\documentclass{article}
\usepackage{stackengine}
\usepackage[final]{neurips_2025}
\usepackage[dvipsnames]{xcolor}
\usepackage[utf8]{inputenc} 
\usepackage[T1]{fontenc}    
\usepackage{hyperref}       
\usepackage{url}           
\usepackage{booktabs}       
\usepackage{amsfonts}      
\usepackage{nicefrac}       
\usepackage{microtype}     
\usepackage{amsmath}
\usepackage{amssymb}
\usepackage{algorithm}
\usepackage{algpseudocode}
\usepackage{CJK}
\usepackage{graphicx}
\usepackage{wrapfig} 
\usepackage{colortbl}
\usepackage{subcaption}
\usepackage{cleveref}
\usepackage{enumitem}
\usepackage{mathrsfs}
\usepackage{amsthm}
\usepackage{mathtools}
\usepackage{multirow}
\usepackage{tikz}
\usepackage{pifont}   
\usepackage[most]{tcolorbox}
\definecolor{SoftLavender}{HTML}{EEF0F8} 
\definecolor{LavenderEdge}{HTML}{C9B3F5} 
\definecolor{new_blue}{rgb}{0, 0.682, 0.937}
\definecolor{new_green}{rgb}{0.4, 0.733, 0.412}
\definecolor{new_yellow}{rgb}{0.973, 0.863, 0.278}
\definecolor{mygreen}{rgb}{0.1608, 0.506, 0.455}

\newtcolorbox{purpleblock}[1][]{%
  enhanced, breakable,
  colback=SoftLavender,   
  colframe=LavenderEdge,  
  boxrule=0pt,            
  arc=2mm,                
  left=8pt,right=8pt,top=6pt,bottom=6pt, 
  before skip=8pt, after skip=8pt,
  #1
}
\newtcolorbox{whiteblock}[1][]{%
  enhanced, breakable,
  colback=white,   
  colframe=white,  
  boxrule=0pt,            
  arc=2mm,                
  left=8pt,right=8pt,top=6pt,bottom=6pt, 
  before skip=8pt, after skip=8pt,
  #1
}

\newcommand{\symbox}[2]{%
  \tikz[baseline=(char.base)]%
    \node[
      fill=#1,          
      rounded corners=1.5pt,
      inner sep=0pt,     
      minimum width=1.8em, 
      minimum height=0.9em,
      text=black,
      font=\tiny,
    ](char){#2};}
\newcommand{\checklash}{%
    \ensurestackMath{%
    \stackinset{l}{1.1pt}{c}{1pt}{%
        \rotatebox{-45}{\scalebox{0.7}{$-$}}%
    }{\checklash}%
    }%
}

\newcommand{\greenP}{\symbox{new_yellow!55}{PA}}
\newcommand{\pinkV} {\symbox{new_blue!25}{VP}}

\newtheorem{assumption}{Assumption}

\newtheorem{definition}{Definition}
\newtheorem{lemma}{Lemma}
\newtheorem{theorem}{Theorem}

\newtheorem*{remark}{Remark}

\Crefname{assumption}{Assumption}{Assumptions}
\Crefname{assumption}{Assumption}{Assumptions}

\title{
Why 1\;+\;1 < 1 in Visual Token Pruning: Beyond Na\"{\i}ve Integration via Multi-Objective Balanced Covering}

\author{%
  \makebox[\textwidth][c]{%
    {\setlength{\tabcolsep}{2.5pt}%
    \begin{tabular}{ccccc}
      Yangfu Li$^{\,\clubsuit}$, & Hongjian Zhan$^{\,\clubsuit,\,\heartsuit,\,}$\thanks{Corresponding Author.} , & Tianyi Chen$^{\spadesuit}$, & Qi Liu$^{\,\clubsuit}$, Yu-jie Xiong$^{\diamondsuit}$, & Yue Lu$^{\,\clubsuit}$
    \end{tabular}%
    }%
  }\\
  \makebox[\textwidth][c]{%
    \begin{tabular}{c}
      {\normalfont $^{\clubsuit}$ East China Normal University, $^{\spadesuit}$ Shanghai Jiao Tong University} \\
      {\normalfont $^{\heartsuit}$ Chongqing Institute of East China Normal University} \\
      {\normalfont $^{\diamondsuit}$ Shanghai University of Engineering Science} \\
      {\texttt{\{yfli\_cee, qiliu\}@stu.ecnu.edu.cn,\ hjzhan@cee.ecnu.edu.cn}} \\
      {\texttt{guleimurray@sjtu.edu.cn,\ xiong@sues.edu.cn,\ ylu@cs.ecnu.edu.cn}}
    \end{tabular}%
  }
}

\begin{document}
\maketitle

\begin{abstract}
\label{Sec:Abstract}
Existing visual token pruning methods target prompt alignment and visual preservation with static strategies, overlooking the varying relative importance of these objectives across tasks, which leads to inconsistent performance. To address this, we derive the first closed-form error bound for visual token pruning based on the Hausdorff distance, uniformly characterizing the contributions of both objectives. Moreover, leveraging $\epsilon$-covering theory, we reveal an intrinsic trade-off between these objectives and quantify their optimal attainment levels under a fixed budget. To practically handle this trade-off, we propose Multi-Objective Balanced Covering (MoB), which reformulates visual token pruning as a bi-objective covering problem. In this framework, the attainment trade-off reduces to budget allocation via greedy radius trading. MoB offers a provable performance bound and linear scalability with respect to the number of input visual tokens, enabling adaptation to challenging pruning scenarios. Extensive experiments show that MoB preserves 96.4\% of performance for LLaVA-1.5-7B using only 11.1\% of the original visual tokens and accelerates LLaVA-Next-7B by 1.3-1.5$\times$ with negligible performance loss. Additionally, evaluations on Qwen2-VL and Video-LLaVA confirm that MoB integrates seamlessly into advanced MLLMs and diverse vision-language tasks. 
The code is available at \url{https://github.com/YChenL/MoB}.
\end{abstract}
\section{Introduction}
\label{Sec:1}

Multimodal large language models (MLLMs) have shown impressive performance across a variety of vision-language tasks, including visual understanding~\cite{liu2023llava,lin2024videollavalearningunitedvisual,li2024llava}, visual question answering~\cite{shao2023prompting,hu2024bliva,peng2024live}, and visual-language reasoning~\cite{chen2023large,wu2024visionllm,wang2024exploring}. Since visual data exhibits much higher spatial redundancy than language, MLLMs are typically required to encode visual inputs as numerous tokens, resulting in substantial computational overhead.

\begin{figure*}[t]
	\centering
	\includegraphics[width=1\linewidth]{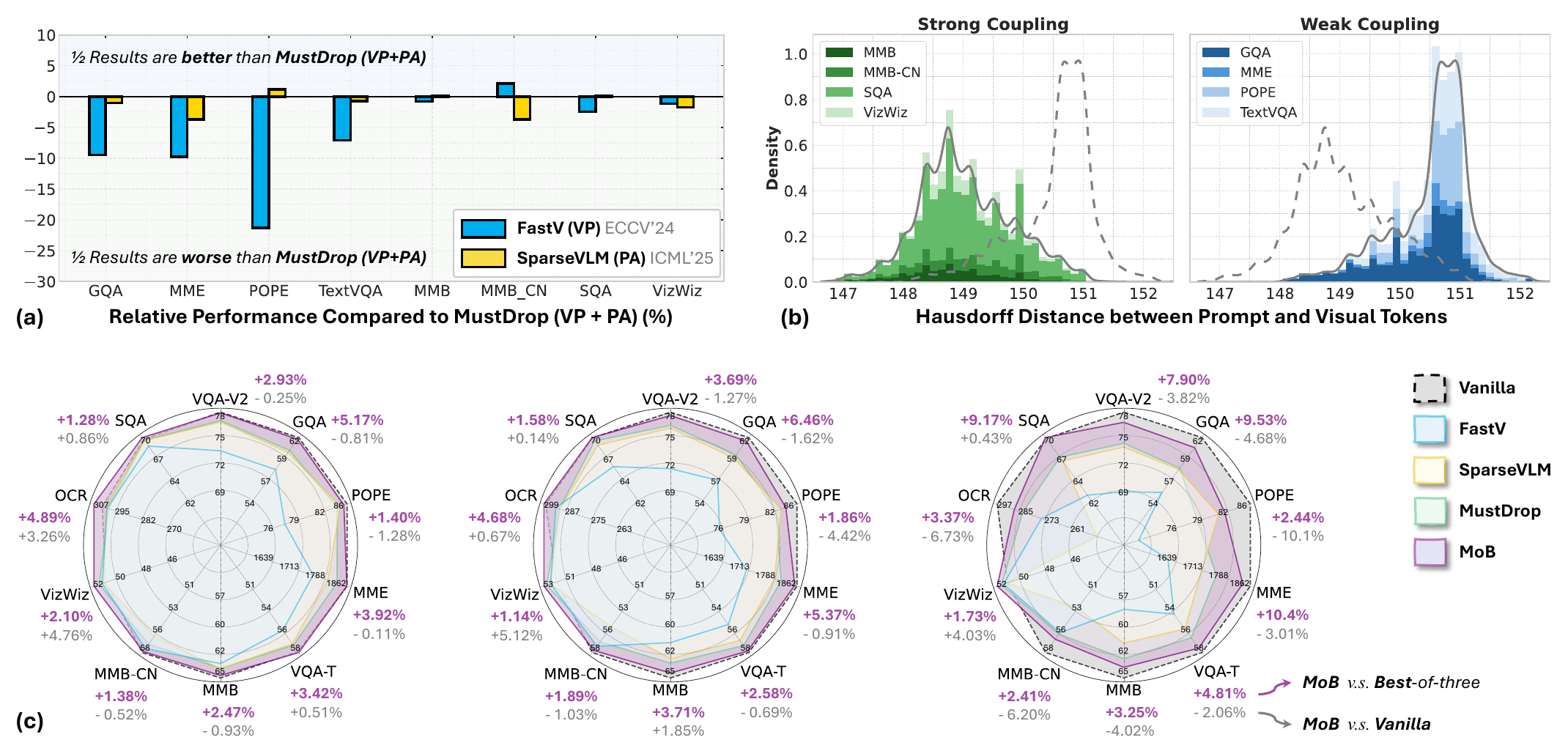}
	\caption{(a) Comparison of single- vs. bi-objective pruning methods on LLaVA-1.5-7B at a $66.7\%$ pruning rate; (b) distribution of the prompt-visual coupling, revealing two distinct patterns across various tasks: weak coupling (large distance) and strong coupling (small distance); (c) radar charts of LLaVA-1.5-7B with visual tokens reduced from $576$ to $192$, $128$, and $64$ (\emph{left-to-right}), demonstrating the consistent improvements of MoB across 10 well-recognized benchmarks.}
	\label{fig:1}
    \vspace{-1.mm}
\end{figure*}

To address this issue, visual token pruning methods are proposed to accelerate MLLMs by selecting representative subsets of visual tokens. Most pruning methods focus on two distinct objectives: Visual Preservation (VP)~\cite{bolyatoken,chen2024image,zhong2024aim,wen2025stop}, which retains tokens by minimizing redundancy or maximizing visual salience, and Prompt Alignment (PA)~\cite{zhang2024sparsevlm,yang2024visionzip,xing2024pyramiddrop}, which selects tokens most relevant to the prompt. Recently, several multi-objective approaches~\cite{liu2024multi,yang2024visionzip,tan2025tokencarve} have been proposed to integrate VP and PA through various complex strategies. Counterintuitively, these methods do not exhibit dominant superiority compared to single-objective approaches, as shown in~\Cref{fig:1}(a). This observation naturally raises a question: \emph{Does integrating different objectives offer fundamental advantages?}

Inspired by this question, we formulate preservation using the \emph{Hausdorff distance} between the original and pruned token sets and derive the first closed-form error bound for visual token pruning (\Cref{lem:1}). This bound depends on VP and PA, while it is also affected by a prompt-visual coupling, measured by the Hausdorff distance between prompt and visual tokens. Notably, we identify two patterns of this coupling across popular benchmarks, as presented in~\Cref{fig:1}(b): weak coupling with large distance (\emph{e.g.}, TextVQA, POPE) and strong coupling with small distance (\emph{e.g.}, MMB, VizWiz). Our further analysis reveals that the effectiveness of the pruning objectives varies under distinct coupling patterns (\Cref{lem:1-relax}). However, existing multi-objective methods overlook this variation and integrate VP and PA via constant strategies, yielding inconsistent improvements over single-objective baselines.

To quantify the effect of prompt-visual coupling, we reexamine visual token pruning from a geometric covering perspective. In this view, the retained tokens can be thought of as the union of two disjoint covers for prompt and visual tokens, where each objective corresponds to a Hausdorff covering radius, and the prompt-visual coupling is represented by the inter-cover diameter. By analyzing the geometric relationship between the radii and the diameter, we reveal an intrinsic trade-off between the two objectives (\Cref{theorem:1}), which \emph{identifies the optimal attainment level of each objective to achieve the performance ceiling under a fixed pruning budget and prompt-visual coupling.} 

For a practical solution to this trade-off, we propose Multi-objective Balanced Covering (MoB), a training-free visual token pruning method with provable performance guarantees and multilinear complexity (\Cref{theorem:2}). MoB partitions the retained tokens into two disjoint subsets for PA and VP, employing greedy radius-trading strategies to reduce the trade-off in objective attainment to a budget allocation problem. This allows MoB to achieve the optimal balance under each coupling pattern by selecting appropriate subset sizes. As shown in~\Cref{fig:1}(c), MoB consistently outperforms both single-objective and multi-objective baselines by a clear margin at identical pruning rates. Besides, MoB accelerates LLaVA-Next-7B by $1.3$-$1.5\times$ with negligible performance loss. Ablation studies further validate our theoretical analysis.
Our key contributions are summarized as follows:
\begin{enumerate}[label=*, leftmargin=10pt, itemsep=3.5pt, topsep=2pt]
    \item[\ding{182}] To our knowledge, we present the first closed-form error bound for visual token pruning and its practical relaxation, characterizing the contributions of the two objectives to preservation quality.

    \item[\ding{183}] We quantify the trade-off between the objectives and identify their optimal attainment level under a fixed budget and prompt-visual coupling, offering valuable insights into visual token pruning.

    \item[\ding{184}] We propose Multi-objective Balanced Covering (MoB) for training-free visual token pruning, which reduces the trade-off of objective attainment to a budget allocation problem via two greedy radius-trading strategies, yielding both a provable performance guarantee and multilinear scalability.

    \item[\ding{185}] Extensive experiments across 14 public benchmarks demonstrate the superiority of MoB. For instance, it retains $96.4\%$ and $97.9\%$ performance for LLaVA-1.5-7B and Video-LLaVA-7B with an $88.9\%$ reduction ratio, outperforming the second-best method by $2.7\%$ and $1.6\%$, respectively. MoB can also be readily incorporated into advanced MLLMs, such as LLaVA-Next and Qwen2-VL.
\end{enumerate}

\section{Background}
\label{Sec:2}

\subsection{Related Work}
\textbf{Multimodal Large Language Model (MLLM).} 
MLLMs~\cite{liu2023llava,li2023blip,zhu2023minigpt,liu2024improved} have achieved remarkable progress in vision-language reasoning, owing to their robust cross-modality modeling via attention mechanisms~\cite{vaswani2017attention,lu2019vilbert}. However, the spatial redundancy inherent in visual signals typically leads to a large number of input tokens~\cite{liangevit,li2024mini,liu2024llavanext,wang2024qwen2}, particularly in high-resolution images and multi-frame videos (\emph{e.g.}, 2048 tokens in Video-LLaVA~\cite{lin2024videollavalearningunitedvisual}). This issue exacerbates the quadratic scaling problem of attention mechanisms, posing significant computational challenges. Moreover, to further enhance the visual capability by incorporating high-quality details, advanced MLLMs are now designed to support higher resolution images~\cite{li2024monkey,chen2024internvl,chen2024expanding,bai2025qwen2}, thereby necessitating the processing of even more visual tokens (\emph{e.g.}, 2880 tokens in LLaVA-NEXT~\cite{liu2024llavanext}). In these scenarios, effectively selecting representative visual tokens becomes a critical requirement for the real-world application of MLLMs.

\textbf{Visual Token Pruning.}
Due to the spatial redundancy, inputs to MLLMs contain numerous less informative visual tokens. 
Visual token pruning accelerates MLLMs by selectively retaining only the most critical tokens during inference. Existing methods typically focus on either visual preservation (VP)~\cite{bolyatoken,shang2024llava,chen2024image,ye2024fit,zhang2024cls,liu2025compression,wen2025stop} or prompt alignment (PA)~\cite{zhang2024sparsevlm,yang2024visionzip,xing2024pyramiddrop}. VP-driven methods, such as ToMe~\cite{bolyatoken} and LLaVA-PruMerge~\cite{shang2024llava}, reduce redundancy by merging similar tokens, while FastV~\cite{chen2024image} and FasterVLM~\cite{zhang2024cls} select tokens based on visual salience. PA-driven approaches like SparseVLM~\cite{zhang2024sparsevlm} rely on cross-modal attention to identify prompt-relevant tokens. More recently, MustDrop~\cite{liu2024multi} integrates VP and PA through a multi-stage pruning pipeline, reporting notable improvements. Despite these advances, existing methods largely overlook the varying relative importance of VP and PA across different scenarios. In this paper, we formally characterize the contribution of each objective under a fixed pruning budget, and propose an algorithm that balances these objectives per scenario, yielding consistent improvements across diverse pruning conditions.

\subsection{Preliminaries}
\label{Sec:3.1}

\textbf{Pipeline of MLLM.} 
MLLMs perform vision-language reasoning by jointly processing multimodal inputs in a shared representation space. Formally, given visual tokens \(\mathcal{V}^{(1)}\) extracted from the visual inputs and prompt tokens \(\mathcal{P}^{(1)}\) encoded from user prompts, the multimodal input is defined as
\[\textstyle
  \mathcal{X}^{(1)} \;=\; \mathcal{V}^{(1)}\,\sqcup\,\mathcal{P}^{(1)},
  \quad
  \mathcal{V}^{(1)}=\{v_1^{(1)},\dots,v_N^{(1)}\},\ 
  \mathcal{P}^{(1)}=\{p_1^{(1)},\dots,p_L^{(1)}\}
  \subseteq \mathbb{R}^d,
\]
where \(N\) and \(L\) denote the numbers of visual and prompt tokens, respectively. We regard both \(\mathcal{V}^{(1)}\) and \(\mathcal{P}^{(1)}\) as compact sets on $d$-dimensional Euclidean space $(\mathbb{R}^d, \|\cdot\|)$. The input $\mathcal{X}^{(1)}$ is then fed into a language model $\mathcal{F}_{[1,I]}$ with $I$ transformer block, and the final output is given by
\[
  y=\mathcal{F}_{[1,I]}\!\bigl(\mathcal{X}^{(1)}\bigr)\quad \text{where}\quad
  \mathcal{F}_{[1,I]}=f_I\circ f_{I-1}\circ\!\dots\!\circ f_1,
\]
In particular, each $f_\ell$ follows the standard Transformer (\emph{e.g.}, multi-head self-attention~\cite{vaswani2017attention}, layer normalization~\cite{ba2016layer,xu2019understanding}). The intermediate feature for any layer $\ell\in\{2,\dots,I\}$ is defined as
\[
  \mathcal{X}^{(\ell)}
  \;\coloneqq\;
  \mathcal{F}_{[1,\ell-1]}\!\bigl(\mathcal{X}^{(1)}\bigr)
  \;=\;
  \mathcal{V}^{(\ell)}\;\sqcup\;\mathcal{P}^{(\ell)},
  \qquad
  \mathcal{F}_{[1,\ell-1]}
  \;\coloneqq\;
  f_{\ell-1}\circ\!\dots\!\circ f_1,
\]
with $\mathcal{V}^{(\ell)}$ and $\mathcal{P}^{(\ell)}$ representing the visual and prompt tokens after $\ell{-}1$ layers, respectively.

\textbf{Visual Token Pruning.} 
To accelerate MLLMs with minimal performance loss, visual token pruning selectively removes less-informative visual tokens at chosen intermediate layers of the language model $\mathcal{F}_{[1,I]}$. Specifically, for any chosen layer $f_\ell,\ \ell\in\{2,\dots,I\}$, pruning algorithms first select a subset $\mathcal{S}^{(\ell)}\subseteq \mathcal{V}^{(\ell)}$ of size $K$ (\emph{i.e.}, pruning budget) and form the pruned input
\scalebox{0.95}{\(
  \mathcal{X}_{\rm s}^{(\ell)} = \mathcal{S}^{(\ell)}  \sqcup\, \mathcal{P}^{(\ell)}
\)}. The corresponding output before and after pruning are then defined as
\[\textstyle
  y
  = \mathcal{F}_{[\ell,I]}\bigl(\mathcal{X}^{(\ell)}\bigr), \quad
  y_{\rm s}
  = \mathcal{F}_{[\ell,I]}\bigl(\mathcal{X}_{\rm s}^{(\ell)}\bigr)\quad \text{where}\quad  \mathcal{F}_{[\ell,I]}\coloneqq f_I\circ\cdots\circ f_{\ell}.
\]
Finally, the objective of visual token pruning is formulated as
\[
\textstyle
    \mathcal{S}^{{(\ell)}\,*} = \operatorname*{argmin}_{\mathcal{S}^{(\ell)} \subseteq \mathcal{V}^{(\ell)},\ |\mathcal{S}^{(\ell)}| = K}
    \| y - y_{\rm s}\|_2.
\]
\textbf{Notation.} 
For brevity we omit the layer index \((\ell)\) and simply write
\(\mathcal{X} = \mathcal{V}\sqcup \mathcal{P}\)
and
\(\mathcal{X}_{s} = \mathcal{S}\sqcup\mathcal{P}\)
to denote the input and its pruned counterpart at an arbitrary layer \(f_\ell\).
We use \(\mathcal{F}\) to denote any composition mapping of the full model \(\mathcal{F}_{[1,I]}\).
Finally, we let \(\|\cdot\|\) denote the Euclidean norm.

\begin{table}[t]
\centering
\scriptsize
\setlength{\tabcolsep}{4pt}
\begin{tabular}{@{}lp{0.38\linewidth}lp{0.38\linewidth}@{}}
\toprule
\textbf{Notation} & \textbf{Description} & \textbf{Notation} & \textbf{Description} \\
\midrule
 $\ell,\ I$ & Pruning layer index; Final layer index.  & $f_\ell$ & Transformer block at layer $\ell$. \\
$d$ & Embedding dimension. & $\mathcal{V}^{(\ell)}$ & Visual tokens at layer $\ell$, \emph{i.e.}, $\mathcal{V}\subseteq \mathbb{R}^d$. \\
$\mathcal{P}^{(\ell)}$ & Prompt tokens at layer $\ell$, \emph{i.e.}, $\mathcal{P}\subseteq \mathbb{R}^d$. &
$\mathcal{S}^{(\ell)}$ & Retained visual tokens at layer $\ell$, \emph{i.e.}, $\mathcal{S}\subseteq \mathcal{V}$. \\
$\mathcal{X}^{(\ell)}$ & All tokens at layer $\ell$, \emph{i.e.}, $\mathcal{X}{=}\mathcal{V}\sqcup\mathcal{P}$. & 
$\mathcal{X}^{(\ell)}_{\rm s}$ & Retained tokens at layer $\ell$, \emph{i.e.}, $\mathcal{X}_{\rm s}{=}\mathcal{S}\sqcup\mathcal{P}$. \\
$N,\ L$ & \#visual tokens, $|\mathcal{V}|{=}N$; \#prompt tokens, $|\mathcal{P}|{=}L$. & $K$ & Pruning budget, $|\mathcal{S}|{=}K$. \\
$\mathcal{F}[1,I]$ & Full model (layers $1\dots I$). & $\mathcal{F}[\ell,I]$ & Submodel from layer $\ell$ to $I$. \\
$y,\ y_{\rm s}$ & Outputs with full tokens $\mathcal{X}$ / pruned tokens $\mathcal{X}_{\rm s}$. & $d_H(\mathcal{A},\mathcal{B})$ & Hausdorff distance between sets $\mathcal{A}$ and $\mathcal{B}$. \\
$C_\ell$ & Lipschitz constant of $\mathcal{F}[\ell,I]$ \emph{w.r.t.} $d_H$. & $\eta$ & Visual-prompt coupling bound: $d_H(\mathcal{V},\mathcal{P})\!\le\!\eta$. \\
$\mathcal{S}_{\rm p},\ \mathcal{S}_{\rm v}$ & Prompt center / Visual center set, $\mathcal{S}{=}\mathcal{S}_{\rm p}\sqcup\mathcal{S}_{\rm v}$. 
& $K_{\rm p},\ K_{\rm v}$ & Budgets for $\mathcal{S}_{\rm p}$ and $\mathcal{S}_{\rm v}$, $K{=}K_{\rm p}+K_{\rm v}$. \\
$\epsilon_{\rm p}$ & Covering radius for $\mathcal{P}$, $d_H(\mathcal{S}_{\rm p},\mathcal{P})$. & $\epsilon_{\rm v}$ & Covering radius for $\mathcal{V}$, $d_H(\mathcal{S}_{\rm v},\mathcal{V})$. \\
$\mathcal{N}(\mathcal{X},\epsilon)$ & Covering number of $\mathcal{X}$ at radius $\epsilon$. & $d_{\mathrm{eff}}$ & Effective dimension of $\mathcal{V},\, \mathcal{P}$. \\
$a,\,b$ & Covering-number lower/upper constants for $\mathcal{P}$. & $a',\,b'$ & Covering-number lower/upper constants for $\mathcal{V}$. \\
$\epsilon_0$ & Validity radius for covering bounds. & $\delta$ & Small dilation radius ($\delta\!\ll\!\eta$). \\
$\mathcal{V}_\delta,\ \mathcal{P}_\delta$ & $\delta$-dilation of $\mathcal{V}$ and $\mathcal{P}$. & $B(c,\epsilon)$ & Ball $\{x:\|x{-}c\|\le\epsilon\}$. \\
$z$ & Radius scaling factor ($>1$). & $D_1$ & Trade-off constant $(4aa')^{1/d_{\mathrm{eff}}}$. \\
$D_2$ & Trade-off constant $1/z^2$. & $\epsilon^\ast$ & Optimal radius $\max\{\eta/z,\ \sqrt{D_1} K^{-1/d_{\mathrm{eff}}}\}$. \\
$k$ & Fold for the proposed nearest-neighbor covering. & $\mathcal{S}'_{\rm p}$ & Candidate set before final truncation by $K_{\rm p}$. \\
$\alpha(\eta,k,L)$ & Alignment constant $\eta\!\left(bkL/a\right)^{1/d_{\mathrm{eff}}}$. & $\beta$ & Preservation constant $2\,(b')^{1/d_{\mathrm{eff}}}$. \\
$\Theta(\cdot)$ & Asymptotically equal (same order); \emph{i.e.}, $\exists\,c_1,c_2>0,\,n_0:c_1 g(n)\le f(n)\le c_2 g(n)$ for $n\ge n_0$. &
$\Omega(\cdot)$ & Asymptotic lower bound (at least on the order of $g$), \emph{i.e.}, $\exists\,c>0,\,n_0: f(n)\ge c\,g(n)$ for $n\ge n_0$. \\
\bottomrule
\end{tabular}
\vspace{2mm}
\caption{Summary of notation used in the theoretical framework.}
\end{table}
\section{Methodology}
\label{Sec:Method}

\subsection{Revisiting Visual Token Pruning: Insights into Prompt-Visual Coupling}
\label{Sec:3.1}
As shown in Fig.~\ref{fig:1}(a), multi-objective pruning methods fail to achieve the expected improvements, and objective-specific methods exhibit inconsistent performance across benchmarks. These observations motivate us to reexamine the problem of visual token pruning. We begin by introducing~\Cref{assump:1}, which quantifies pruning performance in terms of the preservation of the original token set. 
\begin{assumption}[Lipschitz Continuity w.r.t.\ the Hausdorff Distance]
\label{assump:1}
Assume every partial composition \(\mathcal F\) (from layer \(\ell\) to \(I\)) of the language model is Lipschitz continuous w.r.t.\ the Hausdorff distance with constant \(C_\ell\ge1\). Formally, for any intermediate token sets \(\mathcal X,\mathcal X_{\rm s}\subset\mathbb R^d\),
\[\textstyle
  \|\mathcal{F}(\mathcal X)-\mathcal{F}(\mathcal X_{\rm s})\|
  \;\le\;
  C_\ell\;d_H(\mathcal X,\mathcal X_{\rm s}),
\]
where \(d_H\) is the Hausdorff distance induced by the Euclidean norm:
\begin{equation}
\label{Equ:3}
\textstyle
    d_H(\mathcal{X}, \mathcal{X}_{\rm s})\coloneqq \max\Big\{\sup_{x\in\mathcal{X}} \inf_{x_{\rm s}\in\mathcal{X}_{\rm s}} \|x-x_{\rm s}\|,\; \sup_{x_{\rm s}\in\mathcal{X}_{\rm s}} \inf_{x\in\mathcal{X}} \|x-x_{\rm s}\|\Big\}.  
\end{equation}
\end{assumption}
Subsequently, we measure the preservation of the original token set $\mathcal{X}$ using three pairwise distances among visual tokens $\mathcal{V}$, retained tokens $\mathcal{S}$, and prompt tokens $\mathcal{P}$, thereby establishing a unified performance bound for various visual token pruning algorithms, as presented in~\Cref{lem:1}. 

\begin{figure*}[t]
	\centering
	\includegraphics[width=0.98\linewidth]{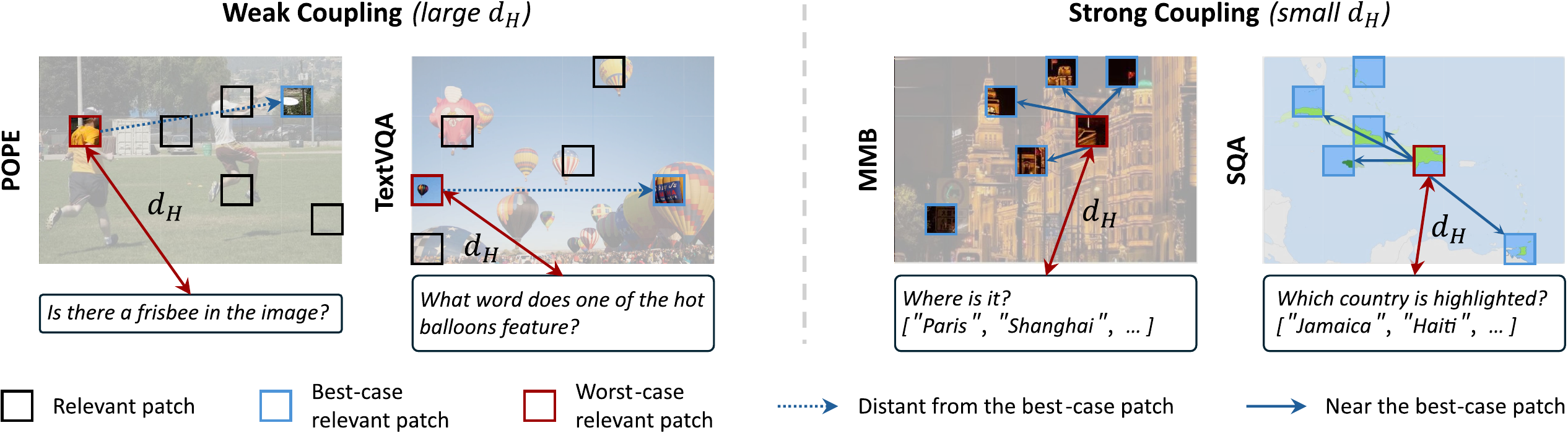}
	\caption{Illustration of prompt-visual coupling with two distinct patterns: In fine-grained tasks (\emph{e.g.} POPE), only a few patches are critical, so the worst-case patch lies far from best-case ones, resulting in a large Hausdorff distance and making prompt alignment valuable. In coarse-grained tasks (\emph{e.g.} MMB), many relevant patches contain the answer cues; thus, the worst-case patch remains close to best-case ones, yielding a small Hausdorff distance and making visual preservation more efficient.}
	\label{fig:2}
\end{figure*}

\begin{purpleblock}
\begin{lemma} [An Error Bound for Visual Token Pruning]
\label{lem:1} 
Under Assump~\ref{assump:1}, given a token set with its pruned counterpart \(\mathcal{X} = \mathcal{V}\sqcup \mathcal{P},\ \ \mathcal{X}_{\rm s} = \mathcal{S}\sqcup\mathcal{P}\subseteq\mathbb{R}^d\), the pruning error bound is given by:
\[\textstyle
    \|\mathcal{F}(\mathcal X)-\mathcal{F}(\mathcal X_{\rm s})\| \;\le\; C_\ell\; \max \Big\{
    \min\big\{ d_H(\mathcal{S},\mathcal{V}),\  d_H(\mathcal{V},\mathcal{P}) \big\},\,
    \min\big\{ d_H(\mathcal{S},\mathcal{V}),\  d_H(\mathcal{S},\mathcal{P}) \big\}
    \Big\}.
\]
\begin{remark}
Here $d_H(\mathcal{S},\mathcal{P})$ and $d_H(\mathcal{S},\mathcal{V})$ describe the prompt alignment and visual preservation, while $d_H(\mathcal{V},\mathcal{P})$ is an inherent term that describes the prompt-visual coupling of input data.     
\end{remark} 
\end{lemma} 
\end{purpleblock}
\emph{Proof} in~\Cref{proof:lem1}.
By~\Cref{lem:1}, in practical settings where \(|\mathcal{S}|\ll |\mathcal{V}|\), pruning performance is governed by a non-trivial interaction among visual preservation, prompt alignment, and prompt-visual coupling. However, existing multi-objective methods typically overlook the coupling term \(d_H(\mathcal V,\mathcal P)\) and statically combine the two objectives across tasks, limiting their effectiveness. Our empirical evidence across popular benchmarks validates two distinct patterns of \(d_H(\mathcal V,\mathcal P)\), each favoring different pruning objectives, as shown in~\Cref{fig:2}. To further explicate the effect of prompt-visual coupling, we introduce~\Cref{assump:2} and propose a practical relaxed error bound in~\Cref{lem:2}.
\begin{assumption}[Prompt-Visual Coupling Bound]
\label{assump:2}
We assume the input visual data and prompts are not entirely unrelated; hence, there exists a constant $\eta>0$ for any intermediate token set \(\mathcal{X} = \mathcal{V}\sqcup \mathcal{P}\subseteq\mathbb{R}^d \) such that \(
d_H(\mathcal{V},\mathcal{P})\le\eta\), ensuring the reasonability of vision-language reasoning.    
\end{assumption} 
\begin{purpleblock}
\begin{lemma} [A Relaxed Error Bound under Practical Budgets] 
\label{lem:1-relax}
Under~\Cref{assump:1,assump:2}, let \(\mathcal{X} = \mathcal{V}\sqcup \mathcal{P},\ \ \mathcal{X}_{\rm s} = \mathcal{S}\sqcup\mathcal{P}\subseteq\mathbb{R}^d\) with $|\mathcal{S}|=K\ll N$. Partition the retained token set $\mathcal{S}$ into two disjoint subsets: $\mathcal{S}=\mathcal{S}_{\rm p}\,\sqcup\,\mathcal{S}_{\rm v}$, devoted to prompt alignment $d_H(\mathcal{S}_{\rm p},\mathcal{P})$ and visual preservation $d_H(\mathcal{S}_{\rm v},\mathcal{V})$, respectively. Then, the pruning error bound reduces to
\[
\textstyle
    \|\mathcal{F}(\mathcal X)-\mathcal{F}(\mathcal X_{\rm s})\| \;\le\; C_\ell\;\max\big\{d_H(\mathcal{S}_{\rm p},\mathcal{P}),\ d_H(\mathcal{S}_{\rm v},\mathcal{V})\big\} + C_\ell\ \eta.
\] 
\end{lemma} 
\end{purpleblock}
\emph{Proof} in~\Cref{proof:lem1-relax}. As~\Cref{lem:1-relax} indicates, under weak coupling (large $\eta$), most visual regions are distant from prompt tokens in the semantic space. Consequently, if $\mathcal S_{\rm p}$ misses the critical patches, $d_H(\mathcal S_{\rm p},\mathcal P)$ dominates the pruning error, making the selection of $\mathcal S_{\rm p}$
\emph{i.e.}, prompt alignment, more significant. Conversely, under strong coupling (small $\eta$), $d_H(\mathcal S_{\rm p},\mathcal P)$ tends to decrease in tandem with $d_H(\mathcal S_{\rm v},\mathcal V)$, reducing the marginal benefit of prompt alignment. To further guide pruning methods design, we next quantify this trade-off governed by $\eta$ through an $\epsilon$-covering argument.

\subsection{Quantifying Prompt-Visual Trade-Off: A Geometric Covering Perspective}
\label{Sec:3.2}
We first introduce some geometric metrics in~\Cref{def:1}, recasting each objective term \(d_H(\mathcal S_{\rm p},\mathcal P)\) and \(d_H(\mathcal S_{\rm v},\mathcal V)\) as covering radii and the coupling term \(d_H(\mathcal V,\mathcal P)\) as an inter-cover diameter. Next, we relate each recasted objective to its token budget $|\mathcal{S}_{\rm p}|, |\mathcal{S}_{\rm v}|$ via covering regularity in~\Cref{lem:2}. Finally, by loading the budget constraint and applying the triangle inequality between radii and diameter, we derive a quantitative trade-off jointly governed by \(K\) and \(\eta\) in~\Cref{theorem:1}.
\begingroup
  \setlength{\abovedisplayskip}{4pt}
  \setlength{\belowdisplayskip}{4pt}
\begin{definition}[$\epsilon$-cover, Covering Number, and Covering Regularity]
\label{def:1}
Let $(\mathbb{R}^d,\|\cdot\|)$ be the $d$-dimensional Euclidean space and let 
$\mathcal{X}\subseteq\mathbb{R}^d$ be a compact set.
\begin{enumerate}[label=(\alph*),leftmargin=16pt,nosep]
\item \textbf{$\epsilon$-cover.}  
if there exists a finite set $\mathcal{C}=\{c_1,\dots,c_M\}\subset\mathbb{R}^d$, an \emph{$\epsilon$-cover} of $\mathcal{X}$ is given by
\[\textstyle
  \mathcal{X}\subseteq\bigcup_{c\in\mathcal{C}} B(c,\epsilon),
  \qquad
  B(c,\epsilon) \coloneqq\{x\in\mathbb{R}^d:\|x-c\|\;\le\;\epsilon\},
\]
where $\mathcal{C}$ is the collection of covering centers, and $\epsilon$ is the covering radius.
\item \textbf{Covering number.}  The minimum cardinality of $\mathcal{C}$ is the \emph{covering number} of $\mathcal{X}$ at radius $\epsilon$:
\[\textstyle
  \mathcal{N}(\mathcal{X},\epsilon)\; \coloneqq\;
  \min\Bigl\{M\in\mathbb{N}\;:\;\exists\,\mathcal{C}\subset\mathbb{R}^d,\ |\mathcal{C}|=M,\
        \mathcal{X}\subseteq\bigcup_{c\in\mathcal{C}}B(c,\epsilon)\Bigr\}.
\]
\item \textbf{Covering regularity.}  
We say that $\mathcal{X}$ satisfies \emph{$d$-dimensional covering regularity} if there exist constants 
$0<A\;\le\; B$ and $\epsilon_0>0$ such that
\[\textstyle
   A\,\epsilon^{-d}\;\le\;
   \mathcal{N}(\mathcal{X},\epsilon)
   \;\le\;
   B\,\epsilon^{-d},
   \qquad\forall\,\epsilon\in(0,\epsilon_0].
\]
\end{enumerate}
\end{definition}
Based on~\Cref{def:1}(a) (b), $\mathcal{S}_{\rm p},\ \mathcal{S}_{\rm v}\subseteq\mathcal{V}$ can be thought of as two collections of centers such that
\[\textstyle
\mathcal{P}\subseteq\bigcup_{i=1}^{K_{\rm p}}B(s_{\rm p}^{(i)},\epsilon_{\rm p}),\ \ \mathcal{V}\subseteq\bigcup_{j=1}^{K_{\rm v}}B(s_{\rm v}^{(j)},\epsilon_{\rm v}),
\]
where the radii are given by
\(
\epsilon_{\rm p}\coloneqq d_H(\mathcal{S}_{\rm p},\mathcal{P}),\ \ 
\epsilon_{\rm v}\coloneqq d_H(\mathcal{S_{\rm v}},\mathcal{V}),
\)
and the covering numbers satisfy
\(
\mathcal{N}(\mathcal{P},\epsilon_{\rm p})\;\le\;|\mathcal{S_{\rm p}}|,\ \ 
\mathcal{N}(\mathcal{V},\epsilon_{\rm v})\;\le\;|\mathcal{S_{\rm v}}|.
\)
Thereby, we derive a lower bound of the required budget, \emph{i.e.}, $|\mathcal{S_{\rm p}}|$, $|\mathcal{S_{\rm v}}|$, to improve each objective, \emph{i.e.}, $\epsilon_{\rm p}$, $\epsilon_{\rm v}$, based on $d_{\mathrm{eff}}$-dimensional covering regularity. 
\begin{lemma}[Covering Number Bounds]
\label{lem:2}
Given $\mathcal{P},\mathcal{V}\subset\mathbb{R}^d$ with an
effective dimension $d_{\mathrm{eff}}$.  Suppose their $\delta$-dilations \smash{$\mathcal{V}_\delta \coloneqq\bigcup_{v\in\mathcal{V}}B(v,\delta)$},
\smash{$\mathcal{P}_\delta \coloneqq\bigcup_{p\in\mathcal{P}}B(p,\delta)$} ($\delta\ll\eta$) satisfy $d_{\mathrm{eff}}$-dimensional covering regularity; thus, there exist constants $b\!>\!a\!>\!0$, $b'\!>\!a'\!>0$ and $\epsilon_0\!>\!\delta$ such that
\[
  a\,\epsilon_{\rm p}^{-d_{\mathrm{eff}}}
  \le\mathcal N(\mathcal P,\epsilon_{\rm p})\le b\,\epsilon_{\rm p}^{-d_{\mathrm{eff}}},
  \qquad
  a'\,\epsilon_{\rm v}^{-d_{\mathrm{eff}}}
  \le\mathcal N(\mathcal V,\epsilon_{\rm v})\le b'\,\epsilon_{\rm v}^{-d_{\mathrm{eff}}},
  \qquad
  \forall\,\epsilon_{\rm p},\epsilon_{\rm v}\in(\delta,\epsilon_0],
\]
\begin{remark}
Previous work suggests that both visual and language embeddings concentrate on a low-dimensional manifold, so the effective covering dimension satisfies the typical relation $d_{\mathrm{eff}}\ll d$.    
\end{remark}
\end{lemma}
\emph{Proof} in~\Cref{proof:lem2}. \Cref{lem:2} demonstrates that once the radius (\emph{i.e.}, the objective) falls below $\epsilon_0$, any further improvement of it demands a $\Theta(\epsilon^{-d_{\mathrm{eff}}})$ increase in the
number of selected token.

By loading~\Cref{lem:2} into the budget constraint: $|\mathcal{S}_{\rm p}|+|\mathcal{S}_{\rm v}|\!=\!K$, and applying a two-step triangle inequality between the covering radii $\epsilon_{\rm p}$, $\epsilon_{\rm v}$ and the inter-cover diameter $\eta$, we establish a $K$-$\eta$-bound in~\Cref{theorem:1}(b), which quantifies the trade-off governed by the budget and prompt-visual coupling.
\begin{purpleblock}
\begin{theorem}[Trade-off between Prompt Alignment and Visual Preservation]
\label{theorem:1}
Under~\Cref{assump:2} and the covering-regularity hypothesis of~\Cref{lem:2} with constants $a,a',d_{\mathrm{eff}}>0$, there exist a radius-scaling factor $z>1$ such that $\eta/z>\delta$ and \(K<\mathcal N(\mathcal P,\eta/z)+\mathcal N(\mathcal V,\eta/z)\), for every pruning results $\mathcal S=(\mathcal S_{\rm p}\sqcup\mathcal S_{\rm v})\subseteq\mathcal{V}$ with budget $K$ satisfying
\[
\max\bigl\{D_1K^{-2/d_{\mathrm{eff}}},\;D_2\,\eta^{2}\bigr\}\;\le\;
d_H(\mathcal S_{\rm p},\mathcal P)\;
d_H(\mathcal S_{\rm v},\mathcal V),
\]
where $D_1 \coloneqq(4\,a\,a')^{1/d_{\mathrm{eff}}}>0$, $D_2\coloneqq 1/z^2>0$.
\begin{remark}[Optimal Attainment Level] 
\(D_1\,K^{-2/d_{\mathrm{eff}}}\) is completely determined by the pruning budget, while \(D_2\,\eta^2\) quantifies the effect of prompt-visual coupling. The optimal attainment level per objective is given by \(\epsilon^* =\max\{\eta/z,\;\sqrt{D_1}\,K^{-1/d_{\mathrm{eff}}}\}.\)
Any attempt to reduce one objective below \(\epsilon^*\) forces the other above \(\epsilon^*\), thereby increasing the overall pruning error.
\end{remark}
\begin{remark}[Effect of Budget and Coupling Strength] 
As $K$ decreases, $z$ correspondingly shrinks, ultimately making \(D_2\,\eta^2\) dominate the bound; while as \(K\) increases, both of the terms reduce, thereby diminishing the trade‐off and tightening the overall error bound.
\end{remark}
\end{theorem}
\end{purpleblock}
\endgroup
\emph{Proof} in~\Cref{proof:theorem1}.
\Cref{theorem:1} characterizes the optimal attainment level for each objective under a fixed pruning budget and prompt-visual coupling. However, it is actually very challenging to dynamically determine the attainment level per objective during the pruning process. To address this, we propose Multi-objective Balanced Covering, which leverages the monotonic relationship between covering radii and numbers to reduce the trade-off of attainment to a budget-allocation problem.

\subsection{Multi-Objective Balanced Covering: From Trade-Off to Budget Allocation
}
\label{Sec:3.3}

Motivated by the insights in \S\ref{Sec:3.2}, Multi-objective Balanced Covering (MoB) recasts visual token pruning as bi-objective covering. Specifically, given a token set $\mathcal{X}=\mathcal{V}\,\sqcup\mathcal{P}\subseteq\mathbb{R}^d$ with a budget $K$, the retained token set $\mathcal{S}$ is defined as the union of a prompt center set $\mathcal{S}_{\rm p}$ and a visual center set $\mathcal{S}_{\rm v}$:
\[\textstyle
\mathcal{S}=\mathcal{S}_{\rm p}\,\sqcup\,\mathcal{S}_{\rm v}\subseteq\mathcal{V}\subseteq\mathbb{R}^d\quad \text{where}\quad \mathcal P\subset\bigcup_{i=1}^{K_{\rm p}}B(s_{\rm p}^{(i)},\epsilon_{\rm p}),\ \ \mathcal V\subset\bigcup_{j=1}^{K-K_{\rm p}}B(s_{\rm v}^{(j)},\epsilon_{\rm v}).
\]
MoB then selects the cover centers (\emph{i.e.}, retained tokens) by minimizing the overall maximum radius:
\[
(\mathcal{S}_{\rm p}^*,\ \mathcal{S}_{\rm v}^*)=\operatorname*{arg\,min}_{\substack{
    \mathcal S_{\rm p}\sqcup\mathcal S_{\rm v}\,\subseteq\mathcal V,\; |\mathcal S_{\rm p}|=K_{\rm p},\;|\mathcal S_{\rm v}|=K-K_{\rm p}
  }} \max\{\epsilon_{\rm p}(\mathcal{S}_{\rm p}), \epsilon_{\rm v}(\mathcal{S}_{\rm v})\}.
\]
In practice, MoB solves this problem approximately by two sequential greedy covering procedures: selection of prompt center set $\mathcal{S}_{\rm p}$ with budget $K_{\rm p}$, and selection of visual center set $\mathcal{S}_{\rm v}$ with the remaining budget $K-K_{\rm p}$. By the covering number bounds given in~\Cref{lem:2}, we have
\[
 K_{\rm p}=\Theta(\epsilon_{\rm p}^{-d_{\mathrm{eff}}}), \quad  K-K_{\rm p}=\Theta(\epsilon_{\rm v}^{-d_{\mathrm{eff}}}),
\]
where $d_{\mathrm{eff}}$ is the effective dimension of $\mathcal{V}$, $\mathcal{P}$. Accordingly, by selecting the unique budget $K_{\rm p}$ (\emph{i.e.}, fixing the remaining budget $K - K_{\rm p}$) under each coupling pattern, MoB ensures
\(
  \epsilon_{\rm p},\,\epsilon_{\rm v}
  = \Omega\bigl(\max\{\eta/z,\;\sqrt{D_1}\,K^{-1/d_{\mathrm{eff}}}\}\bigr),
\)
thus yielding provable performance guarantees across scenarios.

\textbf{Normalization.} For efficiency, MoB applies L2 normalization to each $x\in \mathcal X$ so that $\|x\|=1$. Hence, for any token pair \(x_1,x_2\in\mathcal X\), the Euclidean distance can be induced by their cosine similarity:
\[\textstyle
\|x_1-x_2\|=\sqrt{2-2\cos(x_1,x_2)}.
\]

\textbf{Selection of Prompt Center Set $\mathcal{S}_{\rm p}$.} 
Since all $s_{\rm p}\in \mathcal V$ lie outside $\mathcal P$, a typical solution for minimizing the radius $\epsilon_{\rm p}$ is \emph{Nearest-Neighbor covering} (NN covering) \cite{hochbaum1985best}, which uniformly allocates the nearest $s_{\rm p}\in\mathcal V$ for each prompt token. However, the contribution of each prompt token is inequivalent, especially under weak prompt-visual coupling; thus, equal allocation risks missing the ``best-case tokens.'' To remedy this, we introduce a $k$-fold NN covering procedure. Formally, let $L = |\mathcal P|$ and $k>1$ be a hyperparameter; we first utilize a temporary budget of $kL$ to form a candidate set.
\[\textstyle
  \mathcal{S}_{\rm p}'
  = \bigcup_{p \in \mathcal P}
    \operatorname{arg\,topk}_{s \in \mathcal V}\bigl(\cos(s,p),\,k\bigr),
  \quad
  |\mathcal{S}_{\rm p}'| \ge K_{\rm p},
\]
thereby over-sampling the $k$ nearest visual tokens for each prompt token. Subsequently, we refine the candidate set by selecting the final $K_{\rm p}$ centers that maximize their worst-case alignment with $\mathcal P$:
\[\textstyle
  \mathcal{S}_{\rm p}
  = \operatorname{arg\,topk}_{s \in \mathcal{S}_{\rm p}'}
    \bigl(\max_{p \in \mathcal P}\cos(s,p),\,K_{\rm p}\bigr).
\]
By concentrating the limited budget on those visual tokens most strongly aligned with the key prompt tokens, this strategy ensures a better preservation of the critical regions in the visual input. We determine the appropriate $k$ by ablation to avoid the oversampling of a few salient prompt tokens.

\textbf{Selection of Visual Center Set $\mathcal{S}_{\rm v}$.}
Unlike the prompt center selection, each visual center $s_{\rm v}$ lies in $\mathcal{V}$. Thereby, we employ \emph{Farthest Point Sampling} (FPS)~\cite{moenning2003fast} on the remaining tokens, \emph{i.e.}, $\mathcal{V} \setminus \mathcal{S}$, to select the visual centers, which makes the visual centers $\mathcal{S}_{\rm v}$ well-spread over $\mathcal{V}$, minimizing the covering radius $\epsilon_{\rm v}$. Concretely, FPS operates by iteratively selecting the token farthest (\emph{i.e.}, the most different) from the current centers $\mathcal{S}$, where the distance is given by
\[\textstyle
   \operatorname{dist}_{\rm FPS}(s_{\rm v}, \mathcal{S}) = \min_{s \in \mathcal{S}} (1-\cos(s_{\rm v}, s)),\quad \forall s_{\rm v}\in\mathcal{V}\setminus\mathcal{S}.
\]
Subsequently, we initialize the visual centers with the empty set, \emph{i.e.}, $\mathcal{S}_{\rm v}^{\smash{(1)}} \coloneqq \varnothing$. We then successively add the farthest visual token to the current centers $\mathcal{S}_{\rm v}^{\smash{(i)}}\sqcup\mathcal{S}_{\rm p}$ until it contains a total of $K$ elements. Hence, the visual centers at the subsequent iteration, $\mathcal{S}_{\rm v}^{\smash{(i+1)}}$, is given by:
\[
\textstyle
\mathcal{S}_{\rm v}^{\smash{(i+1)}} = \mathcal{S}_{\rm v}^{\smash{(i)}} \sqcup \operatorname*{arg\,max}_{s_{\rm v} \in \mathcal{V} \setminus \bigl(\mathcal{S}_{\rm v}^{\smash{(i)}} \sqcup \mathcal{S}_{\rm p}\bigr)} \operatorname{dist}_{\rm FPS}(s_{\rm v},\ \mathcal{S}_{\rm v}^{\smash{(i)}} \sqcup \mathcal{S}_{\rm p}), \quad \text{for } i \in [1,\dots, K-K_{\rm p}].
\]
More details of the proposed MoB algorithm are provided in~\Cref{sec:apendixA}.
\begin{purpleblock}
\begin{theorem}[Performance Guarantee]
\label{theorem:2}
Under Assump~\ref{assump:1} and the covering-regularity of Lem~\ref{lem:2} with consts $a,a',d_{\mathrm{eff}}\!>\!0$ and $b\!>\!a,\,b'\!>\!a'$, for any budget split $(K_{\rm p},\,K-K_{\rm p})$, covering fold $k$, and token set $\mathcal{X}=\mathcal{V}\sqcup\mathcal{P}\subseteq\mathbb{R}^d$ with $|\mathcal{V}|=N$, $|\mathcal{P}|=L$, $d_H(\mathcal{V},\mathcal{P})\le\eta$, the following hold:
\begin{enumerate}[label=(\alph*),leftmargin=16pt,nosep]
\item \textbf{Performance bound:} The Performance degradation caused by MoB is upper bounded by
\[\textstyle
\|\mathcal{F}(\mathcal{X})-\mathcal{F}(\mathrm{MoB}(\mathcal{X}))\|\le C_\ell \max\Bigl\{
     \alpha(\eta,k,L)\,(K_{\rm p})^{-1/d_{\mathrm{eff}}}, 
     \;\beta\,(K-K_{\rm p})^{-1/d_{\mathrm{eff}}}
 \Bigr\}
 +C_\ell\, \eta
 ,\]
where
\(
\alpha(\eta,k,L)\;=\; \eta\,\bigl(b\,k\,L/a\bigr)^{1/d_{\mathrm{eff}}},\quad
\beta\;=\;2\,(b')^{1/d_{\mathrm{eff}}}.
\)\vspace{0.6mm}
\item \textbf{Multilinear complexity:} The complexity of MoB is given by
\(\textstyle
T_{\text{\rm MoB}}=\mathcal{O}(N\,(L+K)\,d)
.\)
\end{enumerate}

\begin{remark}[Coupling Trade-off]
Under weak coupling (large $\alpha$), minimizing the bound requires a larger $K_{\rm p}$. Conversely, under strong coupling (small $\alpha$), the alignment term decays rapidly, favoring visual preservation (increasing $K-K_{\rm p}$). Specially, under perfect coupling ($\eta=0$), the bound simplifies to \scalebox{0.85}{\(\|\Delta y\|\;\le\;C_\ell\,\beta\,(K-K_{\rm p})^{-1/d_{\mathrm{eff}}}\)}, \emph{i.e.}, MoB reduces to pure visual preservation.
\end{remark}
\begin{remark}[Budget Scaling]
As the budget $K$ increases, the preservation term
\scalebox{0.85}{\(\beta\,(K - K_{\rm p})^{-1/d_{\mathrm{eff}}}\)} decays, requiring a corresponding increase in $K_{\rm p}$ (and thus a reduction in the alignment term) to re-balance the trade‐off and further lower the overall error bound.
\end{remark}
\begin{remark}[Scalability]
MoB exhibits a multilinear scalability \emph{w.r.t} \#visual tokens $N$, \#prompt tokens $L$, and \#retained tokens $K$ ($K, L \ll N$), making it easily adaptable to more challenging scenarios involving large token counts, \emph{e.g.}, higher-resolution inputs or multi-frame video.
\end{remark}
\end{theorem}
\end{purpleblock}
\emph{Proof} in~\Cref{proof:theorem2}.
\vspace{3mm}

\begin{table*}[h]
\centering
\scriptsize
\renewcommand\arraystretch{0.6}
\setlength{\tabcolsep}{1.05mm}{
\begin{tabular}{l c  ccc c  cc ccc c c}
\toprule
\multirow{3}{*}{\textbf{Method}} & \multirow{3}{*}{\textbf{Objectives}} & \multicolumn{4}{c}{\textbf{Strong Coupling}} &  \multicolumn{6}{c}{\textbf{Weak Coupling}}   &  \multirow{3}{*}{\textbf{Avg.}} \\
\cmidrule(lr){3-6} \cmidrule(lr){7-12} &  
& MMB & MMB$_{\rm CN}$ & SQA & VizWiz & GQA & MME & POPE & VQA$\!^\mathrm{T}$ & VQA$\!^\mathrm{V2}$ & OCR &  \\

\midrule
\cellcolor{gray!15}{LLaVA-1.5-7B} & \multicolumn{12}{c}{\cellcolor{gray!15}{\emph{w/o Pruning, $N=576$}; \emph{Token Reduction Rate} = \textbf{0.0\%}}} \\
\textcolor{gray}{Vanilla}~\cite{liu2024improved} & - 
& \textcolor{gray}{64.7} & \textcolor{gray}{58.1} & \textcolor{gray}{69.5} & \textcolor{gray}{50.0} & \textcolor{gray}{61.9} & \textcolor{gray}{1862} & \textcolor{gray}{85.9} & \textcolor{gray}{58.2} & \textcolor{gray}{78.5} & \textcolor{gray}{297} & \textcolor{gray}{100\%} \\
\midrule
\cellcolor{gray!15}{LLaVA-1.5-7B}   & \multicolumn{12}{c}{\cellcolor{gray!15}{\emph{Pruning budget $K=192$}; \emph{Token Reduction Rate} = \textbf{66.7\%}}} \\
FastV (ECCV'24)~\cite{chen2024image}  & \pinkV  &
61.2 & {57.0} & 67.3 & 50.8 & 52.7 & 1612 & 64.8 & 52.5 & 67.1 & 291 & 91.2\% \\
SparseVLM (ICML'25)~\cite{zhang2024sparsevlm}  & \greenP
& 62.5 & 53.7 & 69.1 & 50.5 & 57.6 & 1721 & {83.6} & 56.1 & 75.6 & 292 & 96.3\% \\
MustDrop (24.11)~\cite{liu2024multi}  &  \greenP  \pinkV
& 62.3 & 55.8 & 69.2 & {51.4} & 58.2 & 1787 & 82.6 & 56.5 & 76.0 & 289 & 97.2\% \\
DART (EMNLP'25)~\cite{wen2025stop}   & \pinkV
& {63.6} & {57.0} & {69.8} & 51.2 & {60.0} & {1856} & 82.8 & {57.4} & {76.7} & {296} & {98.8\%} \\
\textbf{MoB} (w/o $\eta$-prior) &  \greenP  \pinkV &
\cellcolor{orange!15}{63.8} & \cellcolor{orange!15}{57.5} & \cellcolor{orange!15}{70.0} & \cellcolor{orange!15}{52.4} & \cellcolor{orange!15}{61.2} & \cellcolor{orange!15}{1858} & \cellcolor{orange!15}{84.5} & \cellcolor{orange!15}{58.2} & \cellcolor{orange!15}{77.9} & \cellcolor{orange!15}{304} & \cellcolor{orange!15}{100.2\%} \\

\quad + $\eta$-prior &  - &
\cellcolor{blue!10}{64.1} & \cellcolor{blue!10}{57.8} & \cellcolor{blue!10}{70.1} & 
\cellcolor{blue!10}{52.5} & \cellcolor{blue!10}{61.4} & \cellcolor{blue!10}{1860} & 
\cellcolor{blue!10}{84.8} & \cellcolor{blue!10}{58.5} & \cellcolor{blue!10}{78.3} & 
\cellcolor{blue!10}{307}  & \cellcolor{blue!10}{100.6\%} \\

\midrule
\cellcolor{gray!15}{LLaVA-1.5-7B} & \multicolumn{12}{c}{\cellcolor{gray!15}{\emph{Pruning budget $K=128$}; \emph{Token Reduction Rate} = \textbf{77.8\%}}} \\
FastV (ECCV'24)  &  \pinkV
& 56.1 & 56.4 & 60.2 & 51.3 & 49.6 & 1490 & 59.6 & 50.6 & 61.8 & 285 & 86.4\% \\
SparseVLM (ICML'25) &  \greenP  
& 60.0 & 51.1 & 67.1 & 51.4 & 56.0 & 1696 & {80.5} & 54.9 & 73.8 & 280 & 93.8\% \\
MustDrop (24.11)  & \greenP  \pinkV
& 61.1 & 55.2 & 68.5 & {52.1} & 56.9 & 1745 & 78.7 & 56.3 & 74.6 & 281 & 95.6\% \\
DART (EMNLP'25)  & \pinkV
& {63.2} & \cellcolor{blue!10}{57.5} & {69.1} & 51.7 & {58.7} & {1840} & 80.1 & {56.4} & {75.9} & {296} & {98.0\%} \\
\textbf{MoB} (w/o $\eta$-prior) &  \greenP  \pinkV &
\cellcolor{orange!15}{63.2} & \cellcolor{orange!15}{57.3} & \cellcolor{orange!15}{69.3} & \cellcolor{blue!10}{52.8} & \cellcolor{orange!15}{60.7} & \cellcolor{orange!15}{1842} & \cellcolor{orange!15}{81.7} & \cellcolor{orange!15}{57.5} & \cellcolor{orange!15}{77.2} & \cellcolor{orange!15}{299} & \cellcolor{orange!15}{99.2\%} \\

\quad + $\eta$-prior & - & 
\cellcolor{blue!10}{63.5} & \cellcolor{blue!10}{57.5} & \cellcolor{blue!10}{69.6} & \cellcolor{orange!15}{52.7} & \cellcolor{blue!10}{60.9} & \cellcolor{blue!10}{1845} & \cellcolor{blue!10}{82.1} & \cellcolor{blue!10}{57.8} & \cellcolor{blue!10}{77.5} & \cellcolor{blue!10}{299} & \cellcolor{blue!10}{99.4\%} \\

\midrule
\cellcolor{gray!15}{LLaVA-1.5-7B} & \multicolumn{12}{c}{\cellcolor{gray!15}{\emph{Pruning budget $K=64$}; \emph{Token Reduction Rate} = \textbf{88.9\%}}}\\
FastV (ECCV'24)  &  \pinkV
& 48.0 & 52.7 & 51.1 & 50.8 & 46.1 & 1256 & 48.0 & 47.8 & 55.0 & 245 & 77.3\% \\
SparseVLM (ICML'25) &\greenP  
& 56.2 & 46.1 & 62.2 & 50.1 & 52.7 & 1505 & 75.1 & 51.8 & 68.2 & 180 & 84.6\% \\
MustDrop (24.11)  & \greenP  \pinkV
& 60.0 & 53.1 & 63.4 & 51.2 & 53.1 & 1612 & 68.0 & 54.2 & 69.3 & 267 & 90.1\% \\
DART (EMNLP'25)   &   \pinkV
& {60.6} & {53.2} &  \cellcolor{blue!10}{69.8} & {51.6} & {55.9} & {1765} & {73.9} & {54.4} & {72.4} & {270} & {93.7\%} \\
\textbf{MoB} (w/o $\eta$-prior) &  \greenP  \pinkV &
\cellcolor{orange!15}{61.7} & \cellcolor{orange!15}{54.2} & \cellcolor{orange!15}{69.7} & \cellcolor{orange!15}{52.0} & \cellcolor{orange!15}{59.0} & \cellcolor{orange!15}{1806} & \cellcolor{orange!15}{77.2} & \cellcolor{orange!15}{57.0} & \cellcolor{orange!15}{75.5} & \cellcolor{orange!15}{277} & \cellcolor{orange!15}{96.3\%} \\

\quad + $\eta$-prior &  - & 
\cellcolor{blue!10}{62.1} & \cellcolor{blue!10}{54.5} & \cellcolor{blue!10}{69.8} & \cellcolor{blue!10}{52.1} & \cellcolor{blue!10}{59.0} & \cellcolor{blue!10}{1806} & \cellcolor{blue!10}{77.2} & \cellcolor{blue!10}{57.0} & \cellcolor{blue!10}{75.5} & \cellcolor{blue!10}{277} & \cellcolor{blue!10}{96.4\%} \\

\midrule
\midrule
\cellcolor{gray!15}{LLaVA-Next-7B} & \multicolumn{12}{c}{\cellcolor{gray!15}{\emph{w/o Pruning, $N=2880$}; \emph{Token Reduction Rate} = \textbf{0.0\%}}}\\
\textcolor{gray}{Vanilla}~\cite{liu2024llavanext} & - 
& \textcolor{gray}{67.4} & \textcolor{gray}{60.6} & \textcolor{gray}{70.1} & \textcolor{gray}{57.6} & \textcolor{gray}{64.2} & \textcolor{gray}{1851} & \textcolor{gray}{86.5} & \textcolor{gray}{64.9} & \textcolor{gray}{81.8} & \textcolor{gray}{517} & \textcolor{gray}{100\%} \\

\midrule
\cellcolor{gray!15}{LLaVA-Next-7B}   & \multicolumn{12}{c}{\cellcolor{gray!15}{\emph{Pruning budget $K=320$}; \emph{Token Reduction Rate} = \textbf{88.9\%}}}\\

FastV (ECCV'24) &   \pinkV  
& 61.6 & 51.9 & 62.8 & 53.1 & 55.9 & 1661 & 71.7 & 55.7 & 71.9 & 374 & 86.4\% \\

SparseVLM (ICML'25)  & \greenP 
& 60.6 & 54.5 & 66.1 & 52.0 & 56.1 & 1533 & 82.4 & 58.4 & 71.5 & 270 & 85.9\% \\

MustDrop (24.11)  & \greenP  \pinkV
& 62.8 & 55.1 & 68.0 & 54.0 & 57.3 & 1641 & 82.1 & \cellcolor{orange!15}{59.9} & 73.7 & 382 & 90.4\% \\

FasterVLM (24.12)~\cite{zhang2024cls}  & \pinkV
 & 61.6 & 53.5 & 66.5 & 52.6 & 56.9 & 1701 & 83.6 & 56.5 & 74.0 & 401 & 89.8\% \\

DART (EMNLP'25)   &  \pinkV
& \cellcolor{orange!15}{65.3} & \cellcolor{orange!15}{58.2} & \cellcolor{orange!15}{68.4} & \cellcolor{orange!15}{56.1} & \cellcolor{orange!15}{61.7} & \cellcolor{orange!15}{1710} & \cellcolor{orange!15}{84.1} & 58.7 & \cellcolor{orange!15}{79.1} & \cellcolor{orange!15}{406} & \cellcolor{orange!15}{93.9\%} \\

\textbf{MoB} (with $\eta$-prior)  & \greenP  \pinkV
& \cellcolor{blue!10}{65.8} & \cellcolor{blue!10}{58.9} & \cellcolor{blue!10}{68.7} & \cellcolor{blue!10}{57.0} & \cellcolor{blue!10}{62.6} & \cellcolor{blue!10}{1760} & \cellcolor{blue!10}{84.4}  & \cellcolor{blue!10}{60.2} & \cellcolor{blue!10}{80.1} & \cellcolor{blue!10}{418} & \cellcolor{blue!10}{95.4\%} \\

\bottomrule
\end{tabular}}
\caption{Partial comparison of image understanding on the LLaVA-7B series. For MoB, we set $K_{\rm p}\in\{64,48,32\}$ and $k\in\{4,6,8\}$, corresponding to token-reduction rates of $\{88.9\%,77.8\%,66.7\%\}$. For MoB with the $\eta$ prior, we use $K_{\rm p}\in\{\tfrac{3K}{8},\tfrac{K}{4},\tfrac{K}{4}\}$ with $k=\tfrac{3K_{\rm p}}{40}$ for strong-coupling benchmarks and $K_{\rm p}\in\{\tfrac{K}{2},\tfrac{7K}{16},\tfrac{5K}{12}\}$ with $k=\tfrac{K_{\rm p}}{8}$ for weak-coupling benchmarks, corresponding to the same token-reduction rates; the pruning layer is fixed at $\ell=2$. {\setlength{\fboxsep}{1pt} \colorbox{blue!10}{Blue} and \colorbox{orange!15}{Orange} denote the best and the second.}
See~\Cref{Implement_Details} for the detailed setting, and see~\Cref{Exp:full_results} for the full results.}
\label{tab:1}
\vspace{-1mm}
\end{table*}

\section{Experimental Results}
\label{Sec:4}

\begin{wraptable}{r}{0.48\textwidth}
\vspace*{0pt} 
\begin{minipage}[t]{\linewidth}
  \captionsetup{font=footnotesize,aboveskip=4pt,belowskip=6pt}%
  \centering
  \scriptsize
  \renewcommand\arraystretch{0.87}
  \setlength{\tabcolsep}{0.5mm}
  \begin{tabular}{l c c c c  c c  c}
    \toprule
    \textbf{Method} & \textbf{GQA} &  \textbf{MME} & \textbf{POPE} & \textbf{VQA$^\mathrm{T}$} & \textbf{MMB} & \textbf{SQA} & \textbf{Avg.} \\
    \midrule
    \cellcolor{gray!15}{Qwen2-VL-7B} & \multicolumn{7}{c}{\cellcolor{gray!15}{\emph{w/o Pruning}; \emph{Token Reduction Rate =} \textbf{0.0\%}}} \\
    \textcolor{gray}{Vanilla}~\cite{wang2024qwen2} & \textcolor{gray}{62.2} & \textcolor{gray}{2317} &  \textcolor{gray}{86.1} & \textcolor{gray}{82.1} & \textcolor{gray}{80.5} & \textcolor{gray}{84.7}& \textcolor{gray}{100\%} \\
    \midrule
    \cellcolor{gray!15}{Qwen2-VL-7B} & \multicolumn{7}{c}{\cellcolor{gray!15}{\emph{Token Reduction Rate =} \textbf{66.7\%}}} \\
    FastV  & 58.0 & 2130 & 82.1 & 77.3 & 76.1 & 80.0 &  94.0\% \\
    DART   & \cellcolor{orange!15}{60.2} & \cellcolor{orange!15}{2245} & \cellcolor{orange!15}{83.9} & \cellcolor{orange!15}{80.5} & \cellcolor{orange!15}{78.9} & \cellcolor{orange!15}{81.4} & \cellcolor{orange!15}{97.0\%} \\
    \textbf{MoB} (with $\eta$) & \cellcolor{blue!10}{61.8} &  \cellcolor{blue!10}{2268} & \cellcolor{blue!10}{84.7} & \cellcolor{blue!10}{81.1} & \cellcolor{blue!10}{79.5} & \cellcolor{blue!10}{82.3} & \cellcolor{blue!10}{98.4\%} \\
    \midrule
    \cellcolor{gray!15}{Qwen2-VL-7B} & \multicolumn{7}{c}{\cellcolor{gray!15}{\emph{Token Reduction Rate =} \textbf{77.8\%}}} \\
    FastV  & 56.7 & 2031 & 79.2 & 72.0 & 74.1 & 78.3 & 91.0\% \\
    DART   & \cellcolor{orange!15}{58.5} & \cellcolor{orange!15}{2175} & \cellcolor{orange!15}{82.1} & \cellcolor{orange!15}{75.3} & \cellcolor{orange!15}{77.3} &  \cellcolor{orange!15}{79.6} & \cellcolor{orange!15}{94.3\%} \\
    \textbf{MoB} (with $\eta$) & \cellcolor{blue!10}{59.4}  & \cellcolor{blue!10}{2203} & \cellcolor{blue!10}{82.8} & \cellcolor{blue!10}{75.8} & \cellcolor{blue!10}{78.1} & \cellcolor{blue!10}{80.4} & \cellcolor{blue!10}{95.2\%}  \\
    \midrule
    \cellcolor{gray!15}{Qwen2-VL-7B} & \multicolumn{7}{c}{\cellcolor{gray!15}{\emph{Token Reduction Rate =} \textbf{88.9\%}}} \\
    FastV  & 51.9 & 1962 & 76.1 & 60.3 & 70.1 & 75.8 & 84.4\% \\
    DART   & \cellcolor{orange!15}{55.5} & \cellcolor{orange!15}{2052} & \cellcolor{orange!15}{77.9} & \cellcolor{orange!15}{61.8} & \cellcolor{orange!15}{72.0} & \cellcolor{orange!15}{77.6}  & \cellcolor{orange!15}{87.4\%} \\
    \textbf{MoB} (with $\eta$) & \cellcolor{blue!10}{56.5} & \cellcolor{blue!10}{2094} & \cellcolor{blue!10}{78.5} & \cellcolor{blue!10}{62.7} & \cellcolor{blue!10}{72.8} & \cellcolor{blue!10}{78.4} & \cellcolor{blue!10}{88.6\%}  \\
    \bottomrule
  \end{tabular}
  \caption{Comparative experiments on image understanding with Qwen2-VL-7B.}
  \label{tab:2}
\end{minipage}
\vspace{-5mm}
\end{wraptable}

\textbf{Experiment Setting.} 
We perform a comprehensive evaluation of the proposed MoB and several representative methods on two visual tasks: image understanding and visual understanding, together with an efficiency analysis. Our experiments employ four popular MLLMs and include a total of $14$ widely recognized benchmarks. For further details regarding the benchmarks, models, baselines, and implement details please refer to~\Cref{Exp:setup}.

\textbf{Image Understanding.}
\Cref{tab:1} and \Cref{tab:2} report the evaluation results across a variety of image-understanding tasks on LLaVA series and Qwen2-VL, respectively. We highlight five key observations: (a) MoB consistently outperforms all baselines on LLaVA-1.5-7B in most cases. This will be more pronounced when incorporating the $\eta$-prior, which highlights the inherent advantage of our approach;
(b) single-objective baselines exhibit complementary strengths under different coupling patterns, whereas MoB consistently outperforms all baselines, demonstrating the benefit of balanced objectives; (c) the superiority of MoB becomes even more significant under aggressive token reduction. Specifically, the improvement of MoB over the best baseline in average scores increases from 1.8\% at a 66.7\% token reduction to 2.7\% at an 88.8\% reduction on LLaVA-1.5-7B; (d) MoB matches the performance of the vanilla LLaVA-1.5-7B with only 33.3\% of visual tokens, which may be attributed to the mitigation of hallucinations caused by redundant tokens; and (e) MoB scales seamlessly to advanced models, preserving 95.2\% performance on Qwen2-VL-7B using only 22.2\% of visual tokens. These observations demonstrate the superiority of MoB in leveraging limited visual tokens while minimizing performance degradation.

\begin{wraptable}{r}{0.48\textwidth}
\vspace*{0pt} 
\begin{minipage}[t]{\linewidth}
  \captionsetup{font=footnotesize,aboveskip=4pt,belowskip=6pt}%
  \centering
  \scriptsize
  \renewcommand\arraystretch{0.915}
  \setlength{\tabcolsep}{0.5mm}
  \begin{tabular}{l cccc c}
    \toprule
    \textbf{Method} & \textbf{TGIF} & \textbf{MSVD} & \textbf{MSRV} & \textbf{ActNet} & \textbf{Avg.} \\
    \midrule
    \cellcolor{gray!15}{Video-LLaVA-7B} & \multicolumn{5}{c}{\cellcolor{gray!15}{{\emph{Token Reduction Rate} = \textbf{0.0\%}}}}\\
    \textcolor{gray}{Vanilla}~\cite{lin2024videollavalearningunitedvisual} & \textcolor{gray}{47.1} & \textcolor{gray}{69.8} & \textcolor{gray}{56.7} & \textcolor{gray}{43.1} & \textcolor{gray}{100\%} \\
    \midrule    
    \cellcolor{gray!15}{Video-LLaVA-7B} & \multicolumn{5}{c}{\cellcolor{gray!15}{\emph{Token Reduction Rate} = \textbf{93.4\%}}} \\  
    FastV (ECCV’24)    & 23.1 & 38.0 & 19.3 & 30.6 & 52.1\% \\
    SparseVLM (ICML'25) & \cellcolor{orange!15}{44.7} & 68.2 & 31.0 & 42.6 & 86.5\% \\
    VisionZip (24.12)~\cite{yang2024visionzip} & 42.4 & 63.5 & 52.1 & \cellcolor{blue!10}{43.0} & 93.2\% \\
    TwigVLM (ICCV'25)~\cite{shao2025growing} & \cellcolor{orange!15}{44.7} & \cellcolor{orange!15}{68.3} & \cellcolor{orange!15}{54.6} & 41.5 & \cellcolor{orange!15}{96.3\%} \\
    \textbf{MoB} (with $\eta$-prior) & \cellcolor{blue!10}{45.3} & \cellcolor{blue!10}{68.8} & \cellcolor{blue!10}{55.2} & \cellcolor{orange!15}{42.8} & \cellcolor{blue!10}{97.9\%} \\
    \bottomrule
  \end{tabular}
  \caption{Comparative experiments on video understanding with Video-LLaVA-7B.}
  \label{tab:3}
\end{minipage}
\vspace{-3mm}
\end{wraptable}

\textbf{Video Understanding.}
As presented in~\Cref{tab:3}, MoB is general and can be readily extended to more challenging video scenarios without incurring additional cost. Specifically, MoB preserves 97.9\% of average performance for Video-LLaVA-7B using only 6.6\% of the visual tokens, which sets new records in most VideoQA benchmarks, achieving 1.6\% and 4.7\% improvements over TwigVLM and VisionZip, respectively. These results validate the generalization ability of MoB.

\begin{figure*}[t]
	\centering
	\includegraphics[width=1\linewidth]{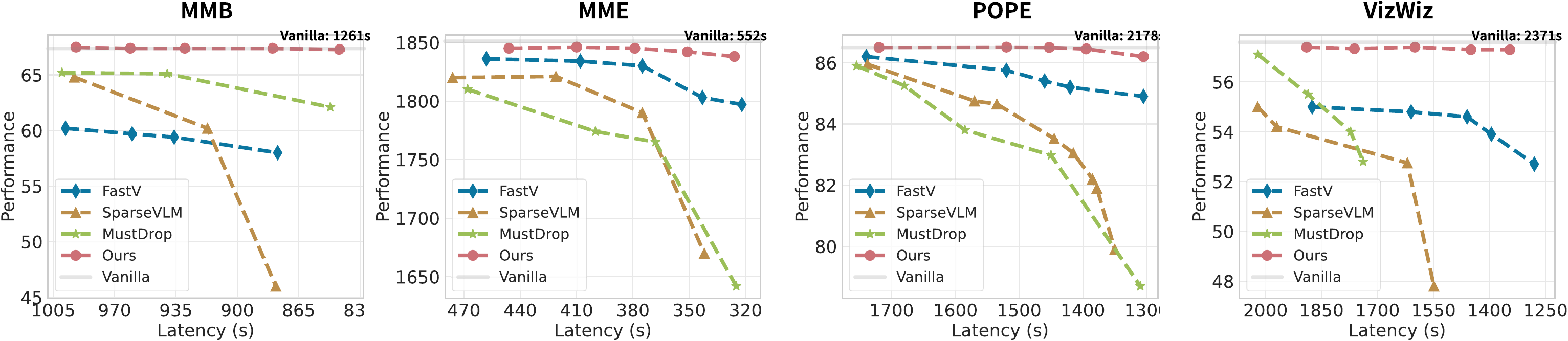}
	\caption{Performance-Latency trade-off comparisons across four benchmarks on LLaVA-Next-7B.}
	\label{fig:3}
    \vspace{-2.5mm}
\end{figure*}

\textbf{Efficiency Analysis.}
We present a performance-latency trade-off measured on an NVIDIA A800-80GB GPU in~\Cref{fig:3}. The results show that (a) MoB achieves a strong performance-latency trade-off, delivering a $1.3$-$1.5\times$ speed-up for LLaVA-NEXT-7B with negligible performance loss; (b) due to ignoring the $K$-$\eta$ trade-off, the multi-stage method MustDrop is outperformed by single-objective methods FastV and SparseVLM on MME and POPE, and suffers significant performance drops as token budgets shrink (\emph{i.e.}, latency decreases). In contrast, MoB consistently maintains a robust trade-off across all benchmarks, surpassing all the baselines by a clear margin; (c) MoB does not rely on attention scores to identify important tokens, making it compatible with flash attention and more efficient than attention-based methods such as SparseVLM and FastV.

\section{Ablation and Discussion}
\label{Sec:5}

\begin{figure*}[t]
	\centering
	\includegraphics[width=1\linewidth]{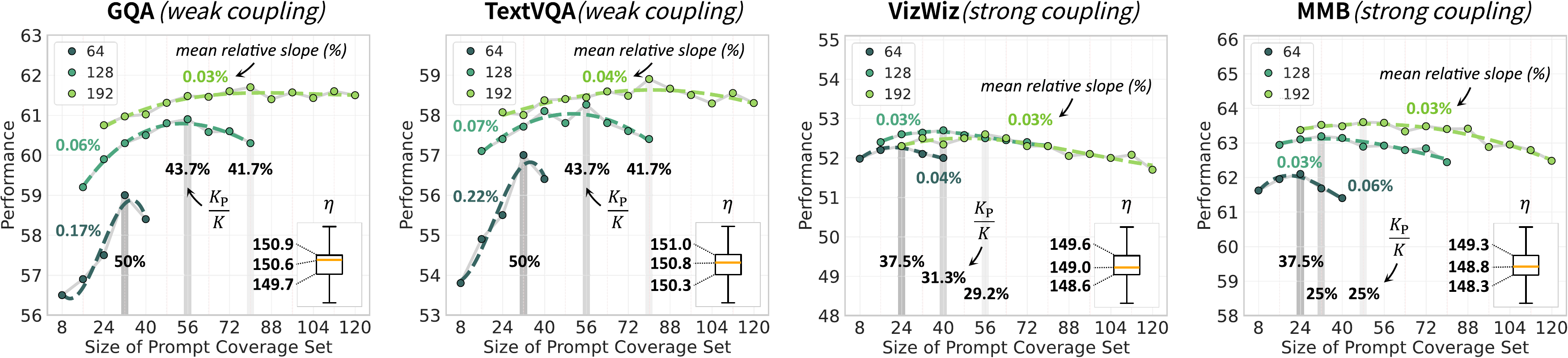}
	\caption{Comprehensive ablation on the budget configuration $\langle K_{\rm p}, K\rangle$ across four benchmarks with distinct prompt-visual coupling $\eta$ on LLaVA-1.5-7B, where $K=\{64, 128, 192\}$; the \emph{mean relative slope (\%)} is given by $\tfrac{100}{x_n-x_1}\sum_{i=1}^{n-1}\tfrac{y_{i+1}-y_i}{y_i}$, quantifying the trade-off intensity; the ratio $\tfrac{K_{\rm p}}{K}$ reflects the cost-effectiveness of prompt alignment, and the box plot presents the distribution of $\eta$.}
	\label{fig:4}
\end{figure*}

\textbf{Impact of $\langle K,\eta, K_{\rm p},\rangle$.} 
We study the impact of $K$, $\eta$, and $K_{\rm p}$ on pruning performance across four benchmarks: GQA and TextVQA (weak coupling); VizWiz and MMB (strong coupling). As shown in~\Cref{fig:4}, the results can be interpreted by~\Cref{theorem:1} and~\Cref{theorem:2}(a), respectively. \\
\emph{A.~\Cref{theorem:1} Perspective:}
When $K$ is large, \emph{e.g.}, $K=192$,  the trade-off is governed by $D_1K^{-2/d_{\mathrm{eff}}}$, hence the trade-off intensity remains nearly identical across benchmarks. Conversely, When $K$ is small, especially $K=64$, in weak-coupling benchmarks, the trade-off turns to be governed by $D_2\eta^2$; thus, the trade-off intensity is obviously more pronounced in GQA and TextVQA than that in VizWiz and MMB. These observations exactly confirm the validity of~\Cref{theorem:1}. \\
\emph{B.~\Cref{theorem:2}(a) Perspective.}
(a) Under weak coupling, the alignment term \(\alpha(\eta,k,L)(K_{\rm p})^{-1/d_{\mathrm{eff}}}\) is amplified, which requires a larger \(K_{\rm p}\) to suppress the overall error. However, across benchmarks sharing the same coupling pattern, the optimal \(K_{\rm p}\) values exhibit only minor variation.
(b) Increasing the total budget \(K\) pushes the optimal \(K_{\rm p}\) upward to rebalance the two bound terms. Since the prompt length \(L\) is fixed, adding more tokens yields diminishing returns for prompt alignment, which is reflected in the declining ratio \(K_{\rm p}/K\). These validate the performance bound in~\Cref{theorem:2}(a). \\
Remarkably, the experimental results suggest that simply determining the optimal \(K_{\rm p}\) for each of the two coupling patterns suffices to guarantee effective generalization across all scenarios.

\begin{figure*}[t]   
    \centering
    \begin{minipage}[t]{0.49\textwidth}
        \centering
        \raisebox{-0.3mm}{%
            \includegraphics[width=0.98\linewidth]{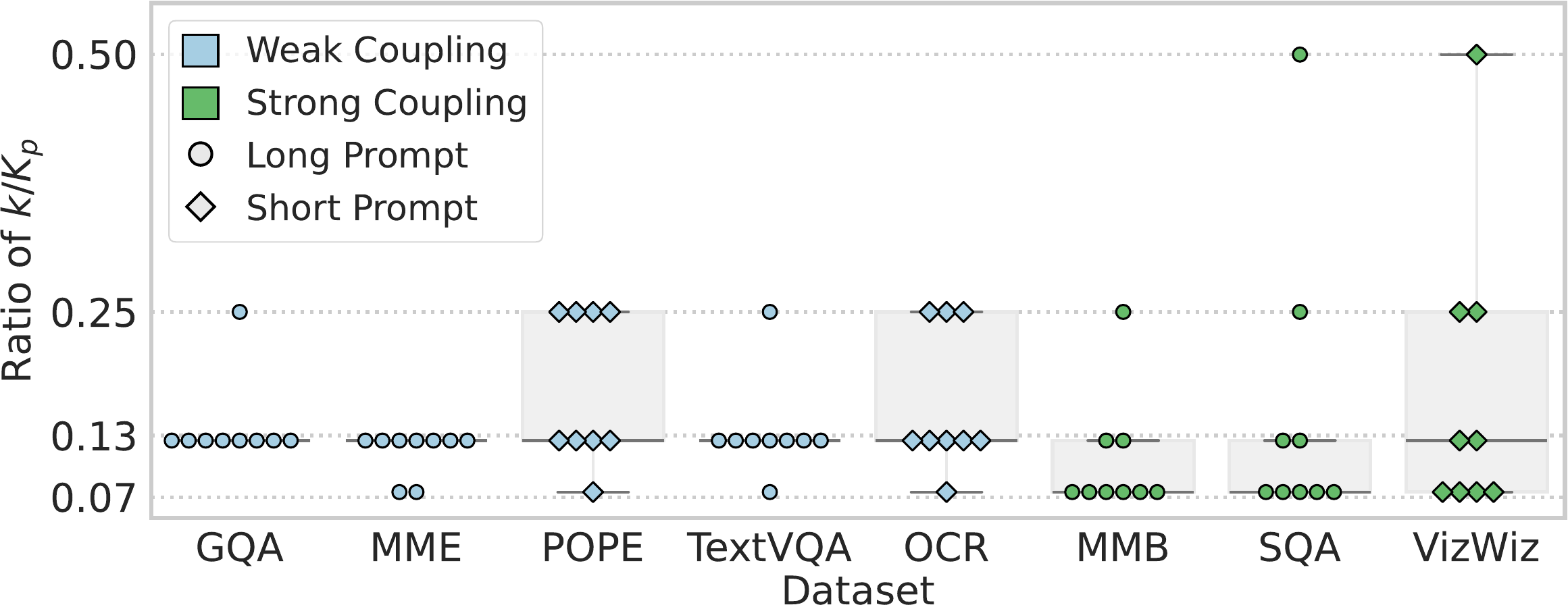}
        }
        \captionof{figure}{Ablation on the ratio of $k/K_{\rm p}$.}
        \label{fig:5}
    \end{minipage}
    \hfill
    \begin{minipage}[t]{0.49\textwidth}
        \centering
        \includegraphics[width=0.98\linewidth]{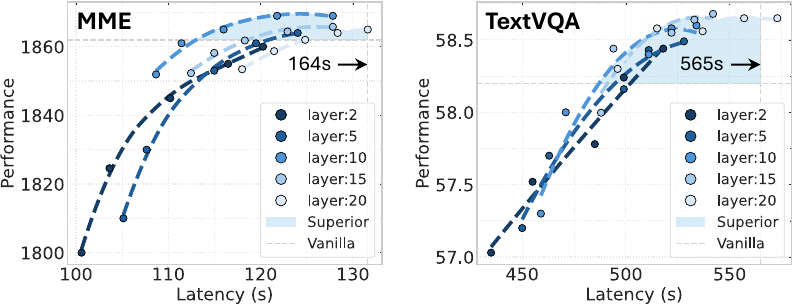}
        \captionof{figure}{Ablation on the pruning layer.}
        \label{fig:6}
    \end{minipage}
\end{figure*}

\textbf{Impact of Covering Fold $k$.}
We chose the covering fold $k$ by examining the normalized ratio $k/K_{\rm p}$ across eight benchmarks and nine budget configurations. As shown in~\Cref{fig:5},
(a) weak-coupling benchmarks generally require a larger $k$ to ensure critical region coverage, whereas strong-coupling settings suffice with a smaller $k$;
(b) benchmarks with longer prompts impose a lower cap on $k$ to preserve sampling diversity and avoid redundant selection of salient tokens.
Notably, weak-coupling benchmarks with long prompts (\emph{e.g.}, GQA, TextVQA) exhibit a narrowly clustered optimal $k/K_{\rm p}$ range, reflecting their strict requirement to cover key tokens without excessive redundancy.

\textbf{Impact of Pruning Layer.}
As shown in~\Cref{fig:6}, (a) models with visual token pruning consistently achieve a more favorable performance-efficiency trade-off than the vanilla model on both benchmarks. 
(b) Pruning in deeper layers provides more significant benefits for the weak-coupling TextVQA than strong-coupling MME. We attribute this to stronger cross-modal interactions in deeper MLLM layers, which facilitate identification of answer-relevant tokens under weak coupling, whereas pruning in shallow layers disrupts these interactions and incurs greater performance degradation. 

\section{Conclusion}
\label{Sec:6}
In this paper, we present a comprehensive analysis of visual token pruning, deriving the first closed-form error bound with a practical relaxation. Leveraging $\epsilon$-covering theory, we quantify the intrinsic trade-off between the fundamental pruning objectives, \emph{i.e.}, visual preservation and prompt alignment, and identify their optimal attainment levels under a fixed pruning budget. Building on these insights, we introduce MoB, a training-free algorithm for visual token pruning. Based on greedy radius trading, MoB ensures the near-optimal attainment per objective via budget allocation, offering a provable performance bound and multilinear scalability. Experimental results indicate that MoB matches the performance (100.6\%) of LLaVA-1.5-7B with only 33.3\% of visual tokens and can be seamlessly integrated into advanced MLLMs, such as LLaVA-Next-7B and Qwen2-VL-7B. Our work advances the understanding of visual token pruning and offers valuable insights for future MLLM compression.

\textbf{Limitations.}
Our theoretical guarantees rely on \cref{assump:1}, which is generally satisfied in practice but may not hold for all MLLMs. Besides, MoB applies a preliminary search to select the proper $K_{\rm p}$, which potentially introduces extra tuning overhead in practical applications. Future work will focus on developing an adaptive $K_{\rm p}$ selection mechanism driven by online estimation of the coupling $\eta$.

\section*{Acknowledgments}
The work was performed at the Shanghai Key Laboratory of Multidimensional Information Processing, East China Normal University; the Institute of Natural Sciences and School of Mathematical Sciences, Shanghai Jiao Tong University; and the Chongqing Key Laboratory of Precision Optics, Chongqing Institute of East China Normal University, with joint support from the National Natural Science Foundation of China (62176091), the Natural Science Foundation of Chongqing (CSTB2024NSCQ-MSX0877), the Science and Technology Commission of Shanghai Municipality (21DZ2203100) and the Fundamental Research Funds for the Central Universities.

\bibliographystyle{plain} 
\bibliography{mybib}  

\appendix

\setcounter{definition}{0}
\setcounter{lemma}{0}

\renewcommand{\thedefinition}{F.\arabic{definition}}
\renewcommand{\thelemma}{F.\arabic{lemma}}

\newpage
\section*{Appendix}
\addcontentsline{toc}{section}{Appendix}

In the appendix, we provide additional information as listed below:
\begin{itemize}[leftmargin=1em]
    \item \S\ref{sec:SI} provides the broader impacts of MoB
    \item \S\ref{sec:apendixA} provides the algorithm details and pseudocode of MoB
    \item \S\ref{Exp:setup} provides the overview of the data, models, baselines and implementation details.
    \item \S\ref{Exp:add_results} provides the additional experimental results.
    \item \S\ref{sec:proof} provides the omitted technical details.
\end{itemize}

\section{Broader Impacts and Limitations}
\label{sec:SI} 

\paragraph{Theory Impacts.}
Beyond the \emph{visual} setting, MoB’s theoretical lens—balancing \emph{Visual Preservation} (retaining sufficient context) and \emph{Prompt Alignment} (isolating “golden evidence”)—naturally transfers to \emph{language} domain. It makes the key challenging (context vs.\ evidence) in long-context LLM explicit and offers actionable guidance for token-level compression and scheduling under fixed context budgets. In practice, this perspective informs RAG (calibrating recall vs.\ precision) and summarization/LLM memory (trading coherence vs.\ conciseness). 

\paragraph{Application Impacts.} 
The proposed MoB yields substantial acceleration of MLLMs with negligible performance loss, thereby enabling high-resolution vision-language models to operate on resource-constrained platforms such as edge devices and mobile systems while supporting low-latency applications—including assistive technologies for the visually impaired, autonomous navigation, and AR/VR. Besides, MoB potentially benefits other redundancy-heavy domains (\emph{e.g.}, point clouds and multi-sensor fusion), guiding efficient token-level compression beyond vision.

\paragraph{Theory Limitations.}
The theoretical analysis (\cref{lem:1,theorem:1}) and the performance guarantees (\cref{theorem:2}) rely on \cref{assump:1} (Lipschitz Continuity) and \cref{lem:2} (Covering Regularity). In embedding spaces that violate metric properties or exhibit highly irregular token distributions, these conditions may fail to hold, and the provable performance bounds may no longer apply.

\paragraph{Application Limitations.}
Our deployment presently requires an \emph{a priori} estimate of \(\eta\) to set the pruning hyperparameters \(K_p\) and \(k\). When \(\eta\) is misestimated for a new model, domain, or input distribution, the selected \(K_p\) and \(k\) can deviate from their optimum, leading to suboptimal speed–accuracy trade-offs.


\newpage
\section{Algorithm}
\label{sec:apendixA}

\begin{algorithm}[H]
\caption{Multi-Objective Balanced Covering (MoB)}
\label{alg:MoB}
\begin{algorithmic}[1]
\Require 
  Visual token $\mathtt{\mathbf{V}\in\mathbb{R}^{N\times d}}$,
  Prompt token $\mathtt{\mathbf{P}\in\mathbb{R}^{L\times d}}$,
  Budget $\mathtt{K_{p}, K_{v}}$, Covering fold $\mathtt{k}$
\Ensure
  Index list for select tokens $\mathtt{\mathbf{S}\in\mathbb{N}^{K_{p}+K_{v}}}$ 
\State  {Normalize all token embeddings to unit $\ell_2$ norm:}
\(\mathtt{
  \mathbf{V} \gets \mathbf{V} 
  / \|\mathbf{V}\|_{2,\mathrm{row}},\quad
  \mathbf{P} \gets \mathbf{P} 
  / \|\mathbf{P}\|_{2,\mathrm{row}}
}\)
\vspace{3mm}
\Statex{\textbf{Step 1. Select Prompt Centers via Nearest-Neighbor Covering}}
\State  {Compute cosine-similarity matrix via $\mathbf{P}\,\mathbf{V}^\top$:}
\(\mathtt{
  \mathbf{M} \gets \mathbf{P}\,\mathbf{V}^\top
}\)
\Comment{\textcolor{mygreen}{$\mathtt{\mathbf{M}\in\mathbb{R}^{L\times N}}$.}}
\State  {Retrieve $k$ nearest token indices per prompt:}
\[\mathtt{
   \mathbf{C}_{idx} \gets ArgTopK(\mathbf{M},\,k,\,axis=1),\quad
   \mathbf{C}_{sim} \gets TopK(\mathbf{M},\,k,\,axis=1)
}\]
\Comment{\textcolor{mygreen}{$\mathtt{\mathbf{C}_{idx},\mathbf{C}_{sim}\in\mathbb{R}^{L\times k}}$ collects index and similarity of $k$ closest centers per prompt token.}}
\Statex\textcolor{mygreen}{\# Deduplicate candidate indices}
\State  {Flatten index and similarity arrays:}
\(\mathtt{
\mathbf{C}_{idx}\gets Flatten(\mathbf{C}_{idx}),\quad 
\mathbf{C}_{sim}\gets Flatten(\mathbf{C}_{sim})
}\)
\Comment{\textcolor{mygreen}{$\mathtt{\mathbf{C}_{idx}\in\mathbb{N}^{Lk}}$, $\mathtt{\mathbf{C}_{sim}\in\mathbb{R}^{Lk}}$}}
\State  {Remove duplicate indices, preserving associated similarities:}
\[\mathtt{
\langle \mathbf{C}_{idx}^*, \mathbf{C}_{sim}^* \rangle \gets UniqueIndices(\mathbf{C}_{idx},\,\mathbf{C}_{sim})
}\]
\Comment{\textcolor{mygreen}{$\mathtt{K_{p}\le |\mathbf{C}_{idx}^*|\le Lk}$}}

\State  {Identify top-$K_{\rm p}$ prompt centers by similarity:}
\(\mathtt{
  \mathbf{i}_p \gets ArgTopK(\mathbf{C}_{sim}^*,\,K_{p})
}\)
\State  {Form the prompt-center index list:}
\(\mathtt{
  \mathbf{S}_{\rm p} \gets \mathbf{C}_{idx}^*[\mathbf{i}_{\rm p}]
}\)
\Comment{\textcolor{mygreen}{$\mathtt{\mathbf{S}_{\rm p}\in\mathbb{N}^{K_{\rm p}}}$}}

\vspace{3mm}
\Statex{\textbf{Step 2. Select Visual Centers via Farthest-Point Sampling}}
\State  {Initialize selected centers:} $\mathbf{S}\gets \mathbf{S}_{\rm p}$

\Statex\textcolor{mygreen}{\# Initialize token-to-prompt minimum distances}
\State  {Compute pairwise minimum distances between all tokens and selected prompt centers:}
\[\mathtt{
  \mathbf{d} \gets \mathbf{1}_{N\times K_{p}} - \mathbf{V}\,\mathbf{V}[\mathbf{S}_{\rm p}]^\top,\quad  \mathbf{d}\gets Min(\mathbf{d},\,axis=1)
}\]
\Comment{\textcolor{mygreen}{Selected centers have zero distance in $\mathbf{d} \in \mathbb{R}^N$.}}
\Statex\textcolor{mygreen}{\# Farthest-Point Sampling}
\For{$t=1$  {to} $K_{\rm v}$}
  \State  {Select the token farthest from current centers:}
    \(\mathtt{
      i^* \gets ArgMax(\mathbf{d}),\quad \mathbf{S}\gets Concat(\mathbf{S},\,i^*)
    }\)
  \Comment{\textcolor{mygreen}{Selected tokens are excluded (distance = 0) from further sampling.}}
  \State  {Compute cosine distances to the newly selected token:}
    \(\mathtt{
      \mathbf{d}_{\Delta}\gets \mathbf{1}_{N} - \mathbf{V}\,\mathbf{V}[i^*]^\top
    }\)
  \State  {Update each token’s minimum distance:}
    \(\mathtt{
      \mathbf{d} \;\gets\; ElementwiseMin(\mathbf{d},\,\mathbf{d}_{\Delta})
    }\)
  \Comment{\textcolor{mygreen}{Distance of newly selected token $\mathtt i^*$ set to zero in $\mathbf{d}$.}}
\EndFor

\State \Return $\mathbf{S}$
\end{algorithmic}
\end{algorithm}

\vspace{7mm}
\begin{algorithm}[H]
\caption{Compute Prompt-Visual Coupling}
\label{alg:hausdorff}
\begin{algorithmic}[1]
\Require 
  Visual embeddings $\mathtt{\mathbf{V}\in\mathbb{R}^{n_v\times d}}$, 
  Prompt embeddings $\mathtt{\mathbf{P}\in\mathbb{R}^{n_p\times d}}$
\Ensure
  Hausdorff distance $\mathtt{h(\mathbf V,\mathbf P)}$
\vspace{1mm}
\Statex{\textbf{Step 1. Compute Pairwise Euclidean Distances}}
\State  {Compute distance matrix via $\mathrm{cdist}$:}
\(
  \mathtt{\mathbf{D}\;\gets\;\mathrm{cdist}(\mathbf{V},\,\mathbf{P},\,p=2)}
\)
\Comment{\textcolor{mygreen}{$\mathtt{\mathbf{D}\in\mathbb{R}^{n_v\times n_p}}$}}
\vspace{1mm}
\Statex{\textbf{Step 2. Directed Hausdorff Distances}}
\State  {Visual-to-prompt directed distance:}
\[
  \mathtt{d_{v\to p},\,\_ \;\gets\; \min(\mathbf{D},\,\mathrm{axis}=2)}
  \quad,\quad
  \mathtt{h_{v\to p}\;\gets\;\max(d_{v\to p})}
\]
\State  {Prompt-to-visual directed distance:}
\[
  \mathtt{d_{p\to v},\,\_ \;\gets\; \min(\mathbf{D},\,\mathrm{axis}=1)}
  \quad,\quad
  \mathtt{h_{p\to v}\;\gets\;\max(d_{p\to v})}
\]
\Statex{\textbf{Step 3. Final Hausdorff Distance}}
\State \Return $\mathtt{\max\bigl(h_{v\to p},\,h_{p\to v}\bigr)}$
\end{algorithmic}
\end{algorithm}

\newpage
\section{Experiment Details}
\label{Exp:setup}
\subsection{Benchmarks}
Our experiments evaluate the vision-language reasoning abilities of multimodal large language models using a comprehensive suite of widely recognized benchmarks. For image understanding tasks, we assess performance on ten public benchmarks: GQA, MMBench (MMB) and MMBench-CN (MMB$_{\rm CN}$), MME, POPE, VizWiz, ScienceQA (SQA), VQA$^{\rm V2}$, TextVQA (VQA$^{\rm T}$), and OCRBench (OCR). For video understanding tasks, we conduct experiments on four popular benchmarks: TGIF-QA (TGIF), MSVD-QA (MSVD), MSRVTT-QA (MSRV), and ActivityNet-QA (ActNet). The following section provides a concise overview of these benchmarks:

\textbf{GQA}~\cite{hudson2019gqa} leverages scene graphs, questions, and images to evaluate visual scene understanding and reasoning. By incorporating detailed spatial relationships and object-level attributes, it poses significant challenges for models to perform accurate visual reasoning in complex environments.


\textbf{MMBench}~\cite{MMBench} introduces a hierarchical evaluation framework where model capabilities are dissected into three levels. Level-1 focuses on basic perception and reasoning; Level-2 subdivides these abilities into six distinct sub-skills; and Level-3 further refines the evaluation into $20$ specific dimensions. Its Chinese counterpart, \textbf{MMBench-CN}, adopts a similar structure.


\textbf{MME}~\cite{liang2024survey} rigorously tests perceptual and cognitive abilities across 14 sub-tasks. By employing carefully crafted instruction-answer pairs and succinct instructions, MME minimizes data leakage and provides a robust, fair assessment of a model’s multifaceted performance.


\textbf{POPE}~\cite{li2023evaluating} targets the evaluation of object hallucination by posing binary questions about object presence in images. It quantifies hallucination levels using metrics, \emph{e.g.}, accuracy, recall, precision, and F1 score, offering a precise and focused measure of model reliability.


\textbf{VizWiz}~\cite{gurari2018vizwiz} is a visual question answering benchmark derived from interactions with blind users. Comprising over $31,000$ image-question pairs with $10$ human-annotated answers per query, it encapsulates the challenges of low-quality image capture and conversational spoken queries, thereby emphasizing real-world visual understanding.



\textbf{ScienceQA}~\cite{lu2022learn} spans multiple scientific domains by organizing questions into $26$ topics, $127$ categories, and $379$ skills. This hierarchical categorization provides a diverse and rigorous testbed for evaluating multimodal understanding, multi-step reasoning, and interpretability across natural, language, and social sciences.


\textbf{VQA$^{\rm V2}$}~\cite{goyal2017making} challenges models with open-ended questions based on $265,016$ images that depict a variety of real-world scenes. Each question is paired with $10$ human-annotated answers, facilitating a thorough evaluation of a model’s capacity to interpret and respond to diverse visual queries.


\textbf{TextVQA}~\cite{singh2019towards} focuses on the integration of text within visual content. It evaluates a model’s proficiency in reading and reasoning about textual information embedded in images, thereby requiring a balanced understanding of both visual and linguistic cues.

\textbf{OCRBench}~\cite{liu2024ocrbench} is a comprehensive benchmark for evaluating the OCR capabilities of multi-modal language models across five key tasks: text recognition, scene text-centric and document-oriented VQA, key information extraction, and handwritten mathematical expression recognition.

\textbf{TGIF-QA}~\cite{jang2017tgif} adapts the visual question answering task to the video domain by focusing on GIFs. With $165$K question-answer pairs, it incorporates tasks, \emph{e.g.}, counting repetitions, identifying repeating actions, detecting state transitions, and frame-specific question answering, thereby demanding detailed spatio-temporal analysis.

\textbf{MSVD-QA}~\cite{xu2017video} builds upon the MSVD dataset by pairing $1,970$ video clips with approximately $50.5$K QA pairs. Questions are categorized into five distinct types, \emph{e.g.}, what, who, how, when, and where, making it a versatile tool for evaluating video understanding.

\textbf{MSRVTT-QA}~\cite{chen2011collecting} features $10$K video clips and $243$K QA pairs designed to test the integration of visual and temporal information. Its structure, which parallels that of MSVD-QA through the inclusion of five question types, further enriches the evaluation landscape for video-based tasks.

\textbf{ActivityNet-QA}~\cite{yu2019activitynet} provides $58$K human-annotated question-answer pairs drawn from $5.8$K videos. Its focus on questions related to motion, spatial relationships, and temporal dynamics necessitates long-term spatio-temporal reasoning, thus serving as a benchmark for advanced video understanding.

\subsection{Multi-modal Large Language Models}

We evaluate MoB using various open-source multimodal large language models (MLLMs). For image understanding tasks, experiments are conducted on the LLaVA series, including LLaVA-1.5-7B and LLaVA-Next-7B, as well as the Qwen-VL series, such as Qwen2-VL-7B. Specifically, LLaVA-Next and Qwen2-VL are utilized to validate performance on high-resolution images, \emph{i.e.}, those with a larger number of visual tokens. For video understanding tasks, we employ Video-LLaVA-7B as the baseline model, following the settings reported in its original paper to ensure a fair comparison.

\textbf{LLaVA-1.5-7B}~\cite{liu2024improved} is a robust vision-language model built on the LLaVA framework. It processes images resized to $224\times 224$ and tokenizes them into roughly $572$ visual tokens using a patch-based vision encoder. This design balances fine-grained visual representation with computational efficiency, making it effective for diverse multimodal tasks.

\textbf{LLaVA-Next-7B}~\cite{liu2024llavanext} extends the LLaVA-1.5 by incorporating refined training strategies and data curation. It supports higher-resolution inputs (up to $448\times448$), yielding up to $2880$ visual tokens. These enhancements improve its visual reasoning capabilities and enable more precise alignment between visual content and language but also incur significantly increased computational cost.

\textbf{Qwen2-VL-7B}~\cite{wang2024qwen2} augments the Qwen2 language model with visual input capabilities. This model leverages cross-modal pretraining to seamlessly merge vision and language, demonstrating strong performance in complex visual question answering and comprehensive scene understanding.

\textbf{Video-LLaVA-7B}~\cite{lin2024videollavalearningunitedvisual} extends the LLaVA framework into the temporal domain by processing video inputs. It is designed to capture both spatial and temporal dynamics, enabling effective video comprehension and video-based question answering with coherent and context-aware responses.

\subsection{Baselines}

To validate the superiority of the proposed MoB, we construct a robust baseline that integrates a comprehensive set of representative existing methods, which encompass single-stage methods with both two distinct objectives and several multi-stage methods.

\textbf{ToMe}~\cite{bolyatoken} employs a lightweight token-matching scheme to merge visually similar tokens across transformer layers, thereby reducing computation without additional training. Its simple yet effective design makes it well suited for real-time applications.

\textbf{FastV}~\cite{chen2024image} leverages attention maps in the early layers to identify and prune non-critical tokens, significantly reducing initial computational overhead. This focus on early-stage reduction allows the model to operate more efficiently while maintaining performance.

\textbf{SparseVLM}~\cite{zhang2024sparsevlm} ranks tokens based on cross-modal attention to assess image-prompt relevance and adopts adaptive sparsity ratios to retain key information. It further incorporates a token recycling mechanism to balance the trade-off between efficiency and accuracy.

\textbf{HiRED}~\cite{arif2024hired} allocates token budgets across image partitions by using CLS token attention and then selects the most informative tokens within each partition. This spatially aware approach ensures balanced reduction while preserving contextual details.

\textbf{LLaVA-PruMerge}~\cite{shang2024llava} combines pruning and merging strategies by dynamically removing less important tokens using sparse CLS-visual attention. It then clusters the retained tokens based on key similarity, ensuring that crucial visual features remain intact.

\textbf{PyramidDrop}~\cite{xing2024pyramiddrop} adopts a progressive token-dropping strategy across different model stages, resulting in a pyramid-like token structure. This method carefully balances the reduction of tokens with the preservation of performance as the processing advances.

\textbf{MustDrop}~\cite{liu2024multi} integrates several token-reduction strategies including spatial merging, text-guided pruning, and output-aware cache policies. Its multi-faceted approach efficiently reduces token counts across various stages of the model.

\textbf{VisionZip}~\cite{yang2024visionzip} first selects dominant tokens that capture the majority of an image’s information and then merges the remaining tokens based on semantic similarity. This approach dramatically reduce token redundancy while accelerating inference and maintaining robust performance.

\textbf{FasterVLM}~\cite{zhang2024cls} evaluates token importance using CLS attention in the encoder and prunes tokens before they interact with the language model. This preemptive reduction streamlines the overall process and enhances model efficiency.

\textbf{GlobalCom$^2$}~\cite{liu2025compression} employs a hierarchical strategy by coordinating thumbnail tokens to allocate adaptive retention ratios for high-resolution crops. This approach successfully preserves local details while providing effective global context reduction.

\textbf{DART}~\cite{wen2025stop} leverages token duplication to guide its pruning process instead of relying solely on attention scores. By selecting a small set of pivot tokens and retaining only those with minimal redundancy, DART achieves significant acceleration in a training-free manner.

\textbf{TokenCarve}~\cite{tan2025tokencarve} implements a two-stage, training-free compression framework that preserves critical visual information during aggressive token reduction. It first prunes low-information tokens using an information-preservation guided selection and then merges the remaining tokens based on similarity to minimize accuracy loss.

\textbf{TwigVLM}~\cite{shao2025growing} accelerates large vision-language models by appending a lightweight twig block to an early layer of a frozen base VLM. It utilizes twig-guided token pruning coupled with self-speculative decoding to boost generation speed while retaining high accuracy even under aggressive token reduction.

\subsection{Implement Details}
\label{Implement_Details}
From~\Cref{theorem:1,theorem:2}, the balance between the visual preservation and prompt alignment, \emph{i.e.}, the optimal budget $K_{\rm p}$ applied for covering prompt tokens $\mathcal{P}$, is jointly determined by the total budget $K$ and the visual-prompt coupling $\eta$. To ensure fair comparison, we evaluate two settings.

\textbf{(i) Without $\eta$ prior.}
This setting deliberately avoids any benchmark-specific prior (w/o $\eta$ prior). MoB adjusts $K_{\rm p}$ solely as a function of $K$ to balance the two objectives. Based on an ablation over $\langle K, K_{\rm p}\rangle$, we set
$K_{\rm p}\!\in\!\{64,48,32\}$ and $k\!\in\!\{4,6,8\}$, corresponding to token-reduction rates of $\{88.9\%,77.8\%,66.7\%\}$.

\textbf{(ii) With $\eta$ prior.}
To verify the $K$--$\eta$--$K_{\rm p}$ relationship formulated in~\Cref{theorem:1,theorem:2}, we introduce a coarse benchmark prior on $\eta$. Specifically, we \textbf{do not} meticulously search the optimal hyperparameters for MoB, \emph{i.e.}, $K_{\rm p}$ and the covering fold $k$, per benchmark. Instead, we partition benchmarks by their empirical $\eta$ distribution into two groups (strong \emph{v.s.}\ weak coupling) and employ \textbf{the same configuration per group}. From a joint ablation over $\langle K,\eta,K_{\rm p}\rangle$, for image understanding we set
\[
  \text{strong coupling:}\quad
  K_{\rm p}\in\bigl\{\tfrac{3K}{8},\tfrac{K}{4},\tfrac{11K}{24}\bigr\},\quad
  k=\tfrac{3K_{\rm p}}{40};
\]
\[
  \text{weak coupling:}\quad
  K_{\rm p}\in\bigl\{\tfrac{K}{2},\tfrac{7K}{16},\tfrac{5K}{12}\bigr\},\quad
  k=\tfrac{K_{\rm p}}{8},
\]
which again yield token-reduction rates of $\{88.9\%,77.8\%,66.7\%\}$.

As for video understanding, we set $K_{\rm p}=\tfrac{3K}{8},\ k=\tfrac{3K_{\rm p}}{40}$ for MSVD, MSRV, and ActNet, and $K_{\rm p}=\tfrac{K}{2},\ k=\tfrac{K_{\rm p}}{8}$ for TGIF.
Unless otherwise stated, the pruning layer index is fixed to $\ell=2$ for both image and video tasks. The same configurations are applied across all MLLMs, and all baselines are run with their default settings.

To ensure reproducibility, we cross-validated our experimental results using the publicly available MLLMs evaluation tool \emph{lmms-eval} (v0.3.0)~\cite{zhang2024lmmsevalrealitycheckevaluation}, with the random seed set to $1234$. All experiments were conducted on $4\times$ Nvidia A800-80GB GPUs paired with $2\times$ Intel Xeon\textsuperscript{\textregistered} Gold 6348 CPUs. The implementation was carried out in Python~3.10 using PyTorch~2.1.2 and CUDA~11.8.

\newpage
\section{Additional Experimental Results}
\label{Exp:add_results}

\subsection{Quantitative Comparison}
\label{Exp:full_results}
\vspace{5mm}
\begin{table*}[htbp]
\centering
\scriptsize
\renewcommand\arraystretch{0.9}
\setlength{\tabcolsep}{0.95mm}{
\begin{tabular}{l c  ccc c  cc ccc c c}
\toprule
\multirow{3}{*}{\textbf{Method}} & \multirow{3}{*}{\textbf{Objectives}} & \multicolumn{4}{c}{\textbf{Strong Coupling}} &  \multicolumn{6}{c}{\textbf{Weak Coupling}}   &  \multirow{3}{*}{\textbf{Avg.}} \\
\cmidrule(lr){3-6} \cmidrule(lr){7-12}
&  
& MMB & MMB$_{\rm CN}$ & SQA & VizWiz & GQA & MME & POPE & VQA$\!^\mathrm{T}$ & VQA$\!^\mathrm{V2}$ & OCR &  \\

\midrule

\cellcolor{gray!15}{LLaVA-1.5-7B} & \multicolumn{12}{c}{\cellcolor{gray!15}{\emph{w/o Pruning, $N=576$}; \emph{Token Reduction Rate} = \textbf{0.0\%}}} \\
\textcolor{gray}{Vanilla}~\cite{liu2024improved} & - 
& \textcolor{gray}{64.7} & \textcolor{gray}{58.1} & \textcolor{gray}{69.5} & \textcolor{gray}{50.0} & \textcolor{gray}{61.9} & \textcolor{gray}{1862} & \textcolor{gray}{85.9} & \textcolor{gray}{58.2} & \textcolor{gray}{78.5} & \textcolor{gray}{297} & \textcolor{gray}{100\%} \\
\midrule
\cellcolor{gray!15}{LLaVA-1.5-7B}   & \multicolumn{12}{c}{\cellcolor{gray!15}{\emph{Pruning budget $K=192$}; \emph{Token Reduction Rate} = \textbf{66.7\%}}} \\
ToMe (ICLR'23)~\cite{bolyatoken} & \pinkV & 60.5 & - & 65.2 & - & 54.3 & 1563 & 72.4 & 52.1 & 68.0 & - & \textcolor{gray}{88.5\%} \\
FastV (ECCV'24)~\cite{chen2024image}  & \pinkV  &
61.2 & 57.0 & 67.3 & 50.8 & 52.7 & 1612 & 64.8 & 52.5 & 67.1 & 291 & 91.2\% \\
HiRED (AAAI'25)~\cite{arif2024hired}  & \pinkV & 62.8 & 54.7 & 68.4 & 50.1 & 58.7 & 1737 & 82.8 & 47.4 & 74.9 & 190 & 91.5\% \\
LLaVA-PruMerge (24.05)~\cite{shang2024llava}  & \pinkV & 59.6 & 52.9 & 67.9 & 50.1 & 54.3 & 1632 & 71.3 & 54.3 & 70.6 & 253 & 90.8\% \\
SparseVLM (ICML'25)~\cite{zhang2024sparsevlm}  & \greenP
& 62.5 & 53.7 & 69.1 & 50.5 & 57.6 & 1721 & 83.6 & 56.1 & 75.6 & 292 & 96.3\% \\
PyramidDrop (CVPR'25)~\cite{xing2024pyramiddrop} & \greenP & 63.3 & 56.8 & 68.8 & 51.1 & 57.1 & 1797 & 82.3 & 56.1 & 75.1 & 290 & 96.7\% \\
FiCoCo-V (EMNLP'24)~\cite{zhan2024rethinking} & \pinkV & 62.3 & 55.3 & 67.8 & 51.0 & 58.5 & 1732 & 82.5 & 55.7 & 74.4 & - & \textcolor{gray}{96.1\%} \\
MustDrop (24.11)~\cite{liu2024multi}  &  \greenP  \pinkV
& 62.3 & 55.8 & 69.2 & \cellcolor{orange!15}{51.4} & 58.2 & 1787 & 82.6 & 56.5 & 76.0 & 289 & 97.2\% \\
VisionZip (24.12)~\cite{yang2024visionzip} & \pinkV & 63.0 & - & 68.9 & - & 59.3 & 1783 & \cellcolor{orange!15}{85.3} & 57.3 & 76.8 & - & \textcolor{gray}{97.7\%} \\
DART (EMNLP'25)~\cite{wen2025stop}   & \pinkV
& 63.6 & 57.0 & \cellcolor{orange!15}{69.8} & 51.2 & 60.0 & 1856 & 82.8 & 57.4 & 76.7 & 296 & 98.8\% \\
TokenCarve (25.03)~\cite{tan2025tokencarve} & \greenP \pinkV & 63.0 & - & 69.1 & 50.9 & - & 1830 & 84.9 & \cellcolor{blue!10}{58.4} & 78.0 & - & \textcolor{gray}{99.3\%} \\
TwigVLM (ICCV'25)~\cite{shao2025growing} & \greenP & \cellcolor{blue!10}{64.0} & - & 68.8 & - & \cellcolor{orange!15}{61.2} & 1848 &  \cellcolor{blue!10}{87.2} & 58.0 & \cellcolor{orange!15}{78.1} & - & \textcolor{gray}{99.5\%} \\
\textbf{MoB} (w/o $\eta$-prior) &  \greenP  \pinkV &
\cellcolor{orange!15}{63.8} & \cellcolor{orange!15}{57.5} & \cellcolor{blue!10}{70.0} & \cellcolor{blue!10}{52.4} & \cellcolor{orange!15}{61.2} & \cellcolor{orange!15}{1858} & 84.5 & \cellcolor{orange!15}{58.2} & 77.9 & \cellcolor{orange!15}{304} & \cellcolor{orange!15}{100.2\%} \\
\quad + $\eta$-prior &  - &
\cellcolor{blue!10}{64.1} & \cellcolor{blue!10}{57.8} & \cellcolor{blue!10}{70.1} & 
\cellcolor{blue!10}{52.5} & \cellcolor{blue!10}{61.4} & \cellcolor{blue!10}{1860} & 
84.8 & \cellcolor{blue!10}{58.5} & \cellcolor{blue!10}{78.3} & 
\cellcolor{blue!10}{307}  & \cellcolor{blue!10}{100.6\%} \\

\midrule

\cellcolor{gray!15}{LLaVA-1.5-7B} & \multicolumn{12}{c}{\cellcolor{gray!15}{\emph{Pruning budget $K=128$}; \emph{Token Reduction Rate} = \textbf{77.8\%}}} \\
ToMe (ICLR'23) & \pinkV & 53.3 & - & 59.6 & - & 52.4 & 1343 & 62.8 & 49.1 & 63.0 & - & \textcolor{gray}{80.4\%} \\
FastV (ECCV'24)  &  \pinkV
& 56.1 & 56.4 & 60.2 & 51.3 & 49.6 & 1490 & 59.6 & 50.6 & 61.8 & 285 & 86.4\% \\
HiRED (AAAI'25)  & \pinkV & 61.5 & 53.6 & 68.1 & 51.3 & 57.2 & 1710 & 79.8 & 46.1 & 73.4 & 191 & 90.2\% \\
LLaVA-PruMerge (24.05)  & \pinkV & 58.1 & 51.7 & 67.1 & 50.3 & 53.3 & 1554 & 67.2 & 54.3 & 68.8 & 248 & 88.8\% \\
SparseVLM (ICML'25) &  \greenP  
& 60.0 & 51.1 & 67.1 & 51.4 & 56.0 & 1696 & 80.5 & 54.9 & 73.8 & 280 & 93.8\% \\
PyramidDrop (CVPR'25) & \greenP & 61.6 & 56.6 & 68.3 & 51.0 & 56.0 & 1761 & 82.3 & 55.1 & 72.9 & 287 & 95.1\% \\
FiCoCo-V (EMNLP'24) & \pinkV & 61.1 & 54.3 & 68.3 & 49.4 & 57.6 & 1711 & 82.2 & 55.6 & 73.1 & - & \textcolor{gray}{94.9\%} \\
MustDrop (24.11)  & \greenP  \pinkV
& 61.1 & 55.2 & 68.5 & \cellcolor{orange!15}{52.1} & 56.9 & 1745 & 78.7 & 56.3 & 74.6 & 281 & 95.6\% \\
VisionZip (24.12) & \pinkV & 62.0 & - & 68.9 & - & 57.6 & 1762 & 83.2 & 56.8 & 75.6 & - & \textcolor{gray}{96.2\%} \\
DART (EMNLP'25)  & \pinkV
& \cellcolor{orange!15}{63.2} & \cellcolor{blue!10}{57.5} & 69.1 & 51.7 & 58.7 & 1840 & 80.1 & 56.4 & 75.9 & \cellcolor{orange!15}{296} & 98.0\% \\
TokenCarve (25.03) & \greenP \pinkV & 62.7 & - & 68.9 & 51.0 & - & 1829 & \cellcolor{orange!15}{84.5} & \cellcolor{blue!10}{58.1} & 77.3 & - & \textcolor{gray}{99.0\%} \\
TwigVLM (ICCV'25) & \greenP & \cellcolor{blue!10}{63.5} & - & \cellcolor{blue!10}{69.5} & - & \cellcolor{orange!15}{60.6} & 1818 & \cellcolor{blue!10}{86.6} & \cellcolor{orange!15}{57.8} & \cellcolor{blue!10}{77.9} & - & \textcolor{gray}{99.0\%} \\
\textbf{MoB} (w/o $\eta$-prior) &  \greenP  \pinkV &
\cellcolor{orange!15}{63.2} & \cellcolor{orange!15}{57.3} & \cellcolor{orange!15}{69.3} & \cellcolor{blue!10}{52.8} & \cellcolor{orange!15}{60.7} & \cellcolor{orange!15}{1842} & 81.7 & 57.5 & 77.2 & \cellcolor{blue!10}{299} & \cellcolor{orange!15}{99.2\%} \\

\quad + $\eta$-prior & - & 
\cellcolor{blue!10}{63.5} & \cellcolor{blue!10}{57.5} & \cellcolor{blue!10}{69.6} & \cellcolor{blue!10}{52.7} & \cellcolor{blue!10}{60.9} & \cellcolor{blue!10}{1845} & 82.1 & \cellcolor{orange!15}{57.8} & \cellcolor{orange!15}{77.5} & \cellcolor{blue!10}{299} & \cellcolor{blue!10}{99.4\%} \\

\midrule

\cellcolor{gray!15}{LLaVA-1.5-7B} & \multicolumn{12}{c}{\cellcolor{gray!15}{\emph{Pruning budget $K=64$}; \emph{Token Reduction Rate} = \textbf{88.9\%}}}\\
ToMe (ICLR'23) & \pinkV & 43.7 & - & 50.0 & - & 48.6 & 1138 & 52.5 & 45.3 & 57.1 & - & \textcolor{gray}{70.1\%} \\
FastV (ECCV'24)  &  \pinkV
& 48.0 & 52.7 & 51.1 & 50.8 & 46.1 & 1256 & 48.0 & 47.8 & 55.0 & 245 & 77.3\% \\
HiRED (AAAI'25) & \pinkV & 60.2 & 51.4 & 68.2 & 50.2 & 54.6 & 1599 & 73.6 & 44.2 & 69.7 & 191 & 87.0\% \\
LLaVA-PruMerge (24.05) & \pinkV & 55.3 & 49.1 & 68.1 & 50.1 & 51.9 & 1549 & 65.3 & 54.0 & 67.4 & 250 & 87.4\% \\
SparseVLM (ICML'25) &\greenP  
& 56.2 & 46.1 & 62.2 & 50.1 & 52.7 & 1505 & 75.1 & 51.8 & 68.2 & 180 & 84.6\% \\
PyramidDrop (CVPR'25) & \greenP & 58.8 & 50.5 & 68.6 & 50.7 & 41.9 & 1561 & 55.9 & 45.9 & 69.2 & 250 & 78.1\% \\
FiCoCo-V (EMNLP'24) & \pinkV & 60.3 & 53.0 & 68.1 & 49.8 & 52.4 & 1591 & 76.0 & 53.6 & 71.3 & - & \textcolor{gray}{91.5\%} \\
MustDrop (24.11)  & \greenP  \pinkV
& 60.0 & 53.1 & 63.4 & 51.2 & 53.1 & 1612 & 68.0 & 54.2 & 69.3 & 267 & 90.1\% \\
VisionZip (24.12) & \pinkV & 60.1 & - & 69.0 & - & 55.1 & 1690 & 77.0 & 55.5 & 72.4 & - & \textcolor{gray}{92.8\%} \\
DART (EMNLP'25)   &   \pinkV
& 60.6 & 53.2 & \cellcolor{orange!15}{69.8} & \cellcolor{orange!15}{51.6} & 55.9 & \cellcolor{orange!15}{1765} & 73.9 & 54.4 & 72.4 & \cellcolor{orange!15}{270} & 93.7\% \\

TokenCarve (25.03) & \greenP \pinkV & \cellcolor{blue!10}{62.0} & - & \cellcolor{orange!15}{69.7} & 51.4 & - & 1754 & \cellcolor{orange!15}{79.9} & \cellcolor{blue!10}{57.0} & 74.8 & - & \textcolor{gray}{97.0\%} \\
TwigVLM (ICCV'25) & \greenP & 60.4 & - & \cellcolor{blue!10}{70.0} & - & \cellcolor{orange!15}{58.8} & 1760 & \cellcolor{blue!10}{82.7} & \cellcolor{orange!15}{55.8} & \cellcolor{blue!10}{75.6} & - & \textcolor{gray}{96.1\%} \\
\textbf{MoB} (w/o $\eta$-prior) &  \greenP  \pinkV &
\cellcolor{orange!15}{61.7} & \cellcolor{orange!15}{54.2} & \cellcolor{orange!15}{69.7} & \cellcolor{blue!10}{52.0} & \cellcolor{blue!10}{59.0} & \cellcolor{blue!10}{1806} & 77.2 & \cellcolor{blue!10}{57.0} & \cellcolor{orange!15}{75.5} & \cellcolor{blue!10}{277} & \cellcolor{orange!10}{96.3\%} \\

\quad + $\eta$-prior &  - & 
\cellcolor{blue!10}{62.1} & \cellcolor{blue!10}{54.5} & \cellcolor{orange!15}{69.8} & \cellcolor{blue!10}{52.1} & \cellcolor{blue!10}{59.0} & \cellcolor{blue!10}{1806} & 77.2 & \cellcolor{blue!10}{57.0} & \cellcolor{orange!15}{75.5} & \cellcolor{blue!10}{277} & \cellcolor{blue!10}{96.4\%} \\

\midrule
\midrule
\cellcolor{gray!15}{LLaVA-Next-7B} & \multicolumn{12}{c}{\cellcolor{gray!15}{\emph{w/o Pruning, $N=2880$}; \emph{Token Reduction Rate} = \textbf{0.0\%}}}\\
\textcolor{gray}{Vanilla}~\cite{liu2024llavanext} & - 
& \textcolor{gray}{67.4} & \textcolor{gray}{60.6} & \textcolor{gray}{70.1} & \textcolor{gray}{57.6} & \textcolor{gray}{64.2} & \textcolor{gray}{1851} & \textcolor{gray}{86.5} & \textcolor{gray}{64.9} & \textcolor{gray}{81.8} & \textcolor{gray}{517} & \textcolor{gray}{100\%} \\

\midrule
\cellcolor{gray!15}{LLaVA-Next-7B}   & \multicolumn{12}{c}{\cellcolor{gray!15}{\emph{Pruning budget $K=320$}; \emph{Token Reduction Rate} = \textbf{88.9\%}}}\\
FastV (ECCV'24) &   \pinkV  
& 61.6 & 51.9 & 62.8 & 53.1 & 55.9 & 1661 & 71.7 & 55.7 & 71.9 & 374 & 86.4\% \\
HiRED (AAAI'25)  & \pinkV & 64.2 & 55.9 & 66.7 & 54.2 & 59.3 & 1690 & 83.3 & 58.8 & 75.7 & 404 & 91.8\% \\
LLaVA-PruMerge (24.05) & \pinkV & 61.3 & 55.3 & 66.4 & 54.0 & 53.6 & 1534 & 60.8 & 50.6 & 69.7 & 146 & 79.9\% \\
SparseVLM (ICML'25)  & \greenP 
& 60.6 & 54.5 & 66.1 & 52.0 & 56.1 & 1533 & 82.4 & 58.4 & 71.5 & 270 & 85.9\% \\
PyramidDrop (CVPR'25) & \greenP & 63.4 & 56.2 & 67.5 & 54.1 & 56.4 & 1663 & 77.6 & 54.4 & 73.5 & 259 & 86.8\% \\
MustDrop (24.11)  & \greenP  \pinkV
& 62.8 & 55.1 & 68.0 & 54.0 & 57.3 & 1641 & 82.1 & \cellcolor{orange!15}{59.9} & 73.7 & 382 & 90.4\% \\
VisionZip (24.12) & \pinkV & 63.1 & - & 67.3 & - & 59.3 & 1702 & - & 58.9 & 76.2 & - & \textcolor{gray}{93.0\%} \\
FasterVLM (24.12)~\cite{zhang2024cls}  & \pinkV
 & 61.6 & 53.5 & 66.5 & 52.6 & 56.9 & 1701 & 83.6 & 56.5 & 74.0 & 401 & 89.8\% \\

GlobalCom$^2$(25.01)~\cite{liu2025compression} & \pinkV & 61.8 & 53.4 & 67.4 & 54.6 & 57.1 & 1698 & 83.8 & 57.2 & 76.7 & 375 & 90.3\% \\
 
DART (EMNLP'25)   &  \pinkV
& \cellcolor{orange!15}{65.3} & \cellcolor{orange!15}{58.2} & \cellcolor{orange!15}{68.4} & \cellcolor{orange!15}{56.1} & 61.7 & 1710 & \cellcolor{orange!15}{84.1} & 58.7 & 79.1 & \cellcolor{orange!15}{406} & \cellcolor{orange!15}{93.9\%} \\

TwigVLM (ICCV'25) & \greenP & 65.0 & - & \cellcolor{blue!10}{68.7} & - & \cellcolor{orange!15}{62.2} & \cellcolor{orange!15}{1758} & - & 57.4 & \cellcolor{orange!15}{79.7} & - & \textcolor{gray}{95.4\%} \\

\textbf{MoB} (with $\eta$-prior)  & \greenP  \pinkV
& \cellcolor{blue!10}{65.8} & \cellcolor{blue!10}{58.9} & \cellcolor{blue!10}{68.7} & \cellcolor{blue!10}{57.0} & \cellcolor{blue!10}{62.6} & \cellcolor{blue!10}{1760} & \cellcolor{blue!10}{84.4}  & \cellcolor{blue!10}{60.2} & \cellcolor{blue!10}{80.1} & \cellcolor{blue!10}{418} & \cellcolor{blue!10}{95.4\%} \\

\bottomrule
\end{tabular}}
\caption{Full results on image understanding with the LLaVA-7B Series. For MoB, we set $K_{\rm p}\in\{64,48,32\}$ and $k\in\{4,6,8\}$, corresponding to token-reduction rates of $\{88.9\%,77.8\%,66.7\%\}$. For MoB with the $\eta$ prior, we use $K_{\rm p}\in\{\tfrac{3K}{8},\tfrac{K}{4},\tfrac{K}{4}\}$ with $k=\tfrac{3K_{\rm p}}{40}$ for strong-coupling benchmarks and $K_{\rm p}\in\{\tfrac{K}{2},\tfrac{7K}{16},\tfrac{5K}{12}\}$ with $k=\tfrac{K_{\rm p}}{8}$ for weak-coupling benchmarks, corresponding to the same token-reduction rates; the pruning layer is fixed at $\ell=2$. {\setlength{\fboxsep}{1pt}
\colorbox{blue!10}{B} and \colorbox{orange!15}{O} denote the best and the second.}
}
\label{tab:5}
\end{table*}

\clearpage
\subsection{Visualization}

\vspace{10mm}
\begin{figure*}[htbp]
	\centering
	\includegraphics[width=0.94\linewidth]{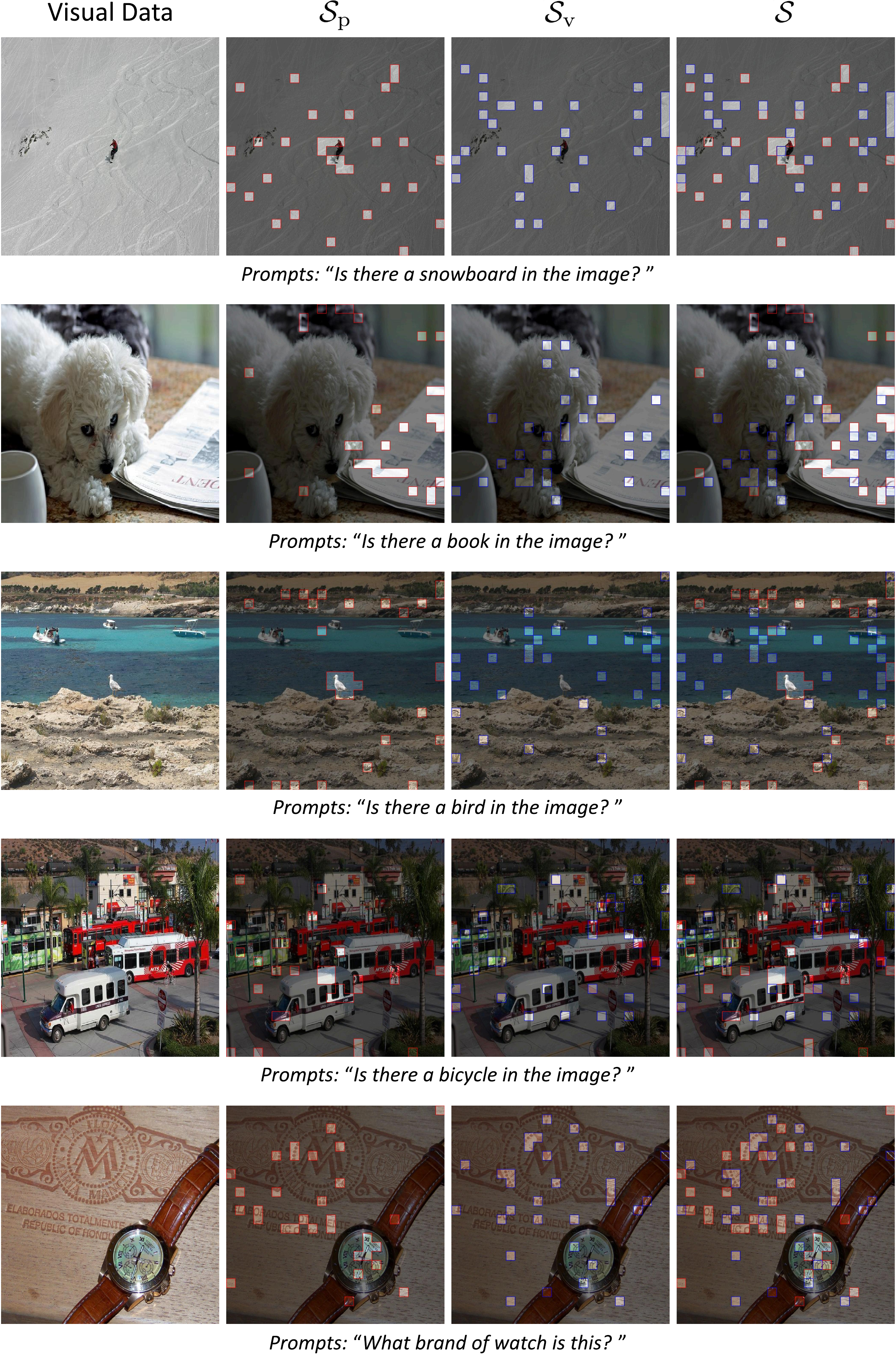}
	\caption{Visualization of the selected prompt and visual centers under weak coupling.}
	\label{fig:7}
\end{figure*}

\vspace{10mm}
\begin{figure*}[htbp]
	\centering
	\includegraphics[width=0.94\linewidth]{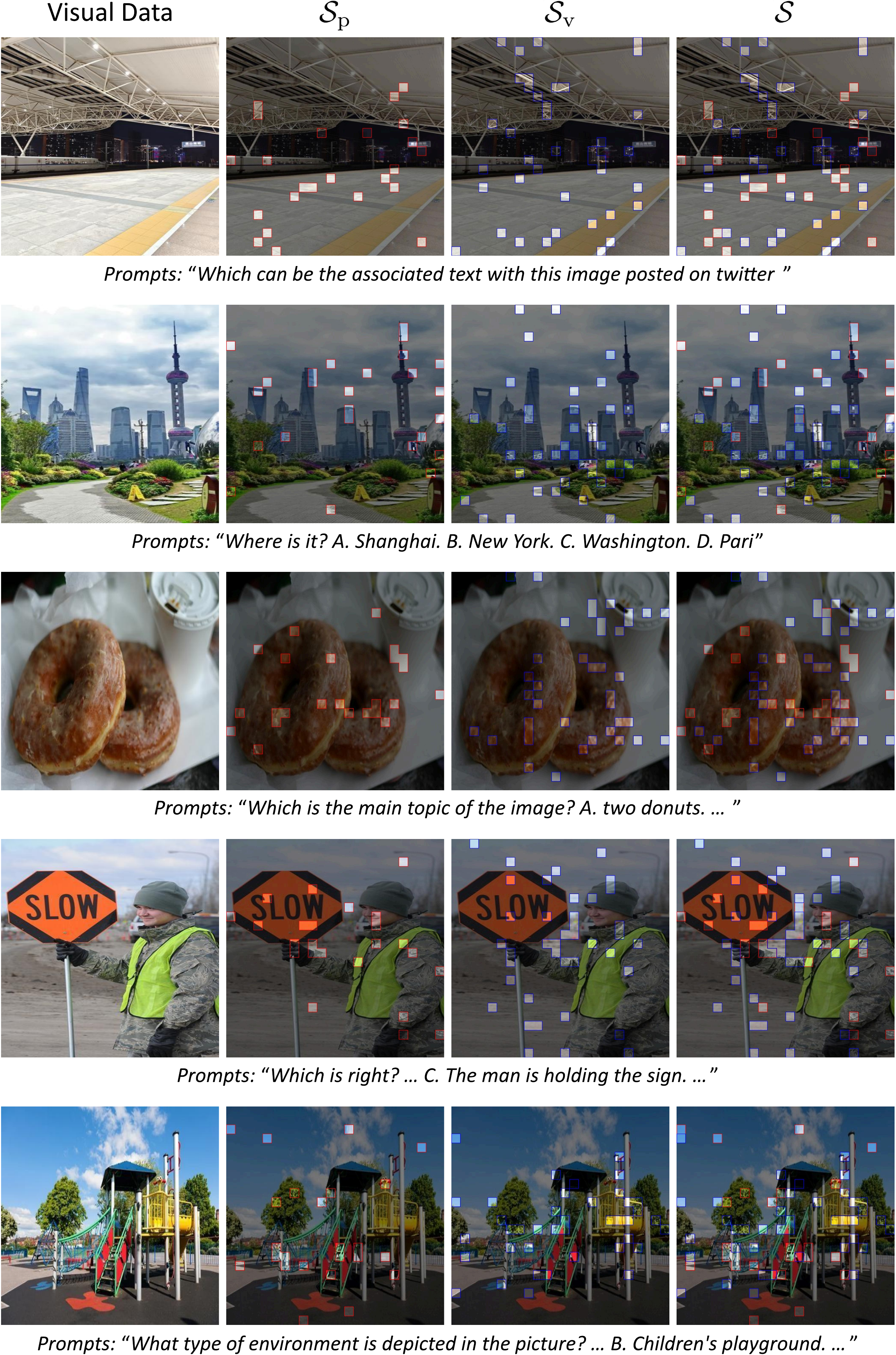}
	\caption{Visualization of the selected prompt and visual centers under strong coupling.}
	\label{fig:8}
\end{figure*}


MoB formulates visual token pruning as a bi-objective covering problem over $(\mathcal{V},\mathcal{P})$, which is expected to gather query-relevant, fine-grained evidence with $\mathcal{S}_{\rm p}$ while preserving global scene context with $\mathcal{S}_{\rm v}$.
The visualizations (\Cref{fig:7,fig:8}) qualitatively validate this design: tokens in $\mathcal{S}_{\rm p}$ concentrate in regions aligned with the text query and key visual evidence, whereas elements of $\mathcal{S}_{\rm v}$ spread more uniformly across the image to maintain the overall context. Together, this complementary allocation enables MoB to retain the most informative visual content for each image–query pair, accounting for its strong empirical performance.

\subsection{Additional Ablation \& Discussion}

\vspace{5mm}
\begin{table}[h]
\centering
\setlength{\tabcolsep}{5pt}
\begin{tabular}{lcccccccccccc}
\toprule
\rowcolor{gray!15}
\multicolumn{13}{c}{GQA}\\
\textbf{$\langle K, K_{\rm p}\rangle$} & $0$ & $2$ & $4$ & $6$ & $8$ & $12$ & $16$ & $24$ & $32$ & $48$ & $64$ & $96$\\
\midrule
$\langle 64,32\rangle$   & 58.3 & \cellcolor{blue!5}{58.8} & \cellcolor{blue!15}{59.0} & \cellcolor{blue!5}{-} & \cellcolor{blue!5}{58.7} & - & 58.2 & - & 57.4 & - & - & - \\
$\langle 128,64\rangle$  & 60.2 & - & \cellcolor{blue!5}{60.5} & \cellcolor{blue!5}{-} & \cellcolor{blue!15}{60.7} & \cellcolor{blue!5}{-} & \cellcolor{blue!5}{60.6} & - & 60.0 & - & 59.5 \\
$\langle 192,96\rangle$  & 60.6 & - & - & \cellcolor{blue!5}{61.1} & \cellcolor{blue!5}{-} & \cellcolor{blue!15}{61.2} & \cellcolor{blue!5}{-} & \cellcolor{blue!5}{60.9} & - & 60.7 & - & 60.5 \\
\midrule
\rowcolor{gray!15}
\multicolumn{13}{c}{TextVQA}\\
\textbf{$\langle K, K_{\rm p}\rangle$} & $0$ & $2$ & $4$ & $6$ & $8$ & $12$ & $16$ & $24$ & $32$ & $48$ & $64$ & $96$ \\
\midrule
$\langle 64,32\rangle$   & 56.5 & \cellcolor{blue!5}{56.9} & \cellcolor{blue!15}{57.0} & \cellcolor{blue!5}{-} & \cellcolor{blue!5}{56.8} & - & 56.5 &- & 56.2 & - & - & - \\
$\langle 128,64\rangle$  & 57.1 & - & \cellcolor{blue!5}{57.5} & \cellcolor{blue!5}{-} & \cellcolor{blue!15}{57.7} & \cellcolor{blue!15}{-}  & \cellcolor{blue!15}{57.7} & - & 57.2 & - & 56.8 & - \\
$\langle 192,96\rangle$  & 57.8 & - & - & \cellcolor{blue!15}{58.2} & \cellcolor{blue!15}{-} & \cellcolor{blue!15}{58.2} & \cellcolor{blue!5}{-} & \cellcolor{blue!5}{58.1} & - & 57.7 & - & 57.5 \\
\bottomrule
\end{tabular}
\vspace{2mm}
\caption{Detailed ablation on the covering fold $k$ for GQA and TextVQA.}
\label{tab:6}
\end{table}

To assess MoB’s sensitivity to the covering-fold parameter $k$---particularly under weak coupling with long prompts---we conduct a detailed ablation on $k$ using GQA and TextVQA.

As~\Cref{tab:6} demonstrates, MoB is not overly sensitive to the choice of $k$, particularly within a clear optimal range. For instance, in both two benchmarks, performance only varies by approximately $0.3\%$ for $k$ values between $[2, 8]$ under $\langle K=64, K_{\rm p}=32\rangle$ setting.

There is also a principled, theoretical reason for this robustness, which stems from the relationship between the covering fold $k$, the budget $K_{\rm p}$, and the length $L$ of prompt tokens $\mathcal{P}$. From covering theory, every prompt token $p\in\mathcal{P}$ is covered by at least one visual token $v\in\mathcal{V}$ under the condition $K_{\rm p}\geq kL$, thereby ensuring the performance guarantee of MoB. Therefore, as selected $k$ satisfies $k\leq K_{\rm p}/L$, the performance will remain stable.

\paragraph{Heuristic for estimating $k$.}
In practice, a robust range for $k$ can be inferred from the prompt length $L$. Given the analysis above, we expect an adaptive, per-sample search for a fine-grained $k$ to yield only limited gains, so we rely on this length-based heuristic instead.

\clearpage

\subsection{Real-life Application}
\vspace{5mm}
\begin{figure*}[h]
	\centering
	\includegraphics[width=1\linewidth]{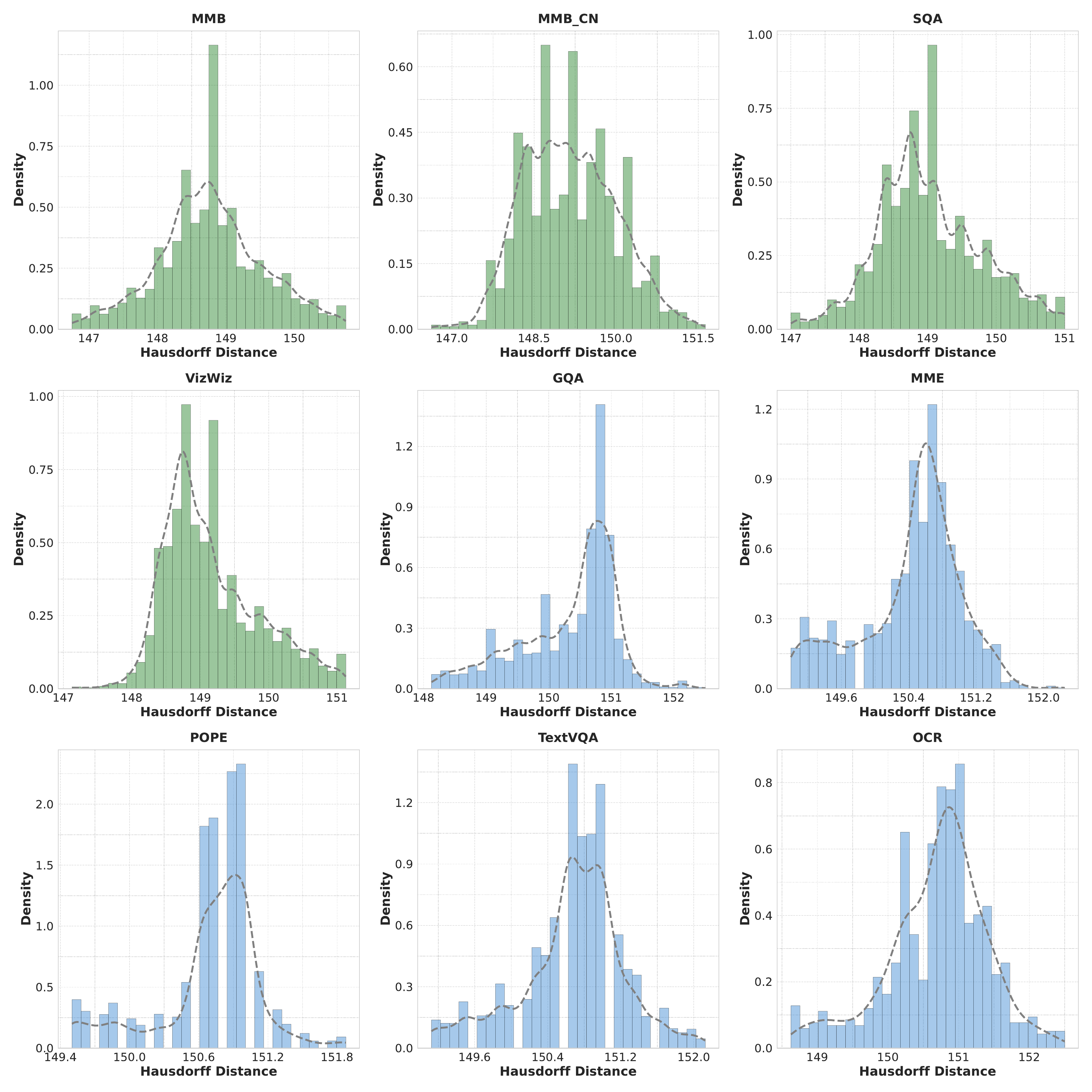}
	\caption{Observations of prompt-visual coupling $\eta$ across 9 popular benchmarks.}
	\label{fig:9}
    \vspace{5mm}
\end{figure*}

\paragraph{Open-domain recipe.}
MoB is task-agnostic, which does not require pre-defined task labels and can operate online by classifying each sample’s coupling pattern. For a given target model, we adopt a two-stage strategy:
\begin{itemize}[leftmargin=12pt]
\item \textbf{Offline calibration.} Analyze the empirical $\eta$ distributions on a set of representative benchmarks (as shown in~\Cref{fig:9}) and set a robust threshold $\tau$ that separates \emph{weak} vs.\ \emph{strong} coupling.
\item \textbf{Online classification and inference.} For each incoming query, compute its Hausdorff distance using~\Cref{alg:hausdorff} with tractable bilinear complexity $\mathcal{O}(N L d)$. Classify the sample by comparing this value to $\tau$, then apply the corresponding budget configuration (\emph{e.g.}, $K_{\rm p}$, $k$) and run MoB + forward inference. In practice, this online cost is negligible relative to the pruned forward pass.
\end{itemize}

\paragraph{Computational Overhead.}
We provide a detailed cost breakdown for online computation of the Hausdorff distance using \Cref{alg:hausdorff} with complexity \(\mathcal{O}(N L d)\) on LLaVA-1.5-7B and LLaVA-Next-7B, where \(N\), \(L\), and \(d\) denote the numbers of visual tokens, prompt tokens, and the feature dimension, respectively. As shown in \Cref{tab:7}, the measured cost (TFLOPs) of exact Hausdorff computation is orders of magnitude smaller than that of MoB itself and the model’s forward pass, yielding a negligible overhead.

Concretely, computing \(d_H\) (\emph{e.g.}, \(\sim 1.2\times 10^{-4}\) TFLOPs on LLaVA-Next) is insignificant relative to the pruned forward pass (\emph{e.g.}, \(\sim 4.6\) TFLOPs at \(K=320\)) and, more importantly, to the savings from pruning (\(\sim 35.9\) TFLOPs). Thus, exact online estimation is not a practical bottleneck; its cost is dwarfed by the efficiency gains of our method. Further acceleration is possible with standard techniques (\emph{e.g.}, heuristic support sampling or low-dimensional random projections), although it is unnecessary in our settings.

\begin{itemize}[leftmargin=12pt]
  \item \textbf{Heuristic Sampling:} It computes the distance on smaller support sets of the tokens \((\mathcal{V}' \subset \mathcal{V},\ \mathcal{P}' \subset \mathcal{P})\), which can be constructed via random sampling~\cite{wen2025stop} or more advanced heuristics such as Key-Norm selection~\cite{akhauri2025tokenbutler,guo2024attention}. This reduces complexity to \(\mathcal{O}(N'L'd)\), where $|\mathcal{V}'|=N',\ |\mathcal{P}'|=L'$.
  \item \textbf{Random Projections:} For a more theoretically grounded approach, the Johnson--Lindenstrauss (JL) lemma~\cite{larsen2014johnson} allows us to project embeddings to a much lower dimension \((d' \ll d)\) while preserving geometric structure, reducing complexity to \(\mathcal{O}(N L\, d')\).
\end{itemize}

\paragraph{Potential Extensions.}
A natural extension is to maintain an online estimate of the coupling statistic \(\eta\) during inference—\emph{e.g.}, a running summary of an approximate \(\hat{\eta}\) computed from shallow-layer tokens. As more samples are processed, we \emph{expect} the empirical distribution of \(\hat{\eta}\) to become bimodal (consistent with the benchmark patterns in~\Cref{fig:9}), enabling a data-driven threshold to be derived on the fly that separates weak vs.\ strong coupling regimes. Using this live threshold, MoB could \emph{adapt} \(K_p\) (and \(k\)) per sample or per mini-batch by selecting from a small budget pool or by scheduling \(K_p\) as a function of \(\hat{\eta}\), with conservative warm-up and safeguards for distribution shift.

\begin{table}[t]
\centering
\setlength{\tabcolsep}{10pt}
\begin{tabular}{lcccc}
\toprule
\rowcolor{gray!15}
\multicolumn{5}{c}{LLaVA-1.5 ($N=576$, $L=10$, $d=4096$)}\\
\textbf{Model} & Vanilla & $K=64$ & $K=128$ & $K=192$ \\
\midrule
Forward        & 8.2              & 1.0             & 1.9              & 2.8 \\
Compute $d_H$  & $2.3{\rm e}-5$ & $2.3{\rm e}-5$ & $2.3{\rm e}-5$ & $2.3{\rm e}-5$ \\
MoB            & --               & $1.7{\rm e}-4$ & $3.3{\rm e}-4$ & $4.8{\rm e}-4$ \\
\midrule
\rowcolor{gray!15}
\multicolumn{5}{c}{LLaVA-Next ($N=2880$, $L=10$, $d=4096$)}\\
\textbf{Model} & Vanilla & $K=320$ & $K=640$ & $K=960$ \\
\midrule
Forward        & 40.5             & 4.6             & 9.1              & 13.6 \\
Compute $d_H$  & $1.2{\rm e}-4$ & $1.2{\rm e}-4$ & $1.2{\rm e}-4$ & $1.2{\rm e}-4$ \\
MoB            & --               & $3.9{\rm e}-3$ & $7.6{\rm e}-3$ & $1.1{\rm e}-2$ \\
\bottomrule
\end{tabular}
\vspace{2mm}
\caption{Computation cost in LLaVA-7B series (TFLOPs)}
\label{tab:7}
\end{table}

\clearpage
\section{Omitted Technical Details}
\label{sec:proof}

\subsection{Proof of~\Cref{lem:1}}
\textbf{Restatement of~\Cref{lem:1}}
(An Error Bound for Visual Token Pruning)\textbf{.} \emph{Under~\Cref{assump:1}, given any token set with its pruned counterpart \(\mathcal{X} = \mathcal{V}\sqcup \mathcal{P},\ \ \mathcal{X}_{\rm s} = \mathcal{S}\sqcup\mathcal{P}\subseteq\mathbb{R}^d\), the pruning error bound is given by:}
\[\textstyle
    \|\mathcal{F}(\mathcal X)-\mathcal{F}(\mathcal X_{\rm s})\| \;\le\; C_\ell\; \max \Big\{
    \min\big\{ d_H(\mathcal{S},\mathcal{V}),\  d_H(\mathcal{V},\mathcal{P}) \big\},\,
    \min\big\{ d_H(\mathcal{S},\mathcal{V}),\  d_H(\mathcal{S},\mathcal{P}) \big\}
    \Big\}.
\]
\begin{remark}
Here $d_H(\mathcal{S},\mathcal{P})$ and $d_H(\mathcal{S},\mathcal{V})$ describe the prompt alignment and visual preservation, while $d_H(\mathcal{V},\mathcal{P})$ is an inherent term that describes the prompt-visual coupling of input data.     
\end{remark}

\begin{proof}
\label{proof:lem1}
The intermediate input for any layer and its pruned counterpart are given by
\[
\mathcal{X} = \mathcal{V} \sqcup \mathcal{P} \text{ and } \mathcal{X}_{\rm s} = \mathcal{S} \sqcup \mathcal{P}.
\]
By~\Cref{Equ:3}, the Hausdorff distance is symmetric, \emph{i.e.}, 
\hypertarget{eq:E1-1}{}
\[
d_H(\mathcal{S},\mathcal{V}) = d_H(\mathcal{V},\mathcal{S}),
\tag*{(E1-1)}
\]
and induced by Euclidean distance.

\medskip
\textbf{Step 1. Bound the one-sided distances.}

We analyze the distances by considering the membership of the points in the subsets.

\textbf{Direction 1} ($\mathcal{X}\rightarrow \mathcal{X}_{\rm s}$) For any \(x\in \mathcal{X}\):

\emph{Case (i):} If \(x\in\mathcal{P}\), then since \(\mathcal{P}\subset \mathcal{X}_{\rm s}\),
\[
\inf_{y\in\mathcal{X}_{\rm s}} \|x-y\| = 0.
\]

\emph{Case (ii):} If \(x\in\mathcal{V}\), then the candidate points in \(\mathcal{X}_{\rm s} = \mathcal{S}\sqcup\mathcal{P}\) can be chosen either from \(\mathcal{S}\) or \(\mathcal{P}\). Thus,
\[
    \inf_{y\in\mathcal{X}_{\rm s}} \|x-y\|
    \;\leq\;
    \min\!\Bigl\{\,
        \inf_{s\in\mathcal{S}}\|x-s\|,
        \;
        \inf_{p\in\mathcal{P}}\|x-p\|
    \Bigr\}.
\]
Taking the supremum over \(x\in\mathcal{V}\) yields
\[
    \sup_{x\in\mathcal{V}}\inf_{y\in\mathcal{X}_{\rm s}}\|x-y\|
    \;\le\;
    \min\!\Bigl\{\,
        \sup_{x\in\mathcal{V}}\inf_{s\in\mathcal{S}}\|x-s\|,
        \;
        \sup_{x\in\mathcal{V}}\inf_{p\in\mathcal{P}}\|x-p\|
    \Bigr\}.
\]

\[
\sup_{x\in\mathcal{V}}\inf_{p\in\mathcal{P}}\|x-p\|
\;\le\;
\max\!\Bigl\{\,
    \sup_{x\in\mathcal{V}}\inf_{p\in\mathcal{P}}\|x-p\|,
    \;
    \sup_{p\in\mathcal{P}}\inf_{x\in\mathcal{V}}\|p-x\|
\Bigr\}
= d_H(\mathcal{V},\mathcal{P}),
\]

By~\Cref{Equ:3}, we derive the distance in direction 1:
\hypertarget{eq:E1-2}{}
\[
    \sup_{x\in\mathcal{X}}\inf_{y\in\mathcal{X}_{\rm s}}\|x-y\|
    \;\le\;
    \min\!\Bigl\{
        d_H(\mathcal{V},\mathcal{S}),
        \;
        d_H(\mathcal{V},\mathcal{P})
    \Bigr\}.
\tag{E1-2}
\]

\textbf{Direction 2} ($\mathcal{X}_{\rm s}\rightarrow \mathcal{X}$) For any \(y\in \mathcal{X}_{\rm s}\):

\emph{Case (i):} If \(y\in\mathcal{P}\), then as \(\mathcal{P}\subset\mathcal{X}\),
\[
\inf_{x\in\mathcal{X}} \|y-x\| = 0.
\]

\emph{Case (ii):} If \(y\in\mathcal{S}\), the candidate points in
\(\mathcal{X}=\mathcal{V}\sqcup\mathcal{P}\) can be chosen from either
\(\mathcal{V}\) or \(\mathcal{P}\); hence
\[
    \inf_{x\in\mathcal{X}} \|y-x\|
    \;\le\;
    \min\!\Bigl\{
        \inf_{v\in\mathcal{V}}\|y-v\|,
        \;
        \inf_{p\in\mathcal{P}}\|y-p\|
    \Bigr\}.
\]
Taking the supremum over \(y\in\mathcal{S}\) yields
\[
    \sup_{y\in\mathcal{S}}\inf_{x\in\mathcal{X}}\|y-x\|
    \;\le\;
    \min\!\Bigl\{
        \sup_{y\in\mathcal{S}}\inf_{v\in\mathcal{V}}\|y-v\|,
        \;
        \sup_{y\in\mathcal{S}}\inf_{p\in\mathcal{P}}\|y-p\|
    \Bigr\}.
\]

\[
\sup_{y\in\mathcal{S}}\inf_{p\in\mathcal{P}}\|y-p\|
\;\le\;
\max\!\Bigl\{
    \sup_{y\in\mathcal{S}}\inf_{p\in\mathcal{P}}\|y-p\|,
    \;
    \sup_{p\in\mathcal{P}}\inf_{y\in\mathcal{S}}\|p-y\|
\Bigr\}
= d_H(\mathcal{S},\mathcal{P}),
\]

By~\Cref{Equ:3}, we derive the distance in direction 2:
\hypertarget{eq:E1-3}{}
\[
    \sup_{y\in\mathcal{X}_{\rm s}}\inf_{x\in\mathcal{X}}\|y-x\|
    \;\le\;
    \min\!\Bigl\{
        d_H(\mathcal{S},\mathcal{V}),
        \;
        d_H(\mathcal{S},\mathcal{P})
    \Bigr\}.
\tag{E1-3}
\]
 
\medskip
\textbf{Step 2. Combine the bounds.}

By~\Cref{Equ:3}, combining the bounds in~\hyperlink{eq:E1-2}{(E1-2)} and~\hyperlink{eq:E1-3}{(E1-3)}, we obtain
\[
    d_H\big(\mathcal{X}, \mathcal{X}_{\rm s}\big)
    \leq \max \Big\{
    \min\big\{ d_H(\mathcal{V},\mathcal{S}),\  d_H(\mathcal{V},\mathcal{P}) \big\},\,
    \min\big\{ d_H(\mathcal{S},\mathcal{V}),\  d_H(\mathcal{S},\mathcal{P}) \big\}
    \Big\}.
\]
Based on~\hyperlink{eq:E1-1}{(E1-1)}, we have
\[
    d_H\big(\mathcal{X}, \mathcal{X}_{\rm s}\big)
    \leq \max \Big\{
    \min\big\{ d_H(\mathcal{S},\mathcal{V}),\  d_H(\mathcal{V},\mathcal{P}) \big\},\,
    \min\big\{ d_H(\mathcal{S},\mathcal{V}),\  d_H(\mathcal{S},\mathcal{P}) \big\}
    \Big\}.
\]

Loading the~\Cref{assump:1}, we have the output discrepancy is bounded by
\[
\begin{aligned}
  \|\mathcal{F}(\mathcal X)-\mathcal{F}(\mathcal X_{\rm s})\|
  \;&\le\;
  C_\ell\;d_H\!\bigl(\mathcal{X}, \mathcal{X}_{\rm s}\bigr).\\
  &= C_\ell\;\max \Big\{
    \min\big\{ d_H(\mathcal{S},\mathcal{V}),\  d_H(\mathcal{V},\mathcal{P}) \big\},\,
    \min\big\{ d_H(\mathcal{S},\mathcal{V}),\  d_H(\mathcal{S},\mathcal{P}) \big\}
    \Big\}.
\end{aligned}
\]
This completes the proof.
\end{proof}

\newpage
\subsection{Proof of~\Cref{lem:1-relax}}
\textbf{Restatement of~\Cref{lem:1-relax}} (A Relaxed Error Bound under Practical Budgets) \textbf{.}
\emph{Under~\Cref{assump:1,assump:2}, let \(\mathcal{X} = \mathcal{V}\sqcup \mathcal{P},\ \ \mathcal{X}_{\rm s} = \mathcal{S}\sqcup\mathcal{P}\subseteq\mathbb{R}^d\) with $|\mathcal{S}|=K\ll N$. Partition the retained token set $\mathcal{S}$ into two disjoint subsets: $\mathcal{S}=\mathcal{S}_{\rm p}\,\sqcup\,\mathcal{S}_{\rm v}$, devoted to prompt alignment $d_H(\mathcal{S}_{\rm p},\mathcal{P})$ and visual preservation $d_H(\mathcal{S}_{\rm v},\mathcal{V})$, respectively. Then, the pruning error bound reduces to}
\[
\textstyle
    \|\mathcal{F}(\mathcal X)-\mathcal{F}(\mathcal X_{\rm s})\| \;\le\; C_\ell\;\max\big\{d_H(\mathcal{S}_{\rm p},\mathcal{P}),\ d_H(\mathcal{S}_{\rm v},\mathcal{V})\big\} + C_\ell\ \eta.
\]

\begin{proof}
\label{proof:lem1-relax}
By~\Cref{lem:1}, we obtain
\[
  \|\mathcal{F}(\mathcal X)-\mathcal{F}(\mathcal X_{\rm s})\|
  \;\le\; C_\ell\;\max \Big\{
    \min\big\{ d_H(\mathcal{S},\mathcal{V}),\  d_H(\mathcal{V},\mathcal{P}) \big\},\,
    \min\big\{ d_H(\mathcal{S},\mathcal{V}),\  d_H(\mathcal{S},\mathcal{P}) \big\}
    \Big\}.
\]
Since $\min\{a,b\} \le \max\{a,b\}$, we have
\hypertarget{eq:E2-1}{}
\[
     \|\mathcal{F}(\mathcal X)-\mathcal{F}(\mathcal X_{\rm s})\|
  \;\le\; C_\ell\; \max\Big\{\ d_H\big(\mathcal{S},\mathcal{P}\big),\ d_H\big(\mathcal{S},\mathcal{V}\big),\ d_H\big(\mathcal{V},\mathcal{P}\big)\Big\}.   
\tag{E2-1}
\]

For any $p\in \mathcal{P}$, we have
\[
\inf_{s\in \mathcal{S}}\|p-s\| = \min\Big\{\inf_{s\in \mathcal{S}_{\rm p}}\|p-s\|,\, \inf_{s\in \mathcal{S}_{\rm v}}\|p-s\|\Big\} \le \inf_{s\in \mathcal{S}_{\rm p}}\|p-s\|.
\]
Taking the supremum over $p\in \mathcal{P}$ yields
\[
\sup_{p\in \mathcal{P}} \inf_{s\in \mathcal{S}} \|p-s\| \le \sup_{p\in \mathcal{P}} \inf_{s\in \mathcal{S}_{\rm p}} \|p-s\|.
\]
Similarly, since $\mathcal{S}_{\rm v}\subset \mathcal{S}$,
\[
\sup_{s\in \mathcal{S}_{\rm v}} \inf_{p\in \mathcal{P}} \|s-p\| \le \sup_{s\in \mathcal{S}} \inf_{p\in \mathcal{P}} \|s-p\|.
\]
Thus, by~\Cref{Equ:3},
\[
d_H(\mathcal{S},\mathcal{P}) \le \max\Big\{ d_H(\mathcal{S}_{\rm p},\mathcal{P}),\; d_H(\mathcal{S}_{\rm v},\mathcal{P}) \Big\}.
\]

Using~\Cref{assump:2} \((d_H(\mathcal V,\mathcal P)\le\eta)\) and the triangle
inequality for Hausdorff distance, we have
\[
  d_H(\mathcal S_{\rm v},\mathcal P)
  \le
  d_H(\mathcal S_{\rm v},\mathcal V)+d_H(\mathcal V,\mathcal P)
  \le
  d_H(\mathcal S_{\rm v},\mathcal V)+\eta,
\]
\[
  d_H(\mathcal S_{\rm p},\mathcal V)
  \le
  d_H(\mathcal S_{\rm p},\mathcal P)+d_H(\mathcal P,\mathcal V)
  \le
  d_H(\mathcal S_{\rm p},\mathcal P)+\eta.
\]
Hence,
\hypertarget{eq:E2-2}{}
\[
d_H(\mathcal{S},\mathcal{P}) \le \max\Big\{ d_H(\mathcal{S}_{\rm p},\mathcal{P}),\; d_H(\mathcal{S}_{\rm v},\mathcal{V})+\eta \Big\}.
\tag{E2-2}
\]
Similarly, one can show that
\hypertarget{eq:E2-3}{}
\[
d_H(\mathcal{S},\mathcal{V}) \le \max\Big\{ d_H(\mathcal{S}_{\rm v},\mathcal{V}),\; d_H(\mathcal{S}_{\rm p},\mathcal{P})+\eta \Big\}.
\tag{E2-3}
\]

Loading the maximum of~\hyperlink{eq:E2-2}{(E2-2)},~\hyperlink{eq:E2-3}{(E2-3)} and $d_H(\mathcal{V},\mathcal{P})$ into~\hyperlink{eq:E2-1}{(E2-1)}, we obtain
\[
\begin{aligned}
  \|\mathcal{F}(\mathcal X)-\mathcal{F}(\mathcal X_{\rm s})\|
  \;&\le
C_\ell\;\max\Big\{d_H(\mathcal{S},\mathcal{P}),\, d_H(\mathcal{S},\mathcal{V}),\ d_H(\mathcal{V},\mathcal{P})\Big\}\\
&\le C_\ell\;\max\Big\{ d_H(\mathcal{S}_{\rm p},\mathcal{P}),\; d_H(\mathcal{S}_{\rm v},\mathcal{V})+\eta,\ \ d_H(\mathcal{S}_{\rm v},\mathcal{V}),\; d_H(\mathcal{S}_{\rm p},\mathcal{P})+\eta, \ \ \eta \Big\} \\
\end{aligned}
\]
Since \(d_H(\mathcal{S}_{\rm p},\mathcal{P})\ge 0\), \( d_H(\mathcal{S}_{\rm v},\mathcal{V})\ge 0\), \(\eta\ge 0\), we have
\[
\begin{aligned}
&\max\bigl\{d_H(\mathcal{S}_{\rm p},\mathcal{P}),\; d_H(\mathcal{S}_{\rm p},\mathcal{P})+\eta,\; \eta\bigr\} = d_H(\mathcal{S}_{\rm p},\mathcal{P})+\eta, \\
&\max\bigl\{d_H(\mathcal{S}_{\rm v},\mathcal{V}),\; d_H(\mathcal{S}_{\rm v},\mathcal{V})+\eta,\; \eta\bigr\} = d_H(\mathcal{S}_{\rm v},\mathcal{V})+\eta. 
\end{aligned}
\]
Hence
\[
\|\mathcal{F}(\mathcal X)-\mathcal{F}(\mathcal X_{\rm s})\|
  \;\le C_\ell\;\max\Big\{ d_H(\mathcal{S}_{\rm p},\mathcal{P}),\; d_H(\mathcal{S}_{\rm v},\mathcal{V}) \Big\} + C_\ell\ \eta.
\]
This completes the proof.
\end{proof}

\newpage
\subsection{Proof of~\Cref{lem:2}}
\textbf{Restatement of Lemma~\ref{lem:2}} ($d_{\mathrm{eff}}$-regular lower bound on covering numbers)\textbf{.}
\emph{Given $\mathcal{P},\mathcal{V}\subset\mathbb{R}^d$ with an
effective dimension $d_{\mathrm{eff}}$.  Suppose their $\delta$-dilations \smash{$\mathcal{V}_\delta:=\bigcup_{v\in\mathcal{V}}B(v,\delta)$},
\smash{$\mathcal{P}_\delta:=\bigcup_{p\in\mathcal{P}}B(p,\delta)$} ($\delta\ll\eta$) satisfy $d_{\mathrm{eff}}$-dimensional covering regularity; thus, there exist constants $b\!>\!a\!>\!0$, $b'\!>\!a'\!>0$ and $\epsilon_0\!>\!\delta$ such that
\[
  a\,\epsilon_{\rm p}^{-d_{\mathrm{eff}}}
  \le\mathcal N(\mathcal P,\epsilon_{\rm p})\le b\,\epsilon_{\rm p}^{-d_{\mathrm{eff}}},
  \qquad
  a'\,\epsilon_{\rm v}^{-d_{\mathrm{eff}}}
  \le\mathcal N(\mathcal V,\epsilon_{\rm v})\le b'\,\epsilon_{\rm v}^{-d_{\mathrm{eff}}},
  \qquad
  \forall\,\epsilon_{\rm p},\epsilon_{\rm v}\in(\delta,\epsilon_0],
\]
\emph{\textbf{Remark}} Previous work suggests that both visual and language embeddings concentrate on a low-dimensional manifold, so the effective covering dimension satisfies the typical relation $d_{\mathrm{eff}}\ll d$.}

\begin{proof}
\label{proof:lem2}
We prove the two-sided bound for \(\mathcal P\); the argument for \(\mathcal V\) is identical.

\paragraph{Notation.}
\begin{itemize}\setlength\itemsep{0pt}
  \item \(\mathcal N(X,r)\): minimal number of closed balls of radius \(r\) covering \(X\).
  \item \(X_\delta = \bigcup_{x\in X} B(x,\delta)\), with \(B(x,\delta)=\{y:\|y-x\|\le\delta\}\).
\end{itemize}

\medskip
\textbf{Step 1. Transfer trick for small \(\epsilon\).}

Fix \(\epsilon\in(\delta,\epsilon_0]\) and define
\(\epsilon'=\min\{\epsilon+\delta,\;\epsilon_0\}\).

\noindent If \(\epsilon\le\epsilon_0-\delta\) (so
\(\epsilon'=\epsilon+\delta\)), then any \(\epsilon\)-cover
\(\{z_i\}_{i=1}^m\) of \(\mathcal P\) satisfies for each
\(y\in\mathcal P_\delta\):
\[
  \exists\,x\in\mathcal P:\|y-x\|\le\delta,
  \quad
  \exists\,i:\|x-z_i\|\le\epsilon
  \;\Longrightarrow\;
  \|y-z_i\|\le\epsilon+\delta=\epsilon'.
\]
Hence
\hypertarget{eq:E3-1}{}
\[
  \mathcal P_\delta\;\subseteq\;\bigcup_{i=1}^mB(z_i,\epsilon')
  \quad\Longrightarrow\quad
  \mathcal N(\mathcal P_\delta,\epsilon')\;\le\;\mathcal N(\mathcal P,\epsilon).
\tag{E3-1}
\]
\emph{Note:} For $\epsilon > \epsilon_0 - \delta$, the above transfer argument is not applied.

\medskip
\textbf{Step 2. Lower bound on \(\mathcal N(\mathcal P,\epsilon)\).}

Split into two cases:

\medskip
\noindent\textbf{Case I: \(\epsilon\le\epsilon_0-\delta\).}
Since \(\mathcal{P}_\delta\) satisfies $d_{\mathrm{eff}}$-dimensional covering regularity; loading the lower-bound for \(\mathcal P_\delta\) at
radius \(\epsilon'=\epsilon+\delta\), there exists a constant $a_\delta\ge 0$ such that
\[
  \mathcal N(\mathcal P_\delta,\epsilon')
  =\mathcal N(\mathcal P_\delta,\epsilon+\delta)
  \;\ge\;a_\delta\,(\epsilon+\delta)^{-d_{\mathrm{eff}}}.
\]
Based on~\hyperlink{eq:E3-1}{(E3-1)}, we obtain
\[
a_\delta(\epsilon+\delta)^{-d_{\mathrm{eff}}}\;\le\; 
\mathcal N(\mathcal P_\delta,\epsilon')\;\le\;
\mathcal N(\mathcal P,\epsilon)
\]
Since \(\delta\le \epsilon\), it follows that \(\epsilon+\delta\le2\epsilon\); thus, we have
\hypertarget{eq:E3-2}{}
\[
  \mathcal N(\mathcal P,\epsilon)
  \;\ge\;
  a_\delta\,2^{-d_{\mathrm{eff}}}\,\epsilon^{-d_{\mathrm{eff}}}.
\tag{E3-2}
\]

\medskip
\noindent\textbf{Case II: $\epsilon>\epsilon_0-\delta$.}
Define
\(
\widetilde{a} \;:=\; (\epsilon_0 - \delta)^{d_{\mathrm{eff}}},
\)
such that
\[
(\epsilon_0 - \delta)^{-d_{\mathrm{eff}}}
= \widetilde{a}^{-1}.
\]
Since 
\(
\epsilon > \epsilon_0 - \delta,
\)
we have
\[
\epsilon^{-d_{\mathrm{eff}}}
\;\le\;
(\epsilon_0 - \delta)^{-d_{\mathrm{eff}}}.
\]
Hence
\[
\epsilon^{-d_{\mathrm{eff}}}
\;\le\;
\widetilde{a}^{-1}
\quad\Longleftrightarrow\quad
\widetilde{a}\,\epsilon^{-d_{\mathrm{eff}}}\;\le\;1.
\]
Since any nonempty set $\mathcal{P}$ has covering number at least one, the following holds
\hypertarget{eq:E3-3}{}
\[
  \widetilde a\,\epsilon^{-d_{\mathrm{eff}}}\;\le\;1\;\le\;
  \mathcal N(\mathcal P,\epsilon).
\tag{E3-3}
\]
Therefore, set
\(\displaystyle a:=\min\{a_\delta2^{-d_{\mathrm{eff}}},\,\widetilde a\}>0\), combining~\hyperlink{eq:E3-2}{(E3-2)} and~\hyperlink{eq:E3-3}{(E3-3)} yields
\hypertarget{eq:E3-4}{}
\[
    \mathcal N(\mathcal P,\epsilon)
    \;\ge\;
    a\,\epsilon^{-d_{\mathrm{eff}}},
  \quad\forall\,\epsilon\in(\delta,\epsilon_0].
\tag{E3-4}
\]
Similarly, \(\mathcal V\) holds
\(\mathcal N(\mathcal V,\epsilon)\ge a'\,\epsilon^{-d_{\mathrm{eff}}},\quad\forall\,\epsilon\in(\delta,\epsilon_0].\)

\medskip
\textbf{Step 3. Upper bound on \(\mathcal N(\mathcal P,\epsilon)\).}

Since \(\mathcal{P}_\delta\) satisfies $d_{\mathrm{eff}}$-dimensional covering regularity, there exists a constant $b_\delta\ge a_\delta\ge0$ such that
\[
  \mathcal N(\mathcal P_\delta,\epsilon)
  \;\le\;
  b_\delta\,\epsilon^{-d_{\mathrm{eff}}}.
\]
Since \(\mathcal P\subseteq\mathcal P_\delta\), we have
\(\mathcal N(\mathcal P,\epsilon)\le\mathcal N(\mathcal P_\delta,\epsilon)\); thus, the following holds
\[
\mathcal N(\mathcal P,\epsilon)\le\mathcal N(\mathcal P_\delta,\epsilon)\le\;
  b_\delta\,\epsilon^{-d_{\mathrm{eff}}}.
\]
Based on the \emph{monotonicity of covering numbers}, for every radius \(\epsilon\ge\delta\), we have
\[
\mathcal N(\mathcal P,\epsilon)\le\mathcal N(\mathcal P,\delta).
\]
Therefore, set
\(
  b \;:=\;\max\{b_\delta,\; \mathcal N(\mathcal P,\delta) \},
\)
for all \(\epsilon\in(\delta,\epsilon_0]\) we have
\hypertarget{eq:E3-5}{}
\[
    \mathcal N(\mathcal P,\epsilon)
    \;\le\;
    b\,\epsilon^{-d_{\mathrm{eff}}}.
\tag{E3-5}
\]
Likewise for \(\mathcal V\), the following holds
\(\mathcal N(\mathcal V,\epsilon)\le b'\,\epsilon^{-d_{\mathrm{eff}}},\quad\forall\,\epsilon\in(\delta,\epsilon_0].\)

\textbf{Step 4. Combine the bounds.}

Based on~\hyperlink{eq:E3-4}{(E3-4)} and~\hyperlink{eq:E3-5}{(E3-5)}, for all \(\epsilon\in(\delta,\epsilon_0]\) the following holds
\[
  a\,\epsilon^{-d_{\mathrm{eff}}}\le\mathcal N(\mathcal P,\epsilon)\le b\,\epsilon^{-d_{\mathrm{eff}}},
\quad
  a'\,\epsilon^{-d_{\mathrm{eff}}}\le\mathcal N(\mathcal V,\epsilon)\le b'\,\epsilon^{-d_{\mathrm{eff}}}.
\]
This completes the proof.
\end{proof}

\newpage
\subsection{Proof of~\Cref{theorem:1}}

\textbf{Restatement of~\Cref{theorem:1}} (Trade-off between Prompt Alignment and Visual Preservation)\textbf{.}
\emph{
Under~\Cref{assump:2} and the covering-regularity hypothesis of~\Cref{lem:2} with constants $a,a',d_{\mathrm{eff}}>0$, there exist a radius-scaling factor $z>1$ such that $\eta/z>\delta$ and \(K<\mathcal N(\mathcal P,\eta/z)+\mathcal N(\mathcal V,\eta/z)\), for every pruning results $\mathcal S=(\mathcal S_{\rm p}\sqcup\mathcal S_{\rm v})\subseteq\mathcal{V}$ with budget $K$ satisfying
\[
\max\bigl\{D_1K^{-2/d_{\mathrm{eff}}},\;D_2\,\eta^{2}\bigr\}\;\le\;
d_H(\mathcal S_{\rm p},\mathcal P)\;
d_H(\mathcal S_{\rm v},\mathcal V),
\]
where $D_1 \coloneqq(4\,a\,a')^{1/d_{\mathrm{eff}}}>0$, $D_2\coloneqq 1/z^2>0$.}
\begin{remark}[Optimal Attainment Level] 
The term \(D_1\,K^{-2/d_{\mathrm{eff}}}\) is completely determined by the pruning budget, while \(D_2\,\eta^2\) quantifies the effect of prompt-visual coupling. Hence, the optimal attainment level per objective is given by \(\epsilon^* =\max\{\eta/z,\;\sqrt{D_1}\,K^{-1/d_{\mathrm{eff}}}\}.\)
Any attempt to reduce one objective below \(\epsilon^*\) forces the other above \(\epsilon^*\), thereby increasing the overall pruning error.
\end{remark}
\begin{remark}[Effect of Budget and Coupling Strength] 
As $K$ decreases, $z$ correspondingly shrinks ($D_{2}$ growing as a power function), ultimately making \(D_2\,\eta^2\) dominate the bound; while as \(K\) increases, both of the terms reduce, thereby diminishing the trade‐off and tightening the overall error bound.
\end{remark}


\begin{proof}
\label{proof:theorem1} We begin the proof by noting
\[
  \epsilon_{\rm p}=d_H(\mathcal S_{\rm p},\mathcal P),\quad
  \epsilon_{\rm v}=d_H(\mathcal S_{\rm v},\mathcal V),\quad
  K_{\rm p}=|\mathcal S_{\rm p}|,\quad
  K_{\rm v}=|\mathcal S_{\rm v}|,\quad
  K_{\rm p}+K_{\rm v}=K.
\]

\medskip
\noindent\textbf{Step 1. Quantify the impact of budget \(K\).}

By Lemma~\ref{lem:2}, for all \(\epsilon_{\rm p},\epsilon_{\rm v}\in(\delta,\epsilon_0]\), we have
\hypertarget{eq:E4-1}{}
\[
  a\,\epsilon_{\rm p}^{-d_{\mathrm{eff}}}\le\mathcal N(\mathcal P,\epsilon_{\rm p})\le K_{\rm p},
  \quad
  a'\,\epsilon_{\rm v}^{-d_{\mathrm{eff}}}\le\mathcal N(\mathcal V,\epsilon_{\rm v})\le K_{\rm v}.
 \tag{E4-1}
\]
By AM-GM inequality, we have
\(
K_{\rm p}K_{\rm v}\le\bigl(\tfrac{K}{2}\bigr)^2;
\)
thus, loading~\hyperlink{eq:E4-1}{(E4-1)} we have
\[
  (a\,a')\,(\epsilon_{\rm p}\,\epsilon_{\rm v})^{-d_{\mathrm{eff}}}
  \;\le\;
  \Bigl(\tfrac{K}{2}\Bigr)^{2} \implies   \epsilon_{\rm p}\,\epsilon_{\rm v}
  \;\ge\;
  (4\,a\,a')^{1/d_{\mathrm{eff}}}\,
  K^{-2/d_{\mathrm{eff}}}.
\]
Define
\(
  D_1:=(4\,a\,a')^{1/d_{\mathrm{eff}}}>0,
\)
the \(K\)-bound is established by
\hypertarget{eq:E4-2}{}
\[
  \epsilon_{\rm p}\,\epsilon_{\rm v}\;\ge\;D_1\,K^{-2/d_{\mathrm{eff}}}.
\tag{E4-2}
\]

\medskip
\noindent\textbf{Step 2. Quantify the impact of prompt-visual coupling $\eta$.}

Based on the budget condition, the radius-scaling factor $z$ holds
\hypertarget{eq:E4-3}{}
\[
  K<\mathcal N\bigl(\mathcal P,\tfrac{\eta}{z}\bigr)
  +\mathcal N\bigl(\mathcal V,\tfrac{\eta}{z}\bigr).
\tag{E4-3}
\]
For contradiction, we suppose two covering radii is simultaneously small, such that
\(\epsilon_{\rm p}<\eta/z\) and \(\epsilon_{\rm v}<\eta/z\). Then, the monotonicity of covering numbers gives
\[
  \mathcal N(\mathcal P,\epsilon_{\rm p})\ge\mathcal N\bigl(\mathcal P,\tfrac{\eta}{z}\bigr),
  \quad
  \mathcal N(\mathcal V,\epsilon_{\rm v})\ge\mathcal N\bigl(\mathcal V,\tfrac{\eta}{z}\bigr).
\]
Hence
\[
  K\;\ge\;
  \mathcal N(\mathcal P,\epsilon_{\rm p})
  +\mathcal N(\mathcal V,\epsilon_{\rm v})
  \;\ge\;
  \mathcal N\bigl(\mathcal P,\tfrac{\eta}{z}\bigr)
  +\mathcal N\bigl(\mathcal V,\tfrac{\eta}{z}\bigr),
\]
contradicting~\hyperlink{eq:E4-3}{(E4-3)}.  Therefore \emph{at least one} of
\(\epsilon_{\rm p},\epsilon_{\rm v}\) is \(\ge\eta/z\).  Consequently
\[
  \epsilon_{\rm p}\,\epsilon_{\rm v}
  \;\ge\;
  \Bigl(\tfrac{\eta}{z}\Bigr)^{2},
\]
Define 
\(
  D_2:=\frac{1}{z^{2}}>0,
\)
the $\eta$-bound is given by
\hypertarget{eq:E4-4}{}
\[
  \epsilon_{\rm p}\,\epsilon_{\rm v}
  \;\ge\;
  D_2\,\eta^{2}.
\tag{E4-4}
\]

\medskip
\noindent\textbf{Step 3. Combine the impacts.}  

By~\hyperlink{eq:E4-2}{(E4-2)} and~\hyperlink{eq:E4-4}{(E4-4)}, we have
\[
\epsilon_{\rm p}\epsilon_{\rm v}\ge D_1K^{-2/d_{\mathrm{eff}}}\quad\text{and}\quad \epsilon_{\rm p}\epsilon_{\rm v}\ge D_2\eta^{2} \implies
  \epsilon_{\rm p}\,\epsilon_{\rm v}
  \;\ge\;
  \max\bigl\{D_1K^{-2/d_{\mathrm{eff}}},\,D_2\eta^{2}\bigr\}.
\]
This completes the proof.
\end{proof}

\subsection{Proof of~\Cref{theorem:2}}
\textbf{Restatement of Theorem~\ref{theorem:2}} (Performance Guarantee)\textbf{.}
\emph{Under~\Cref{assump:1} and the covering-regularity of~\Cref{lem:2} with constants $a,a',d_{\mathrm{eff}}\!>\!0$ and $b\!>\!a,\,b'\!>\!a'$, for any budget split $(K_{\rm p},\,K-K_{\rm p})$, covering fold $k$, and token set $\mathcal{X}=\mathcal{V}\sqcup\mathcal{P}\subseteq\mathbb{R}^d$ with $|\mathcal{V}|=N$, $|\mathcal{P}|=L$, and $d_H(\mathcal{V},\mathcal{P})\le\eta$, the following hold:
\begin{enumerate}[label=(\alph*),leftmargin=16pt,nosep]
\item \textbf{Performance bound:} The Performance degradation caused by MoB is upper bounded by
\[\textstyle
\|\mathcal{F}(\mathcal{X})-\mathcal{F}(\mathrm{MoB}(\mathcal{X}))\|\;\le\;C_\ell\;  \max\Bigl\{
     \alpha(\eta,k,L)\,(K_{\rm p})^{-1/d_{\mathrm{eff}}}, 
     \;\beta\,(K-K_{\rm p})^{-1/d_{\mathrm{eff}}}
 \Bigr\}
 \;+\;C_\ell\, \eta
 ,\]
where
\(
\alpha(\eta,k,L)\;=\; \eta\,\bigl(b\,k\,L/a\bigr)^{1/d_{\mathrm{eff}}},\quad
\beta\;=\;2 (b')^{1/d_{\mathrm{eff}}}.
\)\vspace{0.6mm}
\item \textbf{Multilinear complexity:} The complexity of MoB is given by
\(\textstyle
T_{\text{\rm MoB}}=\mathcal{O}(N\,(L+K)\,d)
.\)
\end{enumerate}}
\begin{remark}[Coupling Trade-off]
Under weak coupling (large $\alpha(\eta,k,L)$), minimizing the bound requires a larger $K_{\rm p}$. Conversely, under strong coupling (small $\alpha(\eta,k,L)$), the alignment term decays rapidly, favoring visual preservation (increasing $K-K_{\rm p}$). Specially, under perfect coupling ($\eta=0$), the bound simplifies to \scalebox{0.85}{\(\|\Delta y\|\;\le\;C_\ell\,\beta\,(K-K_{\rm p})^{-1/d_{\mathrm{eff}}}\)},\emph{i.e.}, MoB reduces to pure visual preservation.
\end{remark}
\begin{remark}[Budget Scaling]
As the total budget $K$ increases, the preservation term
\scalebox{0.85}{\(\beta\,(K - K_{\rm p})^{-1/d_{\mathrm{eff}}}\)}
decays, requiring a corresponding increase in $K_{\rm p}$ (and thus a reduction in the alignment term) to rebalance the trade‐off and further lower the overall error bound.
\end{remark}
\begin{remark}[Scalability]
MoB exhibits a multilinear scalability with respect to visual tokens $N$, prompt tokens $L$, and retained tokens $K$ (especially $K, L \ll N$), making it readily adaptable to more challenging scenarios, such as advanced MLLMs with higher‐resolution inputs or multi‐frame video.
\end{remark}

\paragraph{Notation.} 
\begin{enumerate}[label=$\bullet$,leftmargin=16pt]
\item The intermediate input $\mathcal{X}$ is formulated as  
\[
 \mathcal{X}=\mathcal{V}\,\sqcup\, \mathcal{P}\ \subseteq \mathbb{R}^d
 \quad \text{where}\quad 
 |\mathcal{V}|=N,\ \ |\mathcal{P}|=L,\ \ \text{and}\ \ N\gg L.
\]
Particularly, $\mathcal{V}$, $\mathcal{P}$ are compact sets with $d_{\mathrm{eff}}$ effective dimensions.
\item We define the pruned intermediate input as 
\[
\mathrm{MoB}(\mathcal{X})\coloneqq\mathcal{X}_{\rm s}, \quad \text{where}\quad \mathcal{X}_{\rm s}=\mathcal{S}\,\sqcup\, \mathcal{P}\quad \text{where}\quad |\mathcal{S}|=K.
\]
\item The budget configuration is given by $\langle K_{\rm p}, K_{\rm v}\rangle$, where $K_{\rm p}+K_{\rm v}=K$.

\end{enumerate}

\begin{proof}
\label{proof:theorem2}
We separately proof the Performance Guarantee \& Complexity in Part A \& Part B

\paragraph{Part A: Performance Guarantee}
\paragraph{Part A-1: Performance Guarantee of prompt alignment} \leavevmode \\[2ex]
\noindent\emph{Step A-1.1: Bound of the radius derived by $k$-fold NN-covering} 
\vspace{0.5mm}

Given any union set before $K_{\rm p}$-truncation 
\[
\mathcal{S}_{\rm p}'\coloneqq \bigcup\limits_{p\in\mathcal{P}} \operatorname*{arg\,top-k}_{s_{\rm p}\in\mathcal{V}}(\cos(s_{\rm p},p), k) 
\quad \text{where}\quad
|\mathcal{S}_{\rm p}'| = K_{\rm p}'
  \quad\text{and}\quad
  K_{\rm p}\le K_{\rm p}'\le kL,
\]
we define
\[
  \epsilon_{\rm p}'\;=\;d_H\bigl(\mathcal{S}_{\rm p}',\,\mathcal{P}\bigr).
\]
By previous work~\cite{hochbaum1985best}, NN-covering achieves a \(1\)-approximation for the \(k\)-center problem with sufficient budget; \emph{i.e.}, specifically for any $p\in\mathcal{P}$ we have
\[
\inf_{s_{\rm p}'\in\mathcal{S}_{\rm p}'}\|p-s_{\rm p}'\| = \inf_{v\in\mathcal{V}}\|p-v \|.
\]
Thus, 
\[
\sup_{p\in\mathcal{P}}\inf_{s_{\rm p}'\in\mathcal{S}_{\rm p}'}\|p-s_{\rm p}'\|= \sup_{p\in\mathcal{P}}\inf_{v\in\mathcal{V}}\|p-v \|.
\]
Based on~\Cref{assump:2}, since $s\in\mathcal{S}_{\rm p}'\subseteq\mathcal{V}$, the upper bound of the radius $\epsilon_{\rm p}'$ is given by
\hypertarget{eq:E5-1}{}
\[
\begin{aligned}
 \epsilon_{\rm p}' = d_H\bigl(\mathcal{S}_{\rm p}',\,\mathcal{P}\bigr)&\coloneqq\max\{\sup_{s_{\rm p}'\in\mathcal{S}_{\rm p}'}\inf_{p\in\mathcal{P}}\|p-s_{\rm p}'\|
 ,\ \ \sup_{p\in\mathcal{P}}\inf_{s_{\rm p}'\in\mathcal{S}_{\rm p}'}\|p-s_{\rm p}'\|\}\\
 & \le \max\{\sup_{v\in\mathcal{V}}\inf_{p\in\mathcal{P}}\|p-v\|
 ,\ \ \sup_{p\in\mathcal{P}}\inf_{v\in\mathcal{V}}\|p-v\|\} \\
 &\coloneqq d_H\bigl(\mathcal{V},\,\mathcal{P}\bigr)\le\eta.
\end{aligned}
  \tag{E5-1}    
\]

\noindent\emph{Step A-1.2: Impact of $K_{\rm p}$-truncation on the radius}
\vspace{0.5mm}

Based on~\Cref{lem:2}, we have
\[
ar^{-d_{\mathrm{eff}}}\le\mathcal{N}(\mathcal{P}, r) \leq b\,r^{-d_{\mathrm{eff}}}.
\]
In particular:
\[
  b\,(\epsilon_{\rm p})^{-d_{\mathrm{eff}}}
  \;\ge\;K_{\rm p}
  \quad\Longrightarrow\quad
  \epsilon_{\rm p}
  \;\le\;
  \Bigl(\tfrac{b}{K_{\rm p}}\Bigr)^{1/d_{\mathrm{eff}}}.
\]
and also
\[
  a\,(\epsilon_{\rm p}')^{-d_{\mathrm{eff}}}
  \;\le\;K_{\rm p}'
  \quad\Longrightarrow\quad
  \epsilon_{\rm p}'
  \;\ge\;
  \Bigl(\tfrac{a}{K_{\rm p}'}\Bigr)^{1/d_{\mathrm{eff}}}.
\]
Combining the upper and lower bound for $\epsilon_{\rm p}$ and $\epsilon_{\rm p}'$, respectively in terms of $b,K_{\rm p},K_{\rm p}'$, we obtain
\[
  \epsilon_{\rm p} 
  \;\le\;
  \Bigl(\tfrac{b}{K_{\rm p}}\Bigr)^{1/d_{\mathrm{eff}}}
  \;=\;
  \Bigl(\tfrac{bK_{\rm p}'}{aK_{\rm p}}\Bigr)^{1/d_{\mathrm{eff}}}
  \;\cdot\;
  \Bigl(\tfrac{a}{K_{\rm p}'}\Bigr)^{1/d_{\mathrm{eff}}}
  \;\le\;
  \Bigl(\tfrac{bK_{\rm p}'}{aK_{\rm p}}\Bigr)^{1/d_{\mathrm{eff}}}\,\epsilon_{\rm p}'.
\]
That is, truncating from $K_{\rm p}'$ to $K_{\rm p}$ centers increases the radius by at most the factor
\[
\epsilon_{\rm p}
\;\le\;
\bigl(bK_{\rm p}'/aK_{\rm p}\bigr)^{1/d_{\mathrm{eff}}}\,\epsilon_{\rm p}'.
\]
Since $kL\ge K_{\rm p}'$, loading into above, we have
\[
\epsilon_{\rm p}
\;\le\;
\bigl(bkL/aK_{\rm p}\bigr)^{1/d_{\mathrm{eff}}}\,\epsilon_{\rm p}'.\]
By loading~\hyperlink{eq:E5-1}{(E5-1)} into the above, the performance guarantee of prompt alignment is given by
\hypertarget{eq:E5-2}{}
\[
\epsilon_{\rm p}\coloneqq d_H\bigl(\mathcal{S}_{\rm p},\,\mathcal{P}\bigr) 
\;\le\;
\alpha(\eta,k,L)\, (K_{\rm p})^{-1/d_{\mathrm{eff}}}\quad \text{where}\quad
\alpha(\eta,k,L)\coloneqq \eta\,\bigl(bkL/a\bigr)^{1/d_{\mathrm{eff}}}.
\tag{E5-2}
\]

\vspace{3mm}
\paragraph{Part A-2: Performance Guarantee of Visual Preservation} \leavevmode\\[2ex]
By previous work~\cite{moenning2003fast}, FPS achieves a \(2\)-approximation for the \(k\)-center problem:
\hypertarget{eq:E5-3}{}
\[
\epsilon_{\rm v} \leq 2\,\epsilon^\star(K_{\rm v}),
\tag{E5-3}
\]
where \(\epsilon^\star(K_{\rm v})\) is the optimal radius with \(K_{\rm v}\) centers.
Based on~\Cref{lem:2}, we have
\[
\mathcal{N}(\mathcal{V}, r) \leq b'\,r^{-d_{\mathrm{eff}}},
\]
thereby, the upper bound of optimal radius is given by
\[
\epsilon^\star(K_{\rm v}) \leq (b'/K_{\rm v})^{1/d_{\mathrm{eff}}}.
\]
By loading the above into~\hyperlink{eq:E5-3}{(E5-3)}, the performance guarantee of visual preservation is given by
\hypertarget{eq:E5-4}{}
\[
\epsilon_{\rm v}\coloneqq d_H(\mathcal{S}_{\rm v}, \mathcal{V}) \leq \beta\,(K_{\rm v})^{-1/d_{\mathrm{eff}}}, \quad\text{where}\quad \beta\coloneqq 2\,{b'}^{1/d_{\mathrm{eff}}}.
\tag{E5-4}
\]

\paragraph{Part A-3: Performance Guarantee of MoB} \leavevmode\\[2ex]
By substituting~\hyperlink{eq:E5-2}{(E5-2)} and~\hyperlink{eq:E5-4}{(E5-4)} into~\Cref{lem:1-relax}, the performance guarantee of the MoB is given by:
\[
\|\mathcal{F}(\mathcal{X})-\mathcal{F}(\mathrm{MoB}(\mathcal{X}))\|\;\le\;C_\ell\; \max\Bigl\{
     \alpha(\eta,k,L)\,(K_{\rm p})^{-1/d_{\mathrm{eff}}},
     \;\beta\,(K_{\rm v})^{-1/d_{\mathrm{eff}}}
 \Bigr\}
 \;+\;C_\ell\, \eta,\]
where
\(
\alpha(\eta,k,L)\;=\; \eta\,\bigl(b\,k\,L/a\bigr)^{1/d_{\mathrm{eff}}},\quad
\beta\;=\; 2\,{b'}^{1/d_{\mathrm{eff}}}.
\)

This completes the proof of Part A.

\paragraph{Part B: Complexity}\leavevmode \\[2ex]
Since $k\ll K_{\rm p}\le K\sim L\ll N$, we restrict our complexity analysis to the leading-order terms.

\paragraph{Part B-1: Normalization}\leavevmode \\[2ex]
MoB do a  $L_2$ normalization for each token $x\in\mathcal{X}\subseteq\mathbb{R}^d$; thus, the complexity is given by
\hypertarget{eq:E5-5}{}
\[
T_{\rm norm}=\mathcal{O}((N+L)\, d).
\tag{E5-5}
\]

\paragraph{Part B-2: Selection of Prompt Center}\leavevmode \\[2ex]
Firstly, MoB calculates the cosine similarity with each $p\in\mathcal{P}$ and $v\in\mathcal{V}$ via a matrix multiplication:
\[
\mathbf{M}_{\rm sim} =  \mathbf{P}\, \mathbf{V}^\top \in\mathbb{R}^{L\times N}
\quad\text{where}
\quad
\mathbf{V}\in\mathbb{R}^{N\times d}\ \ \text{and}\ \ \mathbf{P}\in\mathbb{R}^{L\times d}, 
\]
which leads a complexity of $T_{\rm step\, 1\mbox{-}1}=\mathcal{O}(N\,L\,d)$. Subsequent, MoB do a top-$k$ retrieval in the first dimension of $\mathbf{M}_{\rm sim}$ the select $k$ most closed centers for each prompt token $p\in\mathcal{P}$, which can be reduced to a partial sorting, thereby leading to a complexity of $T_{\rm step\, 1\mbox{-}2}=\mathcal{O}(N\,L\,\log k)$. Finally, MoB merge the selected result of each $p\in\mathcal{P}$, and truncated the top-$K_{\rm p}$ ones with largest similarity, leading to a $T_{\rm step\, 1\mbox{-}3}=\mathcal{O}(L\,k\,\log K_{\rm p})$. Consequently, the total complexity $T_{\rm p\mbox{-}select}$ of prompt center selection is given by:
\hypertarget{eq:E5-6}{}
\[
\begin{aligned}
    T_{\rm p\mbox{-}select}&=T_{\rm step\, 1\mbox{-}1}+T_{\rm step\, 1\mbox{-}2}+T_{\rm step\, 1\mbox{-}3},\\
    &=\mathcal{O}(N\,L\,d) + \mathcal{O}(N\,L\,\log k) + \mathcal{O}(L\,k\,\log K_{\rm p}),\\
    &= \mathcal{O}(N\,L\,d). 
\end{aligned}
\tag{E5-6}
\]
\paragraph{Part B-3: Selection of Visual Center}\leavevmode \\[2ex]
Initially, MoB calculates the minimum distance (used in FPS) with each visual token $v\in\mathcal{V}\setminus \mathcal{S}_{\rm p}\coloneqq \mathcal{V}'$ and the selected prompt centers via a matrix multiplication together with an argmin operator:
\[
\mathbf{d}_{\rm FPS}=\operatorname*{arg\,min} \mathbf{V}'^\top\, \mathbf{S}_{\rm p}\in\mathbb{R}^{N-K_{\rm p}}
\quad\text{where}
\quad
\mathbf{V}'\in\mathbb{R}^{(N-K_{\rm p})\times d}\ \ \text{and}\ \ \mathbf{S}_{\rm p}\in\mathbb{R}^{K_{\rm p}\times d}, 
\]
thus, the complexity is given by
\[
\begin{aligned}
T_{\rm step\, 2\mbox{-}1}&=\underbrace{\mathcal{O}((N-K_{\rm p})\,K_{\rm p}\,d)}_{\mathrm{matrix\ \ multiplication}}
+\underbrace{\mathcal{O}((N-K_{\rm p})\,K_{\rm p})}_{\mathrm{argmin}},   \\
                         &=\mathcal{O}((N-K_{\rm p})\,K_{\rm p}\,d).\\
\end{aligned}
\]
Subsequently, in $K-K_{\rm p}$ iterations, MoB add the tokens with largest minimum distance with an argmax operator in $\mathbf{d}_{\rm FPS}$, and update the $\mathbf{d}_{\rm FPS}$ with an inner production together with an $N-K_{\rm p}$-dimensional element-wise comparison; thus the complexity is given by
\[
\begin{aligned}
T_{\rm step\, 2\mbox{-}2}&=\underbrace{\mathcal{O}((N-K_{\rm p})(K-K_{\rm p}))}_{\mathrm{argmax}}+\underbrace{\mathcal{O}((K-K_{\rm p})\,N\, d)}_{\mathrm{inner\ \ productioin}} + \underbrace{\mathcal{O}((K-K_{\rm p})\,d)}_{\mathrm{ele-wise\ \ comparision}},  \\
&=\mathcal{O}((K-K_{\rm p})\,N\, d).
\end{aligned}
\]
Consequently, the total complexity $T_{\rm v\mbox{-}select}$ of visual center selection is given by:
\hypertarget{eq:E5-7}{}
\[
\begin{aligned}
    T_{\rm v\mbox{-}select}&=T_{\rm step\, 2\mbox{-}1}+T_{\rm step\, 2\mbox{-}2},\\
                           &=\mathcal{O}((N-K_{\rm p})\,K_{\rm p}\,d)+\mathcal{O}((K-K_{\rm p})\,N\, d),\\
                           &=\mathcal{O}(N\,K\,d).
\end{aligned}
\tag{E5-7}
\]

\paragraph{Part B-4: Totally complexity}\leavevmode \\[2ex]
By~\hyperlink{eq:E5-5}{(E5-5)},~\hyperlink{eq:E5-6}{(E5-6)} and~\hyperlink{eq:E5-7}{(E5-7)}, the totally complexity of MoB is given by
\[
\begin{aligned}
    T_{\text{MoB}}&= T_{\rm norm} + T_{\rm p\mbox{-}select}+T_{\rm v\mbox{-}select},\\
                  &=\mathcal{O}((N+L)\, d)+ \mathcal{O}(N\,L\,d) + \mathcal{O}(N\,K\,d),\\
                  &=\mathcal{O}(N\,L\,d) + \mathcal{O}(N\,K\,d),\\
                  &=\mathcal{O}(N\,(L+K)\,d).
\end{aligned}
\]
This completes the proof of Part B.

Combining the Part A \& B, we complete the proof.

\end{proof}

\end{document}